\documentclass[sn-mathphys,iicol]{sn-jnl}

\jyear{2023}%

\theoremstyle{thmstyleone}%
\theoremstyle{thmstyletwo}%

\theoremstyle{thmstylethree}%

\usepackage{float} 

\usepackage{tabularx}
\usepackage{longtable}
\usepackage{booktabs}
\usepackage{colortbl}

\usepackage{graphicx}
\usepackage{comment}
\usepackage{amsmath,amssymb} % define this before the line numbering.
\usepackage{color}
\newcommand{\tabincell}[2]{\begin{tabular}{@{}#1@{}}#2\end{tabular}}
\usepackage{makecell}
\usepackage{caption}
\usepackage{enumerate}
\usepackage{multirow}
\usepackage{multicol}
\usepackage{subfigure}
\usepackage{float}
\usepackage{bbding}
\usepackage{xcolor}

\usepackage{mathtools}
\usepackage[linesnumbered,ruled,vlined,algo2e]{algorithm2e}
\SetCommentSty{mycommfont}

\newcommand{\algorithmicbreak}{\textbf{break}}
\newcommand{\BREAK}{\algorithmicbreak}

\newcommand{\ie}{\textit{i}.\textit{e}.}
\newcommand{\eg}{\textit{e}.\textit{g}.}

\begin{document}

\title[SOTVerse]{SOTVerse: A User-defined Task Space of Single Object Tracking}

\author[1,2]{\fnm{Shiyu} \sur{Hu}}\email{hushiyu2019@ia.ac.cn}

\author[1,2]{\fnm{Xin} \sur{Zhao}}\email{xzhao@nlpr.ia.ac.cn}

\author[1,2,3]{\fnm{Kaiqi} \sur{Huang}}\email{kqhuang@nlpr.ia.ac.cn}

\affil[1]{\orgdiv{School of Artificial Intelligence}, \orgname{University of Chinese Academy of Sciences}, \orgaddress{\street{No.19(A) Yuquan Road}, \city{Beijing}, \postcode{100049}, \country{China}}}

\affil[2]{\orgdiv{Institute of Automation}, \orgname{Chinese Academy of Sciences}, \orgaddress{\street{95 Zhongguancun East Road}, \city{Beijing}, \postcode{100190}, \country{China}}}

\affil[3]{\orgdiv{Center for Excellence in Brain Science and Intelligence Technology}, \orgname{Chinese Academy of Sciences}, \orgaddress{\street{320 Yue Yang Road}, \city{Shanghai}, \postcode{200031}, \country{China}}}

\abstract{
    Single object tracking (SOT) research falls into a cycle -- trackers perform well on most benchmarks but quickly fail in challenging scenarios, causing researchers to doubt the insufficient data content and take more effort to construct larger datasets with more challenging situations. However, inefficient data utilization and limited evaluation methods more seriously hinder SOT research. The former causes existing datasets can not be exploited comprehensively, while the latter neglects challenging factors in the evaluation process. In this article, we systematize the representative benchmarks and form a \textbf{s}ingle \textbf{o}bject \textbf{t}racking meta\textbf{verse} (\textbf{SOTVerse}) -- a user-defined SOT task space to break through the bottleneck. We first propose a \textbf{3E Paradigm} to describe tasks by three components (\ie, environment, evaluation, and executor). Then, we summarize task characteristics, clarify the organization standards, and construct SOTVerse with 12.56 million frames. Specifically, SOTVerse automatically labels challenging factors per frame, allowing users to generate user-defined spaces efficiently via construction rules. Besides, SOTVerse provides two mechanisms with new indicators and successfully evaluates trackers under various subtasks. Consequently, SOTVerse first provides a strategy to improve resource utilization in the computer vision area, making research more standardized and scientific. The SOTVerse, toolkit, evaluation server, and results are available at http://metaverse.aitestunion.com.
    }

\keywords{Single object tracking, experimental environment, evaluation system, performance analysis.}

\maketitle

\section{Introduction}
\label{sec:introduction}

\begin{figure*}[t!]
	\centering
	\includegraphics[width=\textwidth]{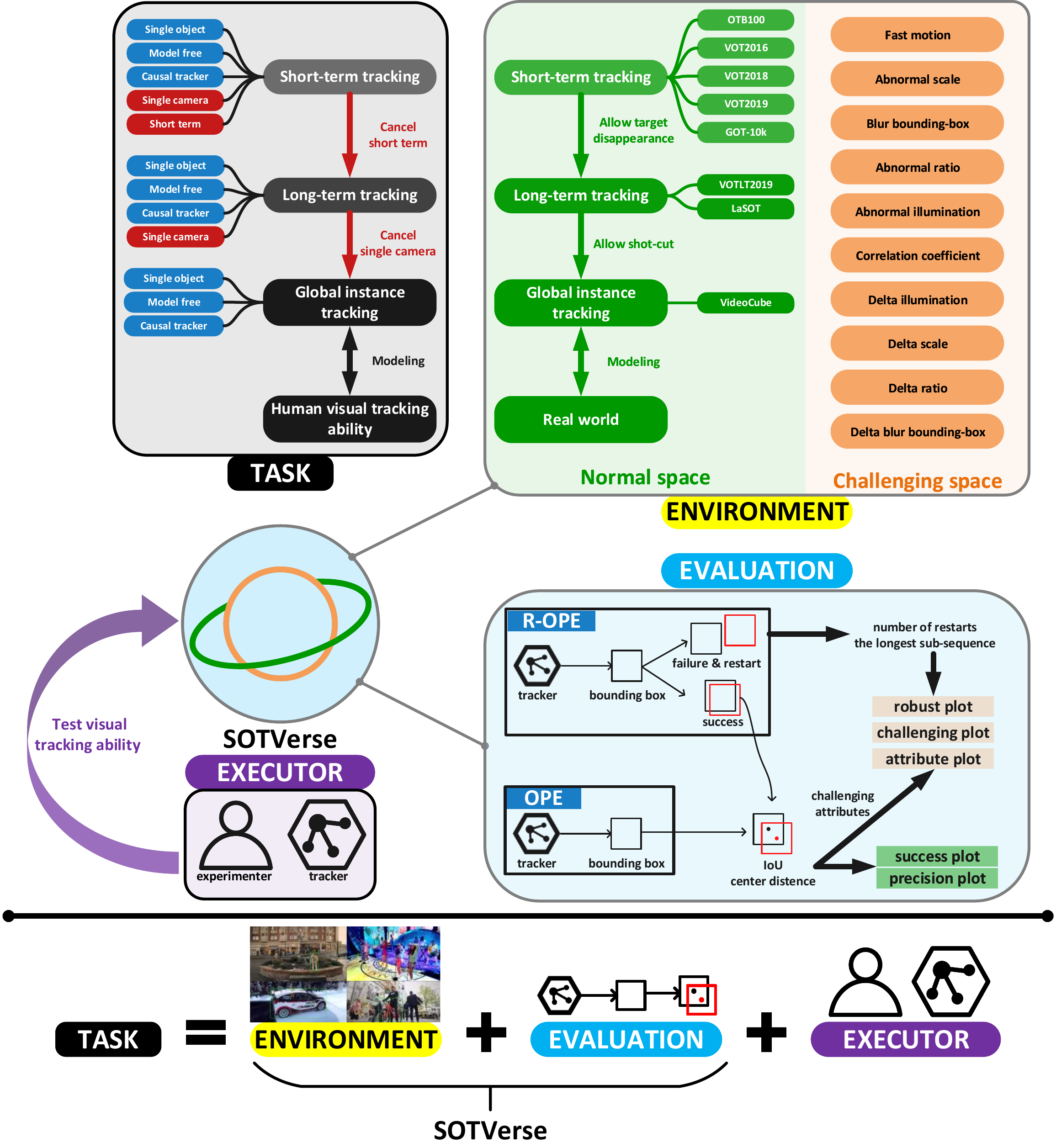}
	\caption{The \textbf{3E Paradigm} to describe the SOT task. A computer vision task can be characterized by three elements (environment, evaluation, and executor). 
	(\textbf{TASK}) For the SOT task, constraints contained in the definition are gradually eliminated during development.
	(\textbf{ENVIRONMENT}) Environment portrays task characteristics. We first select eight representative datasets to form the SOTVerse and then label multiple challenging attributes for each frame. Users can quickly extract related sub-sequences for different tasks, such as selecting abnormal ratio sub-sequences to create a single deformable object tracking space.
	(\textbf{EVALUATION}) SOTVerse provides diverse evaluation mechanisms and evaluation indicators to measure performance.
	(\textbf{EXECUTOR}) Both algorithms and human experimenters can test their visual tracking capabilities via SOTVerse.
	}
	\label{fig:overall}
	\end{figure*}

As the fundamental computer vision task, single object tracking (SOT \cite{VOT2013,OTB2015,LaSOT,GOT-10k,GIT}, \ie, locates a user-specified moving target in a video) aims to model the powerful human dynamic vision ability \cite{jw1962effect,biederman1987recognition,lee1999learning,mcleod2003fielders,land2000eye,beals1971relationship,burg1966visual,kohl1991comparative}, and has been widely used in daily application scenarios like self-driving \cite{self-driving,kong2022human,dendorfer2021motchallenge}, intelligent monitoring \cite{yoon2019structural,intelligent-monitoring,intelligent-monitoring2}, augmented reality \cite{zhang2015good,abu2018augmented,gauglitz2011evaluation} and robot navigation \cite{robot-navigation,robot-navigation2,ramakrishnan2021exploration}. 
When we look back at the evolution of SOT task, we can find that the task definition has drifted three times -- from short-term tracking \cite{OTB2013,OTB2015} to long-term tracking \cite{LaSOT,VOT2019}, and then to spatiotemporal change tracking \cite{GIT}. Obviously, the expansion of task definition prompts SOT to gradually model the human tracking vision ability and evolve towards general vision intelligence.

During the humanoid process of task definition, researchers have also contributed to constructing comprehensive benchmarks, aiming to provide high-quality \textit{datasets} and scientific \textit{evaluation} methods for algorithms.
The first systematic SOT benchmark OTB \cite{OTB2013} was successfully released in 2013. In the following decade, more and more benchmarks have been successfully constructed with larger dataset scales, and richer video content \cite{LaSOT,GOT-10k,GIT}. At the same time, several researchers \cite{ST-TIP,ST-TPAMI,LT-performance} also design various evaluation mechanisms to accomplish performance analysis via different perspectives. These benchmarks provide profitable environments and standardized evaluation processes, greatly facilitating the development of data-driven trackers. 

Through the above analyses, the ideal research route should be an upward spiral: a more human-like task definition promotes a more complicated benchmark construction, and ultimately guides to more intelligent algorithms. However, some bad cases show that current research falls into a cycle -- state-of-the-art (SOTA) algorithms perform well on most benchmarks but quickly fail when facing challenging factors in real application scenarios, causing researchers to doubt the insufficiency of benchmarks; thus, researchers usually spend a lot of effort constructing a larger dataset to solve this problem (\eg, the scale of SOT benchmarks in the past decade has been expended nearly 250 times). But many actual examples, like the frequent self-driving accidents, indicate that \textit{only expanding the dataset scale cannot break this bottleneck.}

To find the core reasons for this phenomenon, we first analyze existing issues separately from the perspective of datasets and evaluation:

\begin{itemize}
\item \textbf{For the dataset aspect, existing data has not been exploited effectively.} SOT task has evolved different characteristics during the development process, which is the primary reason yielding benchmarks to follow miscellaneous data collection rules. This phenomenon leads to inconsistencies in the construction process, causing experimental environments to become isolated. Existing datasets can only be compared in superficial features like dataset scale but are difficult to contrast in vital components such as content difficulty (\eg, challenging factors are always selected to represent the difficulty, while various benchmarks annotate challenging attributes by different metrics, and many classical benchmarks only provide sequence-level annotations rather than frame-level). 
% Experimental conclusions obtained based on isolated datasets are also scattered, and it is difficult to form a comprehensive evaluation of the algorithm's performance. 
When researchers aim to investigate tasks in more complex scenarios, they usually reconstruct a larger dataset rather than extracting relevant data from existing datasets.
\item \textbf{For the evaluation aspect, limitations of evaluation methods lead to neglect of challenging factors.} Multiple researchers always overlook the shortcomings of evaluation methods. In fact, existing benchmarks mainly run trackers on sequences, get frame-by-frame scores, and finally calculate the average value to represent the overall performance. However, SOT is a sequential decision task and is seriously affected by challenging factors (\eg, frames with fast motion or tiny objects), while regular tracking sequences in benchmarks are usually composed of many \textit{simple frames} and scant \textit{challenging frames}. Thus, bad performance is ignored after averaging due to the low proportion of challenging frames.
\end{itemize}

The above problems hinder related research, increase hardship for resource integration and utilization, and ultimately create bottlenecks in research. In this work, we systematize the representative benchmarks and construct a comprehensive \textbf{s}ingle \textbf{o}bject \textbf{t}racking meta\textbf{verse} named \textbf{SOTVerse} to solve the issues, as shown in Figure~\ref{fig:overall}. 
Like DeepMind \cite{XLand} defines reinforcement learning tasks as world, game, and co-players, we propose a \textbf{3E Paradigm} to describe computer vision tasks by three components (\ie, \textit{environment}, \textit{evaluation}, and \textit{executor}).
Among them, datasets provide the \textit{environment} to portray task characteristics, \textit{evaluation} methods measure performance from multiple aspects, and \textit{executors} can estimate their visual tracking abilities via SOTVerse.

\begin{figure*}[t!]
	\centering
	\includegraphics[width=0.75\textwidth]{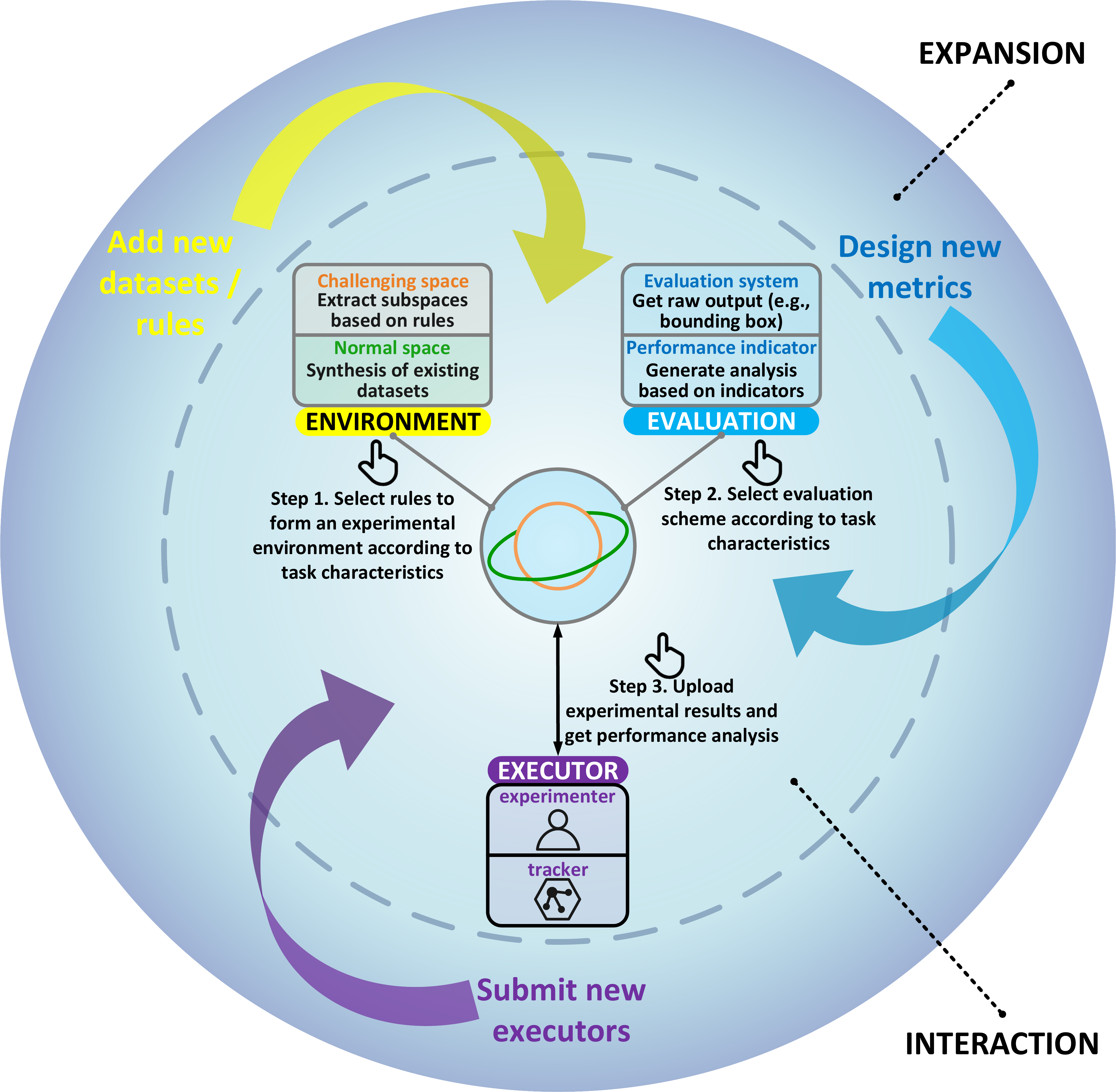}
	\caption{The user-defined process of SOTVerse, can be divided into interaction and expansion. 
	(\textbf{INTERACTION}) Users need three steps to finish the operation: first, select data extraction rules according to task characteristics to generate an experimental environment. Then, determine the appropriate evaluation system and performance indicators. Finally, upload the experimental results and obtain the corresponding performance analysis.
	(\textbf{EXPANSION}) Users can expand the SOTVerse by adding new datasets or extraction rules, designing new metrics, and submitting new executors.
	}
	\label{fig:goal}
	\end{figure*}

Specifically, to integrate different environments into SOTVerse, we first correspond task characteristics with data collection rules, clarify the organization standards, and construct them as the \textit{normal space}. 
In particular, we also provide various challenging attribute labels for each frame, allowing users to extract related sub-sequences from SOTVerse and efficiently generate \textit{challenging spaces} with their research purpose.
Besides, to overcome the limitations of traditional evaluation methods, SOTVerse provides three novel indicators to focus on tracking robustness under challenging factors. 

Obviously, SOTVerse is a customizable and extensible space. We summarize the contributions as follows:

\begin{itemize}
\item \textbf{A paradigm to describe computer vision tasks.} Computer vision tasks can be characterized by \textit{environment}, \textit{evaluation}, and \textit{executor}. Figure~\ref{fig:overall} illustrates the \textbf{3E paradigm} by analyzing SOT in detail: we synthesize the \textit{environment} and \textit{evaluation} to form \textbf{SOTVerse} -- a user-defined single object tracking task space, and conduct experiments in this space to judge \textit{executors}' tracking ability. Definitely, this paradigm can be expanded to describe different visual tasks and help users improve their research efficiency.
\item \textbf{A comprehensive and user-defined environment.} Through precise analyses of the task definition, we organize existing benchmarks to form the \textit{environment} of SOTVerse. It includes 12.56 million frames and frame-level challenging attribute labels to model the real world. Notably, the thresholds for determining challenging factors are selected by their distribution on the whole environment. Besides, an environment generation method can efficiently help researchers form their own task space. Therefore, unlike traditional benchmarks' isolated and static design, SOTVerse is a comprehensive and dynamic experimental environment.
\item \textbf{A thoroughgoing evaluation scheme.} We first point out the limitations of existing systems and indicators through detailed analysis; then design a new evaluation scheme, which includes two mechanisms and new metrics to satisfy various tasks. 
\item \textbf{Various experimental executors and detailed analysis.} We conduct extensive experiments in the SOTVerse and perform performance analysis on various executors. Experimental results show that challenging factors severely hamper tracking performance -- the proposed challenging plot reveals that high scores are mainly obtained in normal frames, while the success rate of most trackers is less than 0.5 under challenging situations. Finally, we point out the necessity of the re-initialization mechanism for evaluation in long sequences. These results indicate the shortcomings of existing work and verify the effectiveness of the evaluation scheme in SOTVerse.
\end{itemize}

We provide a comprehensive online platform at http://metaverse.aitestunion.com to help users operate SOTVerse. The user-defined process illustrated by Figure~\ref{fig:goal} can be divided into \textit{interaction} and \textit{expansion}. With our platform, users can select the environment generation method according to task characteristics and directly download the generated experimental environment. Besides, an open-sourced toolkit is available to accomplish the evaluation process. Finally, users can upload the experimental results and obtain the corresponding performance analysis.
In addition, we accommodate users to expand SOTVerse. For example, users can provide new datasets or develop new environment generation methods to enrich the experimental environment. They can also formulate new evaluation mechanisms and quickly verify the effectiveness of various subtasks.

Evidently, SOTVerse allows users to customize tasks according to their own research purposes. It not only makes research more targeted, but also can significantly improve research efficiency. Furthermore, the 3E Paradigm successfully performed in the SOT area provides an excellent example, which can be referenced by various visual or other domain tasks in the future.

The rest of this paper is organized as follows. Section~\ref{sec:related_work} provides a review of the SOT task. Section~\ref{sec:metaverse} introduces the design principles of SOTVerse. Section~\ref{sec:experiments} describes the experimental results and detailed analysis. Finally, the conclusions and discussions of future work are summarized in Section~\ref{sec:conclusion}.

\begin{figure*}[ht!]
\centering 
\includegraphics[width=\textwidth]{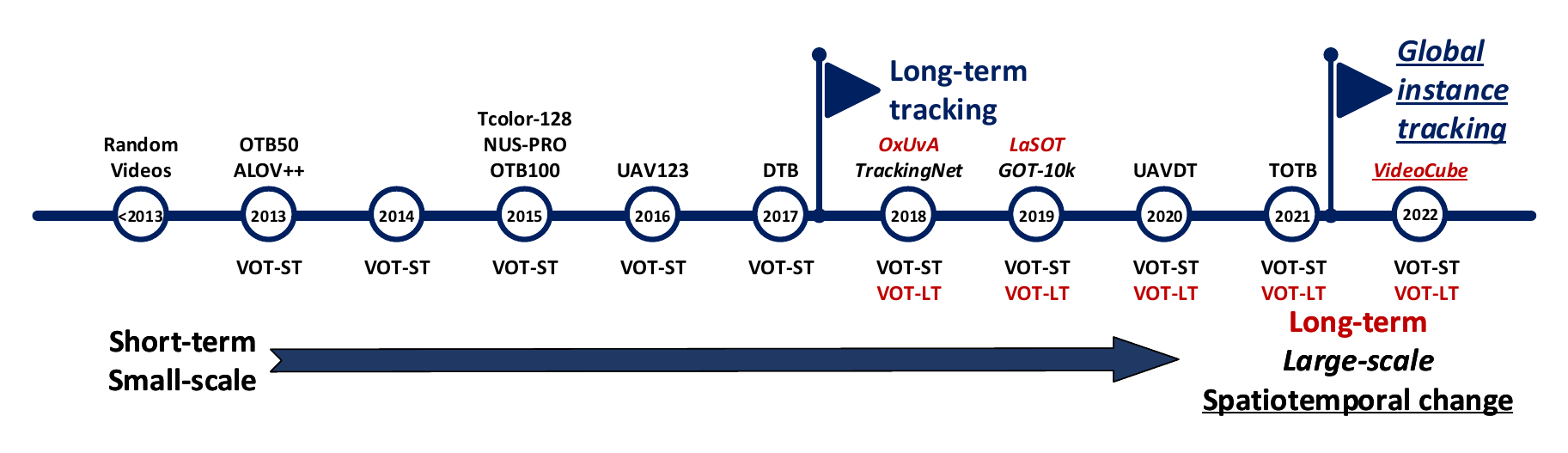}
\caption{The development trend of SOT benchmarks. The \textcolor{red}{red} font represents long-term tracking datasets, the \textit{italic} represents large-scale datasets, and the \underline{underscore} represents the spatiotemporal variations. Clearly, SOT benchmarks are developing toward larger-scale, longer-term, and more challenging tracking.}
\label{fig:timeline}
\end{figure*}

\section{Related Work}
\label{sec:related_work}

\subsection{Task}
Understanding a task includes (1) task definition analysis and (2) task description paradigm. The former is an external description to distinguish a task from others through strict boundaries. The latter is an internal description representing a task through corresponding environment and execution standards.

\subsubsection{Task Definition Analysis}

SOT is usually defined as only providing the initial position of an arbitrary object and continuously locating it in a video sequence \cite{OTB2013,OTB2015}.
Since 2013, researchers have proposed several influential benchmarks \cite{OTB2015,TrackingNet,LaSOT,GOT-10k,GIT} -- the organized datasets and unified metrics promote the SOT research.  
However, limited by the research level, early definition adds additional constraints to simplify the task. The influential VOT competition limits this task to five keywords: \textit{single-target}, \textit{model-free}, \textit{causal trackers}, \textit{single-camera}, and \textit{short-term} \cite{ST-TPAMI}. The first three keywords (\textit{single-target}, \textit{model-free}, \textit{causal trackers}) correspond to the original definition and are the criteria for distinguishing SOT from other visual tasks (\eg, multi-object tracking \cite{MOT1,MOT2} and visual instance detection \cite{VOD1,VOD2,YT-BB,ImageNet-VID}). In comparison, the latter two keywords (\textit{single-camera} and \textit{short-term}) are constraints added to simplify research in the early stage.

The development of SOT is continuously removing hidden constraints and closer to the essential definition, as shown in Figure~\ref{fig:timeline}. Since 2018, some researchers have withdrawn \textit{short-term} and proposed long-term tracking \cite{OxUvA,LaSOT,VOT2019}. 
In 2022, researchers further remove the \textit{single-camera} constraint and propose the global instance tracking (GIT) \cite{GIT}, which is supposed to search an arbitrary user-specified instance in a video without any assumptions about camera or motion consistency. Clearly, GIT realizes the basic definition of SOT by gradually removing the constraints.

\subsubsection{Task Description Paradigm}

The description paradigm analyzes a task from multiple dimensions, establishes specific operating rules, and provides experimental environments for executors.
In other words, the paradigm transforms a monotonous task definition into several operational elements concretely.
In 2021, DeepMind \cite{XLand} provides a task description paradigm for the reinforcement learning task, consisting of a game with a world and co-players. The world is composed of various static and dynamic elements, which can quickly combine an adapted environment according to the task characteristics. A series of goals consist of the game, which aims to guide players to maximize total reward. Players are agents who perform tasks in the world according to the game rules.

Although the computer vision area does not propose a specific task description paradigm like reinforcement learning, different researchers have tried to characterize the task from three aspects: \textit{environment}, \textit{evaluation}, and \textit{executor}. Correspondingly, relevant datasets provide the execution \textit{environment} of the task; \textit{evaluation} methods are similar to the game rules, which measure the performance via different metrics; \textit{executors} are the task entertainer, including related algorithms and human experimenters. 

The following parts introduce the experimental environment (datasets) and evaluation methods of SOT in detail.

\subsection{Environment}

\begin{figure*}[htbp!]
\centering
\includegraphics[width=\textwidth]{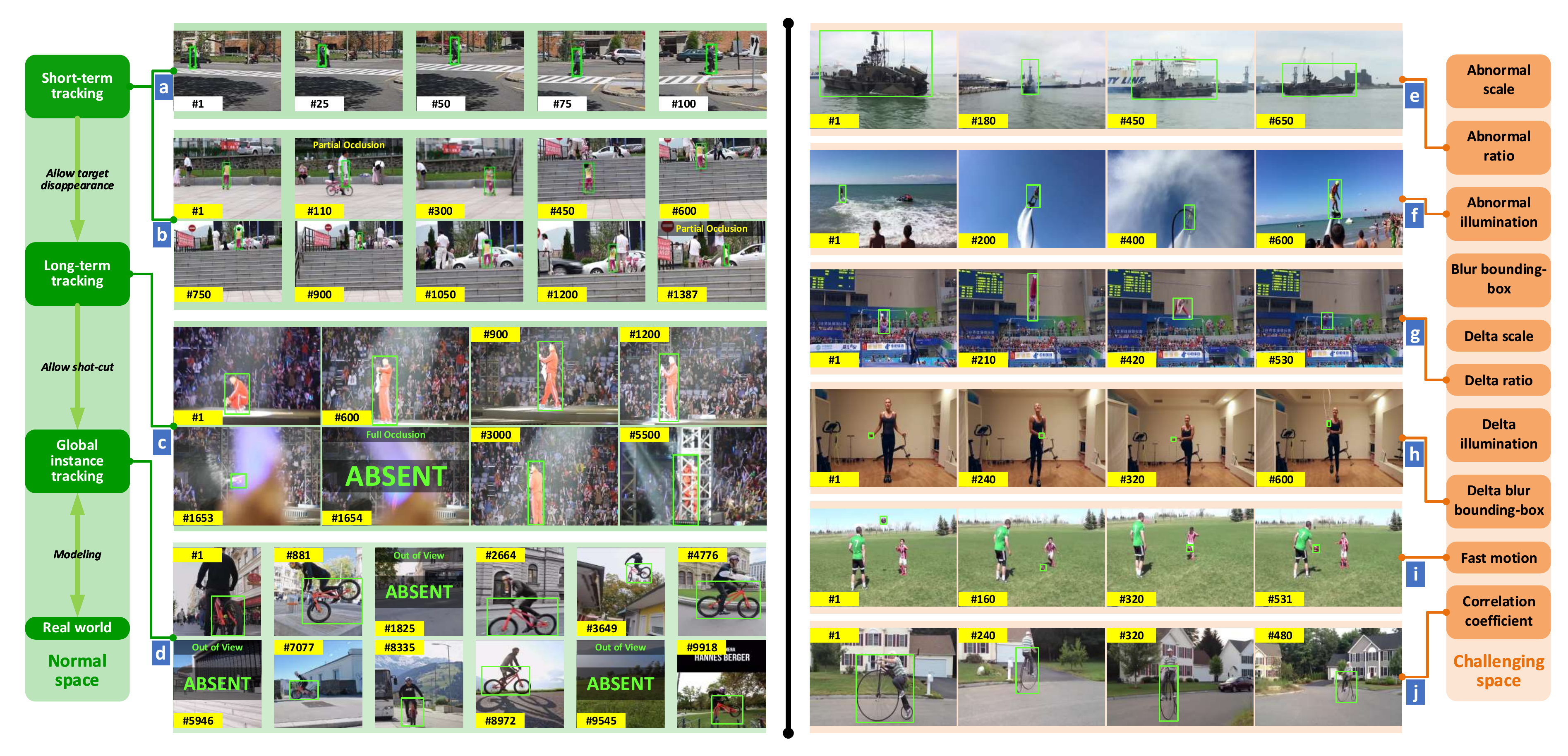}
\caption{Examples of normal space $E_n$ and challenging space $E_c$. Sequences in $E_n$ are selected from existing datasets, while sequences in $E_c$ are obtained based on space construction rules.}
\label{fig:environment}
\end{figure*}

High-quality datasets play a vital role in SOT development. 
Early datasets represented by VIVID \cite{VIVID}, CAVIAR \cite{CAVIAR}, and PETS \cite{PETS} mainly focus on surveillance scenarios, which aim to track humans or cars but lack canonical build standards.
Since 2013, well-organized benchmarks represented by OTB \cite{OTB2013,OTB2015} are mainly designed for short-term tracking tasks \cite{VOT2017,TC128,UAV,NUS-PRO,Nfs}, which assumes no complete occlusion or target out-of-view happened in this video, as shown in Figure~\ref{fig:environment} (a) and (b). 
The average duration of short-term datasets is always less than one minute, and the following benchmarks mainly innovate in video content.
(\eg, TC-128 \cite{TC128} evaluates color-enhanced trackers on color sequences; NUS-PRO \cite{NUS-PRO} focuses on tracking pedestrian and rigid objects; UAV123 \cite{UAV} assesses unmanned aerial vehicle tracking performance; PTB-TIR \cite{PTB-TIR} and VOT-TIR \cite{VOT2016} are thermal tracking datasets; GOT-10k \cite{GOT-10k} includes 563 object classes based on the WordNet \cite{WordNet}).

Recently, several new benchmarks represented by LaSOT \cite{LaSOT} have proposed long-term tracking to satisfy the demands of real scenarios \cite{TLP,OxUvA}. 
However, it is hard to separate the short-term and long-term in the time dimension. Although short-term videos are usually shorter than one minute, only adopting \textit{one minute} as the task boundary is biased. Therefore, the VOT competition proposes a new criterion -- a task that allows the target to disappear completely can be regarded as long-term tracking \cite{LT-performance}. 
In contrast to the two criteria, allowing the object to disappear for a short period is more suitable as the decisive factor for long-term tracking. By removing the constraint hidden in the definition of short-term tracking that the target should present in the tracking process, the experimental environment can include more long-term videos to achieve the expansion from a short to a long term.
As shown in Figure~\ref{fig:environment} (c), a target may disappear utterly due to being out of view or be fully occluded, which is excluded in the short-term tracking environment.

Nonetheless, the implicit continuous motion assumption restricts long-term tracking environments to single-camera and single-scene, which is still far from the application scenarios of SOT. Thus, the global instance tracking environment named VideoCube is proposed \cite{GIT}. It includes videos with shot-cut and scene-switching to model the real world comprehensively (Figure~\ref{fig:environment} (d)).

Existing works build environments from different perspectives with various rules, but no one has tried to unify the environments. When researchers try to analyze problems from new perspectives, they have to build corresponding datasets from scratch, significantly reducing research efficiency.
This status inspired us to summarize and uniform existing environments to construct SOTVerse, and help researchers generate experimental environments effectively.

\subsection{Evaluation}

\subsubsection{Evaluation System}

Initialize a tracker in the first frame and continuously record the tracking results -- this evaluation system is a one-pass evaluation (OPE). To utilize the failure information and analysis the breakdown reasons, OTB \cite{OTB2015} benchmark offers a re-initialization mechanism (OPER).
The re-initialization mechanism also plays a vital role in VOT competitions \cite{VOT2016,VOT2018,VOT2019}. Trackers will be re-initialized after the assessment system detects tracking failure. Unlike the OPER mechanism that can be re-initialized in any frame, VOT2020 selects several anchors as re-initialization frames. The intervals between anchors are constant (\eg, 50 frames), and artificial examination is adopted to ensure each anchor contains complete target information. Trackers are reset in the nearest anchor once they fail, and the tracking direction (forward or backward) may change to track on the longest sub-sequence split by the anchor.

Most long-term tracking benchmarks \cite{LaSOT,OxUvA} select OPE mechanism as an evaluation system. The VOTLT competition \cite{LT-performance}, which regards \textit{target disappearance} as the manifestation of long-term tracking, hopes trackers can re-locate the target and exclude the restart mechanism for long-term evaluation. 

\subsubsection{Performance Indicator}

Most evaluation indicators can be summarized from precision, successful rate, and robustness. Most indicators select the positional relationship between predicted result $p_t$ and ground-truth $g_t$ in the $t$-th frame to accomplish calculation: 

\begin{itemize}
	\item \textit{Precision} proposed by OTB \cite{OTB2015} measures the euclidean distance between center points of $p_t$ and $g_t$ in pixels. Calculating the proportion of frames whose distance is less than a threshold and drawing the statistical results based on different thresholds into a curve generates the \textit{precision plot}. Typically, 20 pixels are selected as a threshold to rank trackers. To eliminate the influence of object scale, TrackingNet \cite{TrackingNet} adopts ground-truth scale (width and height) to normalize the center distance. VideoCube \cite{GIT} provides a normalized precision metric to eliminate the effect of target size.
	\item The overlap of $p_t$ and $g_t$ is calculated by $\Omega (p_t,g_t) = \frac{p_t\bigcap g_t}{p_t\bigcup g_t}$. Frames with $\Omega (p_t,g_t)$ greater than a threshold are defined as successful tracking, and the \textit{successful rate (SR)} measures the percentage of successfully tracked frames. Drawing the results based on various thresholds into a curve is the \textit{success plot}.
	\item \textit{Robustness} evaluates the stability of tracking. VOT competition\cite{ST-TPAMI} initially applies the number of re-initialization $M$ to calculate robustness, then converts it as $R_s = e^{-SM}$ to interpret the reliability. GIT task \cite{GIT} considers the degree of frame variation and the number of failures to measure the robustness.
\end{itemize}

Besides, several long-term tracking benchmarks \cite{OxUvA, LT-performance} require trackers to output  disappearance-judgment for calculating the \textit{tracking accuracy}, \textit{recall}, and \textit{F-measure}.

The above introduction illustrates that existing evaluation systems and performance indicators are fragmented. More importantly, the impact of challenging factors has long been identified \cite{godec2013hough,han2008sequential,nejhum2008visual,collins2003mean,kwon2009tracking}, but ignored by existing mechanisms, which mainly focus on the all-around performance of complete sequences. Therefore, when constructing the SOTVerse, we first clarify the calculation formula of performance indicators, then conduct experiments on various tasks to explore the applicable scope of different evaluation methods.

\begin{table*}[htbp!]
	\begin{center}
	\caption{The symbol of task description paradigm in SOT.}
        \resizebox{\textwidth}{!}{
	\begin{tabular}{llll}
	\toprule
	\multicolumn{2}{l}{\textbf{Symbol}} & {\textbf{Implication}} & {\textbf{Definition in SOT}} \\
	\midrule
	$\mathbb{S}$ & Task & The academic definition of a task & Locating a user-specified target in a video \\
	\midrule
	$S$ & Subtask & \tabincell{l}{Subtask formed by adding constraints \\ to the original task definition} & \tabincell{l}{Normal subtask $S_{n}$ (\eg, short-term tracking, long-term \\ tracking, global instance tracking) and challenging subtask \\ $S_{l}$ (\ie, locating the target in the challenging situation)} \\ 
	\midrule
	$\mathbb{E}$ & Environment & \tabincell{l}{An execution space of a task, usually \\ organized by datasets} & Combination of representative SOT datasets \\ 
	\midrule
	$E$ & Subspace & Execution space of subtask & Normal sub-space $E_{n}$ and challenging sub-space $E_{l}$ \\ 
	\midrule
	$\mathbb{M}$ & Evaluation &	Methods to evaluate the abilities of executors & Evaluation system $M_{s}$ and performance indicator $M_{p}$ \\
	\midrule
	$\mathbb{T}$ & Executor & Task executor & Tracker $T_{t}$ or human subject $T_{h}$ \\
	\botrule
	\end{tabular}}
	\label{table:symbol}

	\end{center}
\end{table*}

\begin{table*}[htbp!]
	\begin{center}
		\caption{The representative benchmarks selected to form the normal space of SOTVerse. OTB \cite{OTB2015}, VOT series \cite{VOT2016, VOT2018, VOT2019} and GOT-10k \cite{GOT-10k} represent the short-term tracking benchmarks; VOTLT2019 \cite{VOT2019} and LaSOT \cite{LaSOT} represent the long-term tracking benchmarks; VideoCube \cite{GIT} represents the global instance tracking benchmark.}
            \resizebox{\textwidth}{!}{
		\begin{tabular}{llllllllllll}
		\toprule
		\multicolumn{3}{l}{\textbf{Environment}} & \textbf{Videos} & \tabincell{l}{\textbf{Min} \\ \textbf{frame}} & \tabincell{l}{\textbf{Mean} \\ \textbf{frame}} & \tabincell{l}{\textbf{Max} \\ \textbf{frame}} & \tabincell{l}{\textbf{Total} \\ \textbf{frame}} & \multicolumn{2}{l}{\textbf{Subtask}} & \tabincell{l}{\textbf{Target} \\ \textbf{absent}} & \tabincell{l}{\textbf{Shot-} \\ \textbf{cut}}\\
		\midrule
		\multicolumn{1}{l}{\multirow{5}{*}{$E_{n_{1}}$}} & $e_{1}$ & OTB2015 \cite{OTB2015} & 100 & 71 & 590 & 3,872 & 59K & \multicolumn{1}{l}{\multirow{5}{*}{$S_{n_{1}}$}}  & \multicolumn{1}{l}{\multirow{5}{*}{Short-term tracking}} & \XSolidBrush & \XSolidBrush \\
		\multicolumn{1}{l}{} & $e_{2}$ & VOT2016 \cite{VOT2016} & 60 & 41 & 357 & 1,500 & 21K & \multicolumn{1}{l}{} & \multicolumn{1}{l}{} & \XSolidBrush & \XSolidBrush  \\
		\multicolumn{1}{l}{} & $e_{3}$ & VOT2018 \cite{VOT2018} & 60 & 41 & 356 & 1,500 & 21K & \multicolumn{1}{l}{} & \multicolumn{1}{l}{} & \XSolidBrush & \XSolidBrush  \\
		\multicolumn{1}{l}{} & $e_{4}$ & VOT2019 \cite{VOT2019} & 60 & 41 & 332 & 1,500 & 20K & \multicolumn{1}{l}{} & \multicolumn{1}{l}{} & \XSolidBrush & \XSolidBrush \\
		\multicolumn{1}{l}{} & $e_{5}$ & GOT-10k \cite{GOT-10k} & 10,000 & 29 & 149 & 1,418 & 1.45M & \multicolumn{1}{l}{} & \multicolumn{1}{l}{} & \XSolidBrush & \XSolidBrush  \\
		\midrule
		\multicolumn{1}{l}{\multirow{2}{*}{$E_{n_{2}}$}} & $e_{6}$ & VOTLT2019 \cite{VOT2019} & 50 & 1,389 & 4,305 & 29,700 & 215K & \multicolumn{1}{l}{\multirow{2}{*}{$S_{n_{2}}$}}  & \multicolumn{1}{l}{\multirow{2}{*}{Long-term tracking}} &  \Checkmark & \XSolidBrush \\
		\multicolumn{1}{l}{} & $e_{7}$ & LaSOT \cite{LaSOT} & 1,400 & 1,000 & 2,502 & 11,397 & 3.5M & \multicolumn{1}{l}{} & \multicolumn{1}{l}{} &  \Checkmark & \XSolidBrush \\
		\midrule
		$E_{n_{3}}$ & $e_{8}$ & VideoCube \cite{GIT} & 500 & 4,008 & 14,920 & 29,834 & 7.46M & $S_{n_{3}}$ & Global instance tracking & \Checkmark & \Checkmark \\
		\botrule
		\end{tabular}}
		\label{table:dataset}
	\end{center}
 \end{table*}

\section{The Construction of SOTVerse Space}
\label{sec:metaverse}

\subsection{3E Paradigm}

As shown in Figure~\ref{fig:overall}, a computer vision task can be described by the combination of environment, evaluation, and executor.
Table~\ref{table:symbol} lists the related concepts in the SOT task.
We assume that $S$ denotes a subtask (\eg, short-term tracking task), $E$ is the corresponding experimental environment organized by several videos (\eg, short-term dataset), $M_{s}$ represents the set of evaluation systems (\eg, OPE mechanism), $M_{p}$ represent the set of performance indicators (\eg, precision), $T$ symbolizes the set of task executors (\eg, trackers and human subjects). 
Particularly, $\times$ represents the Cartesian product. Under the 3E Paradigm, the subtask can be represented as: 

\begin{equation} \label{equ:1}
	S = E \times M_{s} \times M_{p} \times T
	\end{equation}

\noindent
On the one hand, a complete SOT task space $\mathbb{S}$ can be obtained by integrating the various subtasks.
On the other hand, the set of environments, evaluation methods, and task executors can be separately symbolized as $\mathbb{E}$, $\mathbb{M}$, and $\mathbb{T}$, characterizing $\mathbb{S}$ as:

\begin{equation} \label{equ:2}
	\mathbb{S} = \left \{ S_{n_{1}} ,S_{n_{2}},\dots ,S_{c_{1}} ,S_{c_{1}},\dots \right \} = \mathbb{E} \times \mathbb{M} \times \mathbb{T}
\end{equation}

According to 3E Paradigm, we build a user-defined task space named SOTVerse, which integrates the existing SOT datasets into a large environmental space $\mathbb{E}$, and provides multiple indicators to combine a comprehensive evaluation space $\mathbb{M}$. With the help of SOTVerse, users can quickly extract relevant data to form the task environment and select appropriate evaluation methods for performance measurement.
The following parts introduce the experimental environment $\mathbb{E}$ and evaluation methods $\mathbb{M}$ in detail.

\subsection{Environment}

\begin{figure*}[htbp!]
	\centering
	\includegraphics[width=\textwidth]{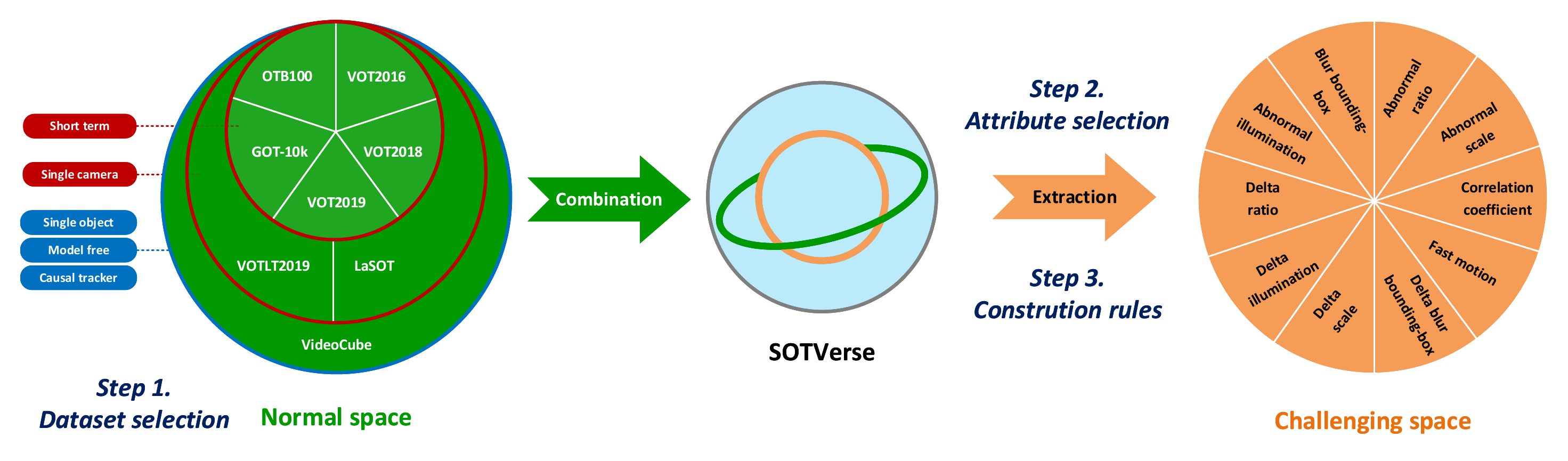}
	\caption{The combination process of SOTVerse. 
	First, representative datasets $e_{i}$ are chosen to form the normal space $E_{n}$ according to the relationship between SOT subtasks (short-term tracking $S_{n_{1}}$, long-term tracking $S_{n_{2}}$, global instance tracking $S_{n_{3}}$). 
	Second, we summarize the challenging factors into ten attributes and automatically label these attributes per frame. 
	Finally, we design space construction rules, which help users quickly extract eligible sub-sequences from SOTVerse to form a challenging space $E_{l}$ based on research goals.}
	\label{fig:sub-space}
	\end{figure*}

Figure~\ref{fig:sub-space} illustrates the combination process of SOTVerse, which can be split into three steps.

\subsubsection{Step One. Dataset Selection}

First, representative datasets $e_{i}$ are chosen to form normal space $E_{n}$ according to the relationship between subtasks (short-term tracking $S_{n_{1}}$, long-term tracking $S_{n_{2}}$, global instance tracking $S_{n_{3}}$).
Select benchmarks can cover all subtasks and reflect the characteristics of SOT. Table~\ref{table:dataset} illustrates that the normal space includes 12.56 million frames to simulate real application scenarios fully.

\subsubsection{Step Two. Attribute Selection}

We unify the attribute calculation and determine the abnormal judgment based on its distribution. All attributes are calculated from original files (sequences and ground-truth) without additional manual annotations. 
We split attributes into two categories: (1) \textit{static attributes} only relate to the current frame, while (2) \textit{dynamic attributes} record changes between consecutive frames.

For the $t$-th frame $F_{t}$ in the sequence $L$, we use four values ($x_{t}$, $y_{t}$, $w_{t}$, $h_{t}$) (\ie, the coordinate information of the upper left corner and the shape of the bounding-box) to represent the target bounding-box. The calculation rules are as follows:

\begin{itemize}
	\item The target \textit{ratio} is defined as $r_{t}=\frac{h_{t}}{w_{t}}$. Original target scale can be calculated via $s_{t}=\sqrt{w_{t}h_{t}}$, to further weaken the impact of image resolution, we calculate \textit{relative scale} by $s_{t}^{'}=\frac{s_{t}}{\sqrt{W_{t}H_{t}}}$ ($W_{t}$ and $H_{t}$ represent the image resolution of $F_{t}$). 
	\item Video recorded in special light conditions (\eg, dim light or blinding light) can be transferred to standard illumination by multiplying a correction matrix $C_{t}$ \cite{ShadeofGray}. \textit{Illumination} can be quantified by the Euclidean distance between $C_{t}$ and $\boldsymbol{1}^{1\times 3}$. 
	\item We use Laplacian transform \cite{Laplacian} to calculate the \textit{blur box} degree. We first convert the RGB bounding-box into gray-scale $G_{t}$, then convolve $G_{t}$ with a Laplacian kernel, and calculate the variance as sharpness. 
	\item Dynamic attributes are generated from the variation of static attributes. Correspondingly, we define the variations in two sequential frames as \textit{delta ratio}, \textit{delta relative scale}, \textit{delta illumination}, and \textit{delta blur box}.
	\item Besides, we use \textit{fast motion} to quantify the target center distance between consecutive frames by $d_{t}=\frac{\left\|c_{t}-c_{t-1}\right\|_{2}}{\sqrt{max(s_{i},s_{t-1})}}$.
	\item \textit{Correlation coefficient} measures the similarity between progressive frames. We select the Pearson product-moment correlation coefficient to calculate the image covariance, and the denominator is the product of the standard deviation. 
	\end{itemize}

\begin{figure*}[htbp!]
	\centering
	\includegraphics[width=\textwidth]{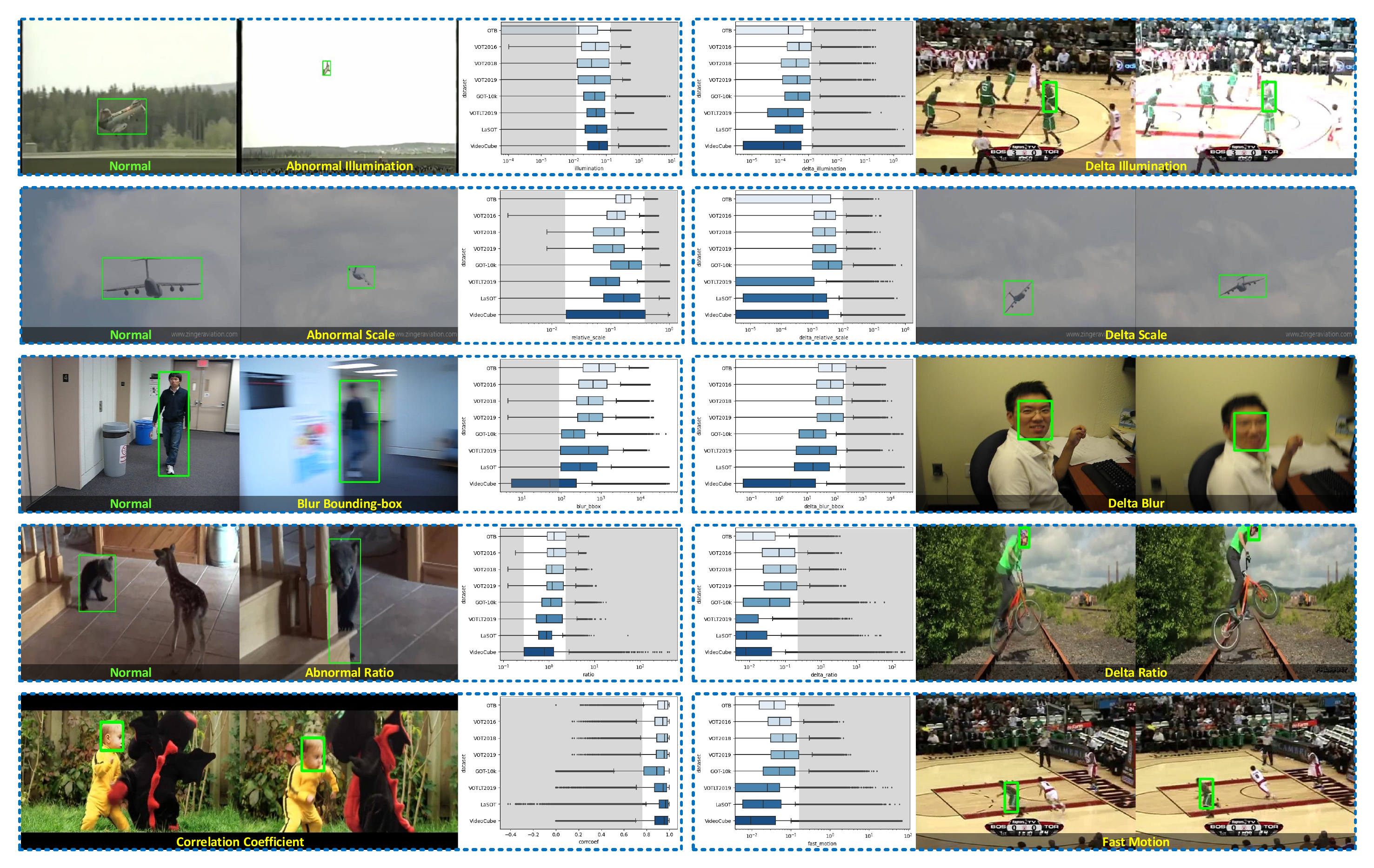}
	\caption{The attribute distribution and example on SOTVerse. We use a box-plot to illustrate the attribute distribution, the distribution boundaries of attribute values over the eight datasets are regarded as abnormal criteria. Notably, the distribution of the OTB \cite{OTB2015} in the abnormal illumination is significantly different from other datasets, since the partial sequences of OTB are grayscale, resulting in the calculation result being 0. Therefore, we remove the OTB before confirming the lower bound of the abnormal illumination. Similarly, we remove the VideoCube \cite{GIT}, which is obviously different from other datasets, to determine the boundary in the blur bounding-box.}
	\label{fig:attribute}
	\end{figure*}

We note that determining the threshold of abnormal attributes in existing benchmarks is subjective. For example, TrackingNet \cite{TrackingNet} and LaSOT \cite{LaSOT} regard the area smaller than 1000 pixels as low resolution (\ie, tiny object), while GOT-10k \cite{GOT-10k} considers the target smaller than half of the frames is tiny. 
Thus, we first ensure the above calculation formulas are applicable to all situations (\eg, we eliminate the influence of image resolution variation). 
The frame whose attribute value lies in the abnormal interval is defined as a \textit{challenging frame}; otherwise, it is a \textit{normal frame}.
To avoid the influence of subjective factors, we select abnormal thresholds via attributes' distribution in 12.56 million frames, as shown in Figure~\ref{fig:attribute} and Table~\ref{table:challenging}.
Consequently, our method excludes human interference and suits all benchmarks.

\subsubsection{Step Three. Space Construction Rules}

\begin{table*}[htbp!]
	\begin{center}
	\caption{The threshold for judging the abnormal value and the corresponding challenging space $E_c$.}
        \small
	\begin{tabular}{llllll}
	\toprule
	\multicolumn{3}{l}{\textbf{Environment}} &  \tabincell{l}{\textbf{Sub-} \\ \textbf{sequences} }& \tabincell{l}{\textbf{Max} \\ \textbf{frame}} & \tabincell{l}{\textbf{Mean} \\ \textbf{frame}} \\
	\midrule
	\multicolumn{1}{l}{\multirow{4}{*}{$E_{c_{1}}$}} & $c_{1}=\left \{ t:\gamma_t \le 0.28 \wedge \gamma_t \ge 2.38,t \in E_n \right \}$ & Ratio & 5,173 & 29,834 & 852 \\
	\multicolumn{1}{l}{} & $c_{2}=\left \{ t:\varsigma_t \le 0.02 \wedge \varsigma_t \ge 0.39,t \in E_n \right \}$ & Relative scale & 4,283 & 29,834 & 1,433 \\
	\multicolumn{1}{l}{} & $c_{3}=\left \{ t:\iota_t \le 0.01 \wedge \iota_t \ge 0.13,t \in E_n \right \}$ & Illumination & 4,965 & 28,828 & 856  \\
	\multicolumn{1}{l}{} & $c_{4}=\left \{ t:\beta_t \le 95,t \in E_n \right \}$ & Blur bbox & 4,353 & 28,828 & 1,439  \\
	\midrule
	\multicolumn{1}{l}{\multirow{6}{*}{$E_{c_{2}}$}} & $c_{5}=\left \{ t:\Delta \gamma_t \ge 0.2,t \in E_n \right \}$ & Delta ratio & 1,507 & 7,162 & 219 \\
	\multicolumn{1}{l}{} & $c_{6}=\left \{ t:\Delta \varsigma_t \ge 0.01,t \in E_n \right \}$ & Delta relative scale & 2,929 & 5,312 & 202 \\
	\multicolumn{1}{l}{} & $c_{7}=\left \{ t:\Delta \iota_t \ge 0.0012,t \in E_n \right \}$ & Delta illumination & 4,893 & 26,268 & 323  \\
	\multicolumn{1}{l}{} & $c_{8}=\left \{ t:\Delta \beta_t \ge 250,t \in E_n \right \}$ & Delta blur bbox & 962 & 28,800 & 523 \\
	\multicolumn{1}{l}{} & $c_{9}=\left \{ t:\varepsilon_t \ge 0.16,t \in E_n \right \}$ & Fast motion & 2,920 & 22,923 & 456 \\
	\multicolumn{1}{l}{} & $c_{10}=\left \{ t:\rho_t \le 0.75,t \in E_n \right \}$ & Corrcoef & 3,658 & 22,923 & 324 \\
	\botrule
	\end{tabular}
	\label{table:challenging}
	\end{center}
	\end{table*}

\begin{figure*}[htbp!]
	\centering
	\includegraphics[width=\textwidth]{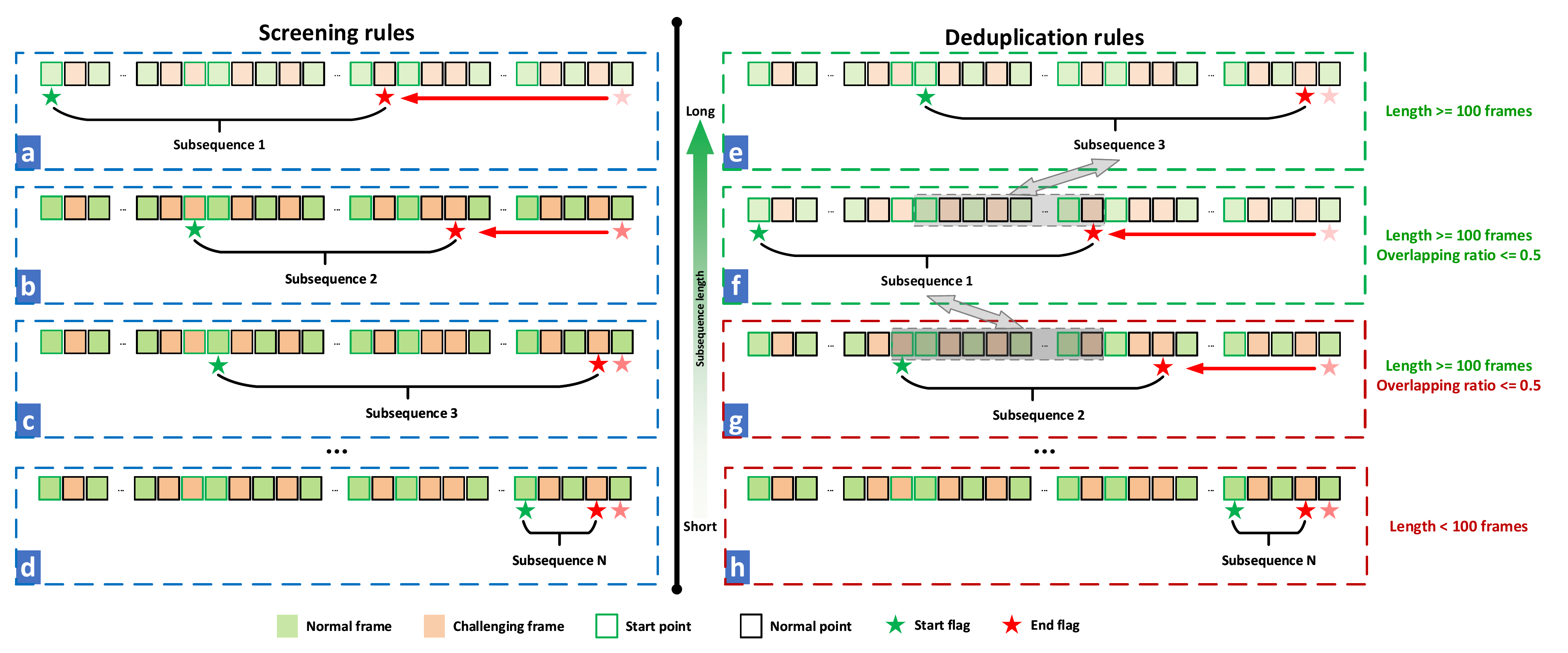}
	\caption{Schematic diagram of space construction. (a)-(d) display screening rules, aiming to find all eligible sub-sequences. (e)-(h) display the deduplication rules, aiming to remove the unqualified sub-sequences. }
	\label{fig:extract}
	\end{figure*}

\begin{algorithm}[htbp!]
	\caption{Framework of space construction.}
	\SetKwInput{KwInput}{Input}                % Set the Input
	\SetKwInput{KwOutput}{Output}              % set the Output
	\DontPrintSemicolon
	\KwInput{$L$: original sequence;
	$ \lvert \cdot \rvert $: the cardinality;
	$S$: the list of start points in $L$, sorted by frame number in ascending order;
	$C$: the list of challenging frames in $L$, sorted by frame number in ascending order}
	\KwOutput{$L_{l}$: the set of challenging sub-sequences in $L$}
	
	\tcc{Step 1: data screening}
	  
	screening set $L_{s} = \varnothing $\\
	\For{$i \leftarrow 0$ \KwTo $\lvert S \rvert-1$} 
	{
		start flag $\alpha  \leftarrow S[i]$ \\
		end flag $\beta \leftarrow \lvert L \rvert $ \\
		\While{$\beta > \alpha$}
		{
			sub-sequence $l \leftarrow L[\alpha:\beta] $\\
			\If{$\frac{\lvert l \cap C \rvert}{\lvert l \rvert} \ge 0.5$}
			{$L_{s} \leftarrow L_{s} \cup l$ \\
			\BREAK}
			\Else{$\beta \leftarrow \beta-1$}
		}
	}

	\tcc{Step 2: data deduplication}

	$L_s \leftarrow \mathrm{DescendingSort} (L_s) $ 
	  
	extraction set $L_{l} = \varnothing $\\
	\For{$i \leftarrow 0$ \KwTo $\lvert L_s \rvert -1$} 
	{
		sub-sequence $l_s \leftarrow L_s[i]$ \\
		\For{$j \leftarrow 0$ \KwTo $\lvert L_c \rvert -1$} 
		{
			sub-sequence $l_c \leftarrow L_c[j]$ \\
			\If{$\frac{ \lvert l_s \cap l_c \rvert}{\lvert l_s \rvert } \ge 0.5$}
			{\BREAK}
			\If{$(j == \lvert L_c \rvert-1) \wedge (\lvert L_s \rvert \ge 100)$}
			{$L_{l} \leftarrow L_{l} \cup l_s$ }
		}
	}
	return $L_{l}$
	\label{alg:extract}
	\end{algorithm}

Space construction rules are proposed based on intensive attribute annotation, allowing users to extract relevant data to form challenging spaces. If more than half of the frames in a sequence are challenging frames of attribute $a_{i}$, the sequence will be regarded as a \textit{challenging sequence}. The \textit{challenging sub-space} $c_{i}$ is consisted of challenging sequences of $a_{i}$.
Figure~\ref{fig:extract} and Algorithm~\ref{alg:extract} shows the process of the space construction method, including data screening and deduplication:

\begin{itemize}
	\item \textbf{Data screening} aims to find all challenging sub-sequences in an original sequence. First, we determine all appropriate start points in the original series. Initializing trackers in frames with tiny or blur targets is unreasonable, while manually selecting start points is time-consuming. Thus, we calculate the scale and clarity to exclude low-quality objects. Frames near the subsequent target absence are excluded, and the remaining are start points (green border in Figure~\ref{fig:extract}). We specify a start point as the first frame of the sub-sequence and move the end flag forward. Once the sub-sequence satisfies the challenging sequence, it will be recorded. Then change the start flag and repeat the above process until all start points are checked. The data screening rules can ensure all eligible sub-sequences are not overlooked.
	\item \textbf{Data deduplication} aims to remove the unqualified sub-sequences. Based on their length, we arrange all sub-sequences in descending order and keep the longest series as the first baseline. Other sub-sequences will be compared with all baselines and calculate the overlapping ratio. A series that has a high overlapping ratio or is less than 100 frames will be discarded; otherwise, it will be regarded as a new baseline. We keep all baselines as the extraction result of the original sequence. Eligible sub-sequences combine into the challenging sub-space $c_{i}$ of attribute $a_i$. All sub-spaces $c_i$ comprise the challenging space $E_c$, as listed in Table~\ref{table:challenging}.
\end{itemize}

\subsection{Evaluation}
\label{subsec:evaluation}

\begin{figure}[htbp!]
	\centering
	\includegraphics[width=\linewidth]{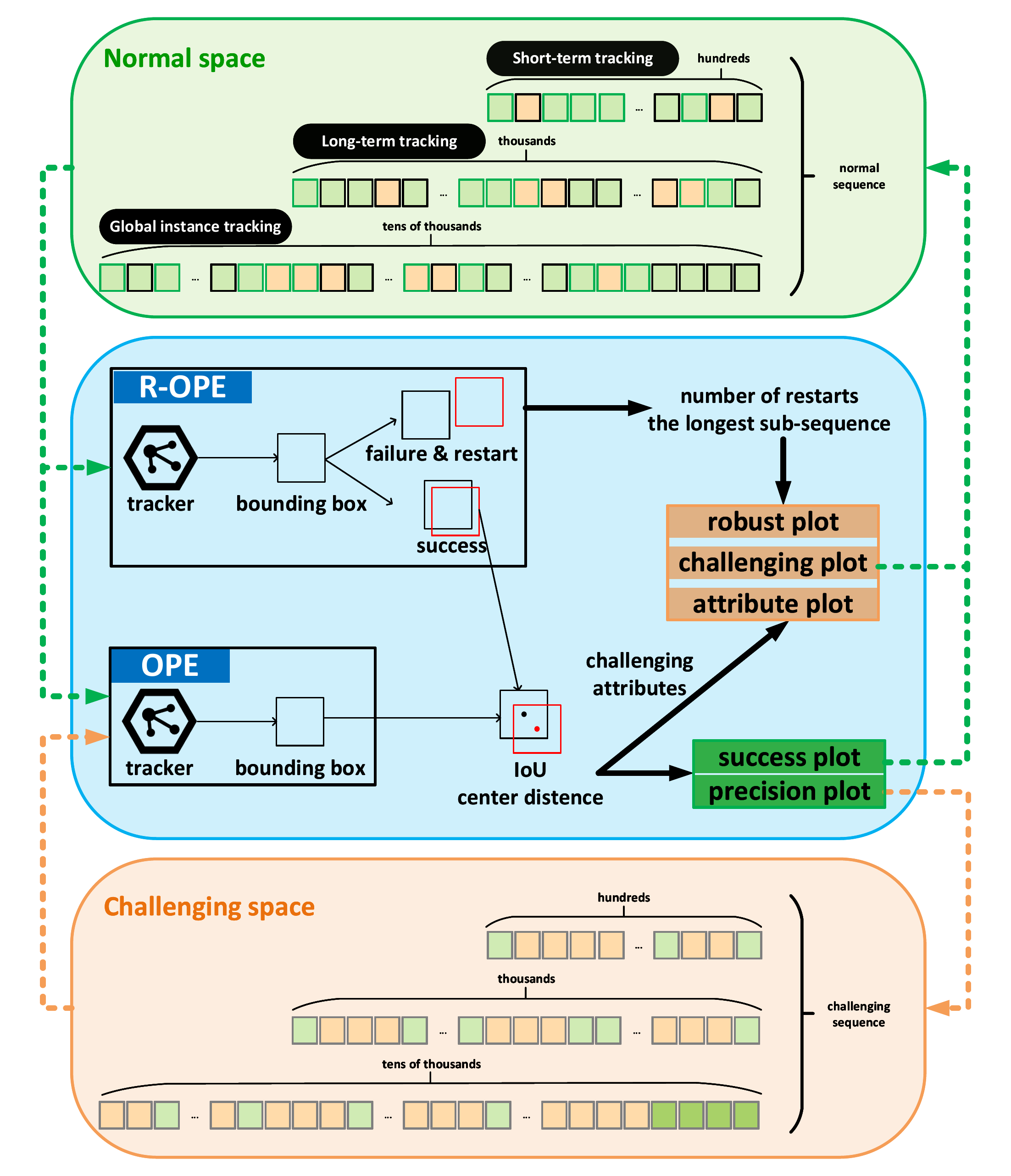}
	\caption{Evaluation process of normal space $E_n$ and challenging space $E_c$. The green dotted line and orange dotted lines respectively indicate the evaluation process of $E_n$ and $E_c$.}
	\label{fig:evaluation}
	\end{figure}

\begin{table*}[htbp!]
	\begin{center}
	\caption{The implementation of experiments, organized by 3E Paradigm. More information about $E_n$ and $E_c$ can reference Table~\ref{table:dataset} and Table~\ref{table:challenging}. Details about $M_s$ and $M_p$ are listed in Section~\ref{subsec:evaluation}.}
		\resizebox{\textwidth}{!}{
	\begin{tabular}{llll}
	\toprule
	\multicolumn{1}{l}{\multirow{2}{*}{\textbf{Task} $\mathbb{S}$}} & \multicolumn{1}{l}{\multirow{2}{*}{\textbf{Environment} $\mathbb{E}$}}& \multicolumn{1}{l}{\textbf{Evaluation} $\mathbb{M}$} & \multicolumn{1}{l}{\multirow{2}{*}{\textbf{Executor} $\mathbb{T}$}} \\
	\cline{3-3} 
	\multicolumn{1}{l}{} & \multicolumn{1}{l}{} & \multicolumn{1}{l}{System $M_s$ and Indicator $M_p$} & \multicolumn{1}{l}{}\\
	\midrule 
	$S_{n_1}$: short-term tracking & $E_{n_1} = (e_1, e_2, e_3,e_4,e_5)$ & \multirow{3}{*}{\tabincell{l}{$M_{s} = (\mathrm{OPE} , \mathrm{R\text{-}OPE})$ \\ $M_{p} = (\mathcal{P} (\cdot), \mathcal{N} (\cdot),\mathcal{S} (\cdot)$, \\ $\mathcal{C} (\cdot),\mathcal{A} (\cdot),\mathcal{R} (\cdot))$}} & \multirow{13}{*}{\tabincell{l}{$T_t$=(MixFormer \cite{MixFormer}, KYS \cite{KYS}, \\ KeepTrack \cite{KeepTrack}, Ocean \cite{Ocean}, \\ SiamRCNN \cite{SiamRCNN}, SuperDiMP \cite{PrDiMP}, \\ PrDiMP \cite{PrDiMP}, SiamCAR \cite{SiamCAR}, \\ SiamFC++ \cite{SiamFC++}, SiamDW \cite{SiamDW}, \\ GlobalTrack \cite{GlobalTrack}, DiMP \cite{DiMP}, \\ SPLT \cite{SPLT}, SiamRPN++ \cite{SiamRPN++}, \\ ATOM \cite{ATOM}, DaSiamRPN \cite{DaSiamRPN},\\ SiamRPN \cite{SiamRPN}, ECO \cite{ECO},\\ SiamFC \cite{SiamFC}, KCF \cite{KCF})}} \\
	$S_{n_2}$: long-term tracking & $E_{n_2} = (e_6, e_7)$ & \multicolumn{1}{l}{} & \multicolumn{1}{l}{} \\
	$S_{n_3}$: global instance tracking & $E_{n_3} = e_8$ & \multicolumn{1}{l}{} & \multicolumn{1}{l}{} \\
	\cline{1-3} 
	$S_{c_1}$: tracking under abnormal ratio & $E_{l} = c_1$ & \multirow{10}{*}{\tabincell{l}{$M_{s} = \mathrm{OPE}$ \\ $M_{p} = (\mathcal{P} (\cdot), \mathcal{N} (\cdot), \mathcal{S} (\cdot))$}} & \multicolumn{1}{l}{} \\
	$S_{c_2}$: tracking under abnormal scale & $E_{l} = c_2$ & \multicolumn{1}{l}{} & \multicolumn{1}{l}{} \\
	$S_{c_3}$: tracking under abnormal illumination & $E_{l} = c_3$ & \multicolumn{1}{l}{} & \multicolumn{1}{l}{}  \\
	$S_{c_4}$: tracking under blur bbox& $E_{l} = c_4$ & \multicolumn{1}{l}{} & \multicolumn{1}{l}{} \\
	$S_{c_5}$: tracking under delta ratio & $E_{l} = c_5$ & \multicolumn{1}{l}{} & \multicolumn{1}{l}{}  \\
	$S_{c_6}$: tracking under delta scale & $E_{l} = c_6$ & \multicolumn{1}{l}{} & \multicolumn{1}{l}{}\\
	$S_{c_7}$: tracking under delta illumination & $E_{l} = c_7$ & \multicolumn{1}{l}{} & \multicolumn{1}{l}{} \\
	$S_{c_8}$: tracking under delta blur bbox & $E_{l} = c_8$ & \multicolumn{1}{l}{} & \multicolumn{1}{l}{} \\
	$S_{c_9}$: tracking under fast motion & $E_{l} = c_9$ & \multicolumn{1}{l}{} & \multicolumn{1}{l}{} \\
	$S_{c_{10}}$: tracking under low correlation coefficient & $E_{l} = c_{10}$ & \multicolumn{1}{l}{} & \multicolumn{1}{l}{} \\
	\botrule
	\end{tabular}}
	\label{table:implementation}
	\end{center}
	\end{table*}

\subsubsection{Evaluation System}
SOTVerse provides two evaluation systems -- the traditional OPE system and the mechanism with re-initialization (R-OPE). Unlike OTB \cite{OTB2015} or VOT competition \cite{VOT2016,VOT2018,VOT2019} that only supports reset in the short-term tracking task, SOTVerse allows re-initialization in all subtasks (\ie, short-term tracking, long-term tracking, and global instance tracking) to maximize sequence utilization. A tracker that fails for ten consecutive frames will trigger the re-initialization mechanism and be reset at the next start point.

\subsubsection{Performance Indicator}

Suppose the evaluation environment $E$ is composed of $\lvert E \rvert$ sequences, where $\lvert \cdot \rvert$ is the cardinality. For the $t$-th frame $F_t$ in a sequence $L$, suppose that $p_t$ is the position predicted by a tracker $T$, and $g_t$ is the ground-truth. Specifically, a frame without the target is regarded as an empty set (\ie, $g_t = \phi $) and excluded by the evaluation process. Traditional \textit{precision score} and \textit{success score} of frame $F_t$ are calculated by: 

\begin{equation} \label{equ:3}
	\begin{aligned}
	d_{t} &= {\left\|c_{p}-c_{g}\right\|_{2}} \\
	  s_{t} &= \Omega (p_t,g_t) = \frac{p_t\bigcap g_t}{p_t\bigcup g_t}\\
	\end{aligned}
	\end{equation}

\noindent
where $d_{t}$ is the distance between center points $c_{p}$ and $c_{g}$, $\Omega (\cdot )$ is the intersection over union.

Recently, \textit{normalized precision score} (\cite{GIT}) is proposed to exclude the influence of target size and frame resolution.
Trackers with a predicted center outside the ground-truth will add a penalty item ${d_t}^{p}$ (\ie, the shortest distance between center point $c_{p}$ and the ground-truth edge). For trackers whose center point falls into the ground-truth, the center distance ${d_t}^{'}$ equals the original precision $d_t$ (\ie, ${d_t}^{p}=0$). 

\begin{equation} \label{equ:4}
	\begin{aligned}
        N({d_t}) &= \frac{{d_t}^{'}} {\max ( \{{d_i}^{'} \mid i \in F_t \} )} \\{d_t}^{'} &= d_t+{d_t}^{p} \\
	\end{aligned}
	\end{equation}

Obviously, the \textit{precision} $\mathcal{P} (E)$, \textit{normalized precision} $\mathcal{N} (E)$, and \textit{success} $\mathcal{S} (E)$ of environment $E$ can be defined as:

\begin{equation} \label{equ:5}
	\begin{aligned}
	\mathcal{P} (E) &= \frac{1}{\lvert E \rvert} \sum_{l=1}^{\lvert E \rvert} \frac{1}{\lvert L \rvert } \lvert \left \{ t:d_t \le \theta_{d} \right \}  \rvert\\
        \mathcal{N} (E) &= \frac{1}{\lvert E \rvert} \sum_{l=1}^{\lvert E \rvert} \frac{1}{\lvert L \rvert } \lvert \left \{ t:N({d_t}) \le \theta_{d}^{'} \right \}  \rvert\\
        \mathcal{S} (E) &= \frac{1}{\lvert E \rvert} \sum_{l=1}^{\lvert E \rvert} \frac{1}{\lvert L \rvert} \lvert \left \{ t:s_t \ge \theta_{s} \right \}  \rvert \\
	\end{aligned}
	\end{equation}

\noindent
Calculating the proportion of frames whose distance $d_t \le \theta_{d}$ and drawing the statistical results based on different $\theta_{d}$ into a curve generates the \textit{precision plot}. Typically, existing benchmarks always select $\theta_{d}=20$ to rank trackers.
Similarly, drawing statistical results based on different ${\theta_{d}}^{'} \in [0,1]$ into a curve generates the \textit{normalized precision plot}. However, directly selecting a ${\theta_{d}}^{'}$ to rank executors may introduce human factors. Thus, the proportion of frames whose predicted center $c_p$ successfully falls in the ground-truth rectangle $g_t$ are selected to rank trackers. 
Frames with overlap $s_t \ge \theta_{s}$ are defined as successful tracking. 
Draw the results based on various overlap threshold $\theta_{s}$ into a curve is the \textit{success plot}, where the mAO (mean average overlap) is wildly used to rank trackers.

Evidently, traditional precision plot, normalized precision plot, and success plot average scores on complete series to generate the final result. 
Challenging space that already contains enough \textit{challenging frames} can directly use these indicators. But for normal sequences composed of most \textit{normal frames} and a few \textit{challenging frames}, the above metrics may ignore the influence of challenging factors due to the average calculation. Thus, SOTVerse provides three novel indicators to concentrate on the impact of challenges:

\begin{itemize}
	\item \textbf{Challenging plot.} For the $t$-th frame $F_t$ in $L$, suppose that $\rho _t$ is the correlation coefficient between $F_t$ and $F_{t-1}$. A frame with $s_t \ge 0.5$ is defined as \textit{success frame}, and vice versa is \textit{fail frame}. The \textit{challenging score}is defined as:
	\begin{equation} \label{equ:6}
		\begin{aligned}
		\mathcal{C} (E) &= \frac{1}{\lvert E \rvert} \sum_{l=1}^{\lvert E \rvert}\frac{\lvert \left \{ t:s_t \ge 0.5 \right \}  \rvert }{\lvert \left \{ t:\rho _t \le \theta_{\rho} \right \}  \rvert  }\\
		\end{aligned}
		\end{equation}
	\noindent
	Calculating the proportion of success frames on the challenging part (\ie, $\rho _t \le \theta_{\rho}$) and drawing the statistical results based on different $\theta_{\rho}$ into a curve generates the \textit{challenging plot}. SOTVerse selects $\theta_{\rho}=0.75$ to rank trackers.
	\item \textbf{Attribute plot.} \textit{Attribute plot} $\mathcal{A} (\cdot) $ aims to find the attribute that affects tracking most. SOTVerse finds all fail frames and determines the attribute with the highest ratio is the main failure reason. Unlike other indicators to rank algorithms, the attribute plot intuitively reveals the most likely reasons causing failures for each tracker.
	\item \textbf{Robust plot.} The \textit{robust plot} $\mathcal{R} (\cdot) $ aims to exhibit the performance of trackers in the R-OPE mechanism. SOTVerse counts the number of restarts for each video, divides the entire video into several segments based on the restart point, and returns the longest sub-sequence that the algorithm successfully runs. Taking the number of restarts and the mean value of the longest sub-sequence as abscissa and ordinate can generate a robust plot. Trackers closer to the upper left corner have better performance (indicating successful tracking in longer sequences with rare re-initializations).
	\end{itemize}

\begin{figure*}[htbp!]
	\centering
	\includegraphics[width=\textwidth]{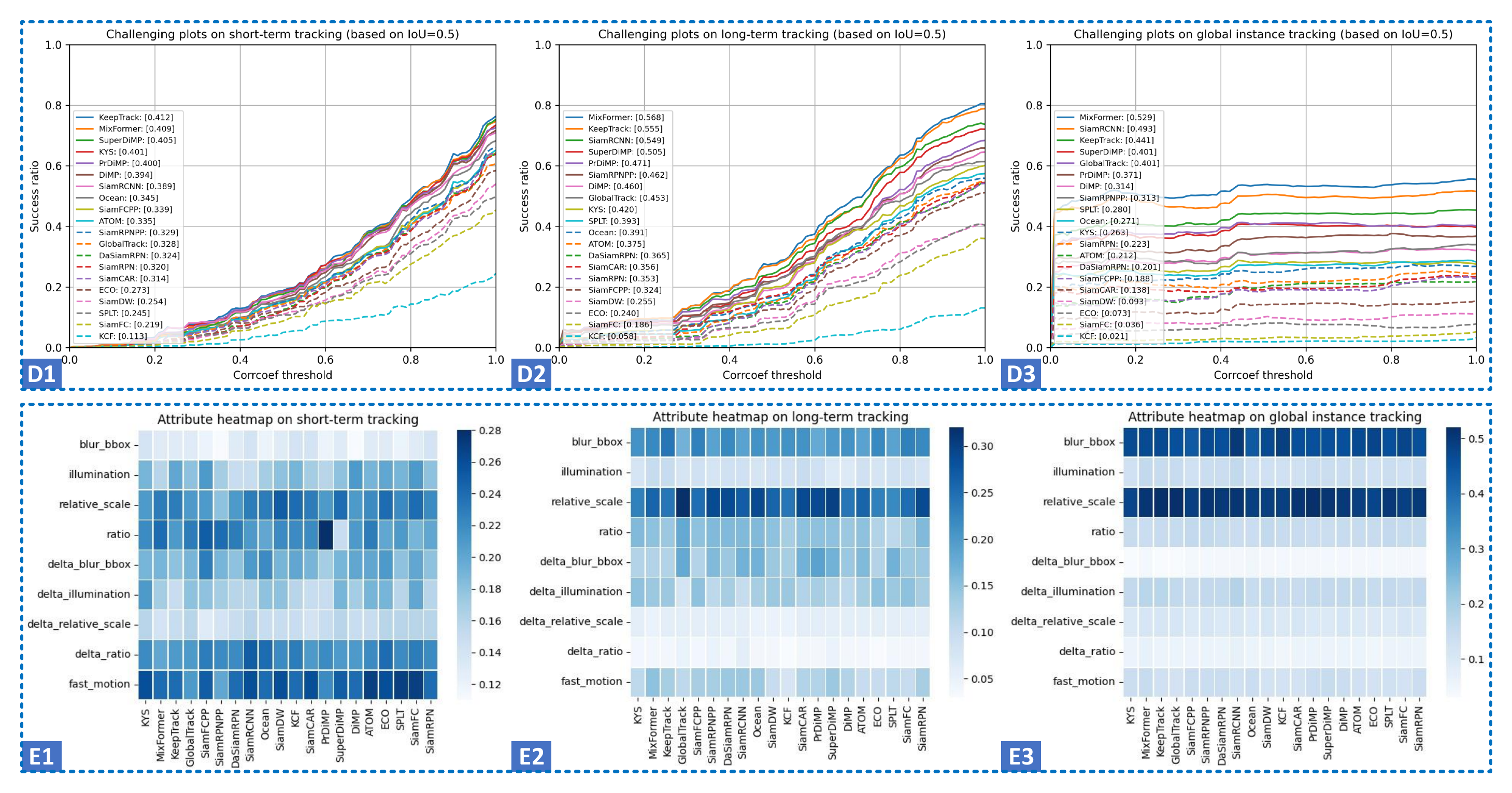}
	\caption{Experiments in normal space with OPE mechanisms. Three columns represent the results in the short-term tracking task (left), long-term tracking task (middle), and global instance tracking task (right). Each task is evaluated by challenging plots (D1-D3) and attribute plots (E1-E3).}
	\label{fig:ope-challenge}
	\end{figure*}

\begin{figure*}[htbp!]
	\centering 
	\includegraphics[width=\textwidth]{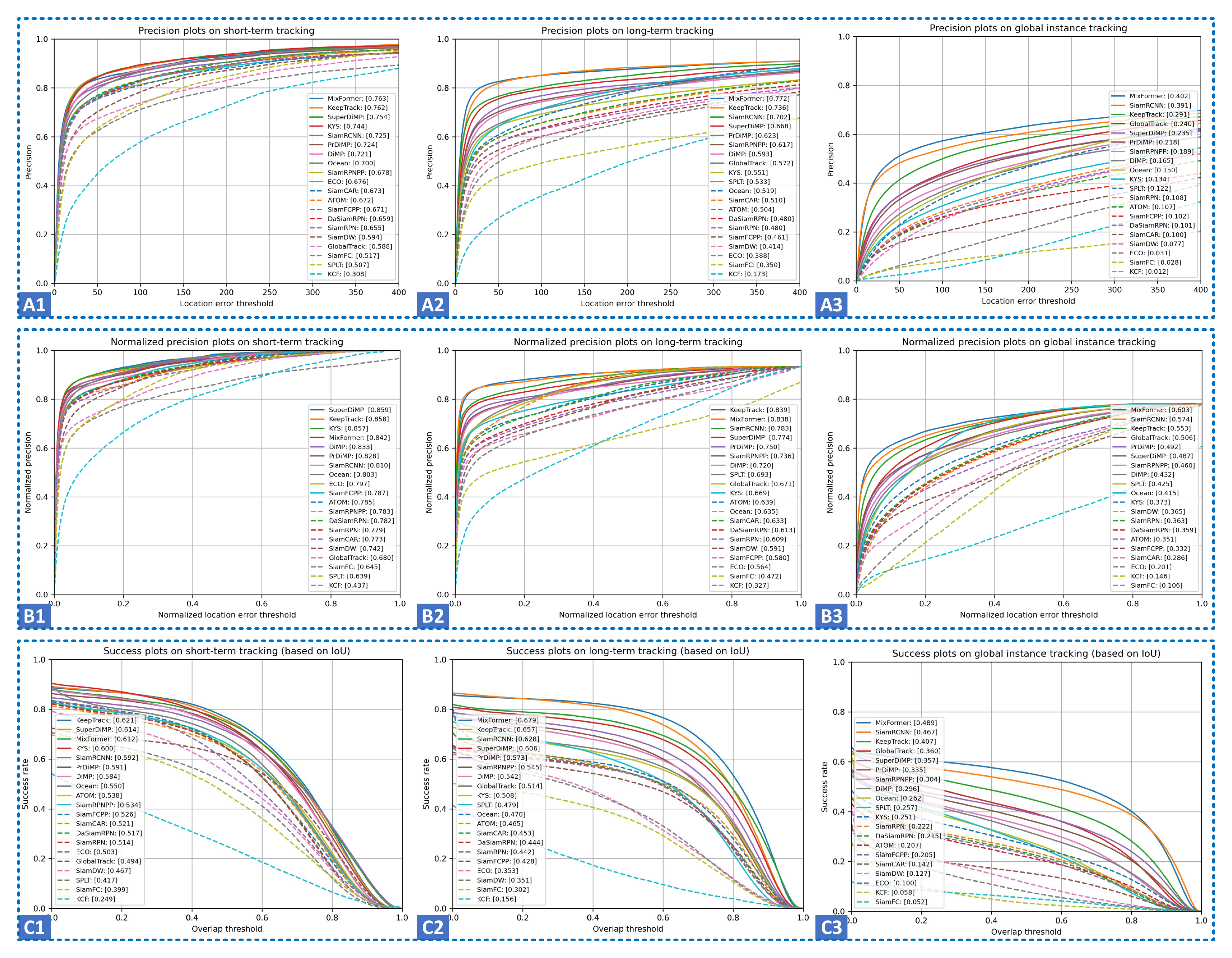}
	\caption{Experiments in normal space with OPE mechanisms. Three columns represent the results in the short-term tracking task (left), long-term tracking task (middle), and global instance tracking task (right). Each task is evaluated by precision plots (A1-A3), normalized precision plots (B1-B3), and success plots (C1-C3).}
	\label{fig:ope-normal}
	\end{figure*}

\begin{figure*}[htbp!]
	\centering 
	\includegraphics[width=\textwidth]{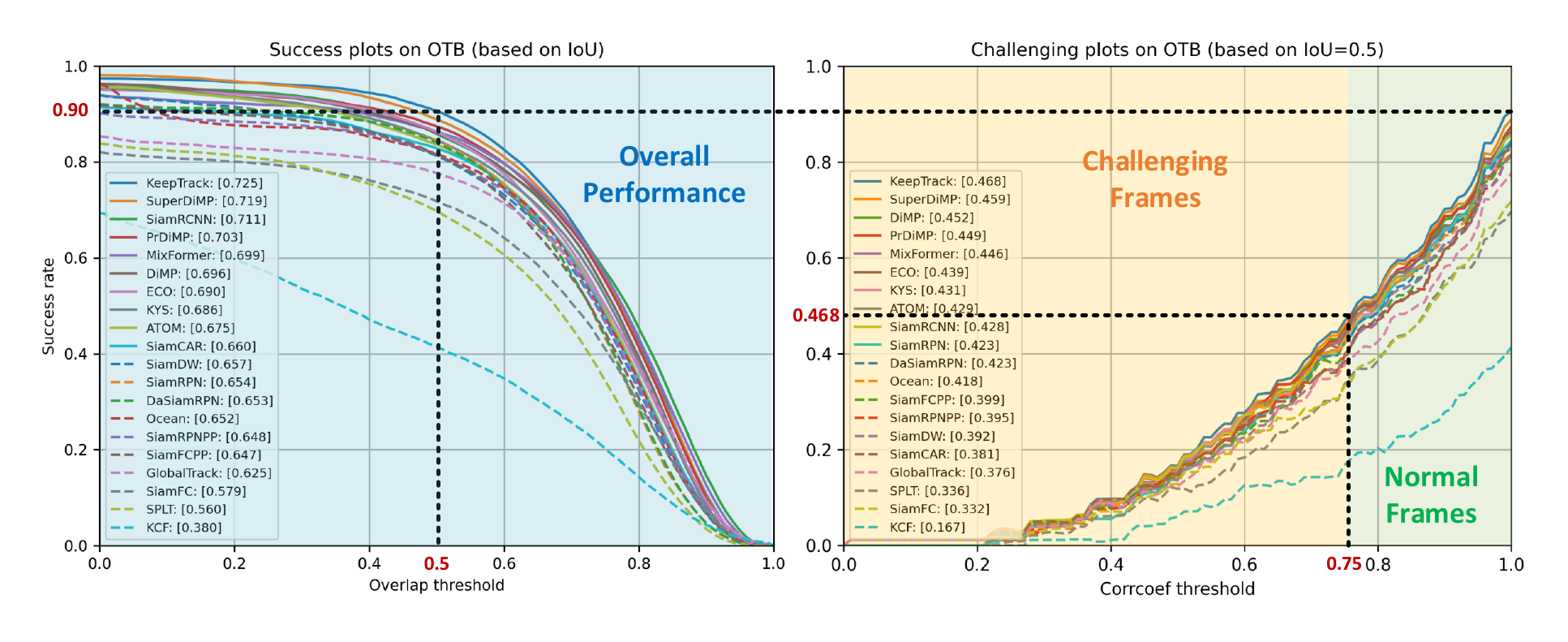}
	\caption{Comparison of performance under success plot (left) and challenging plot (right), taking OTB \cite{OTB2015} as an example.}
	\label{fig:motivation}
	\end{figure*}

\section{Experiments}
\label{sec:experiments}
\subsection{Implementation Details}

We select 20 represent algorithms as task executor $T_t$ and conduct experiments based on 3E Paradigm, as shown in Table~\ref{table:implementation}. 
The 20 trackers can be divided into four categories based on their model architectures. (1) Correlation filter (CF) based trackers: KCF \cite{KCF} and ECO \cite{ECO}. (2) Siamese neural network (SNN) based trackers: SiamFC \cite{SiamFC}, SiamRPN \cite{SiamRPN}, DaSiamRPN \cite{DaSiamRPN}, SiamRPN++ \cite{SiamRPN++}, SPLT \cite{SPLT}, SiamDW \cite{SiamDW}, SiamCAR \cite{SiamCAR}, SiamFC++ \cite{SiamFC++}, Ocean \cite{Ocean}, and SiamRCNN \cite{SiamRCNN}. (3) Trackers that combine CF and SNN: ATOM \cite{ATOM}, DiMP \cite{DiMP}, PrDiMP \cite{PrDiMP}, SuperDiMP \cite{PrDiMP}, and KeepTrack \cite{KeepTrack}. (4) Trackers with custom networks, like 
GlobalTrack \cite{GlobalTrack} with zero cumulative error, KYS \cite{KYS} with scene information, and MixFormer \cite{MixFormer} with an end-to-end transformer-based structure.

Please refer to Section~\ref{supsec:overall} of the appendices for detailed information and comprehensive experimental results for the 20 representative trackers. 

\subsection{Experiments in Normal Space}

\begin{figure*}[htbp!]
	\centering 
	\includegraphics[width=\textwidth]{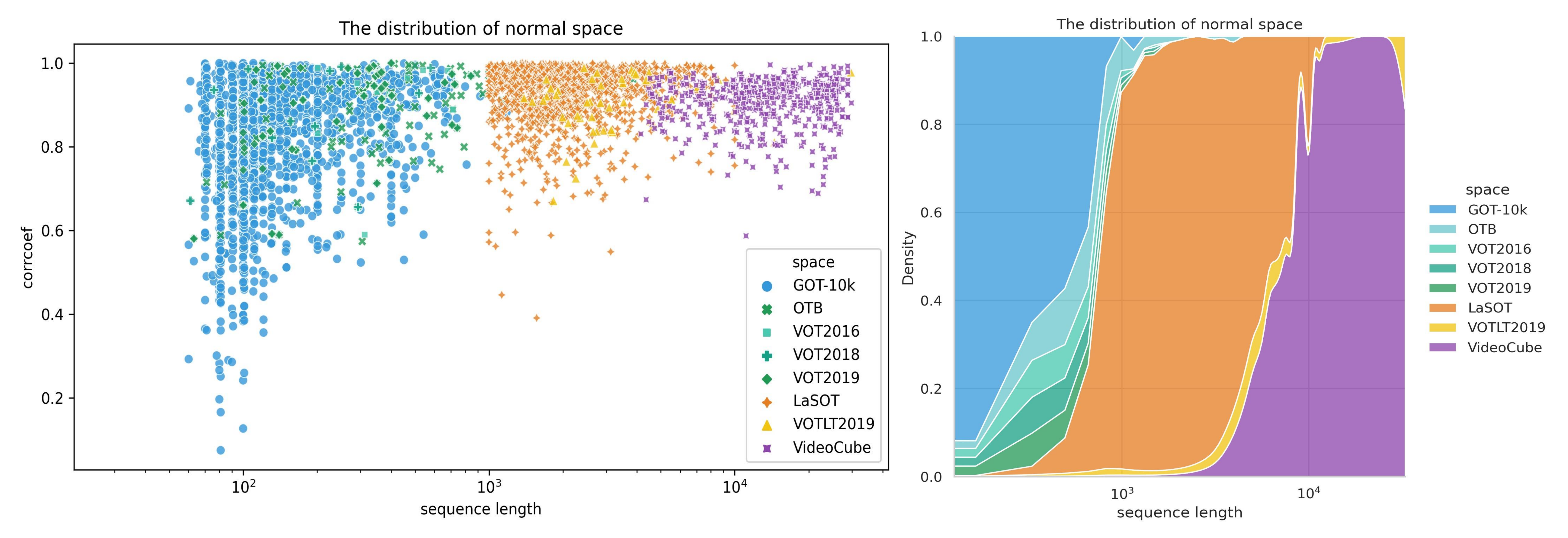}
	\caption{The composition of normal space. (left) The distribution of corrcoef values and sequence lengths, each point representing a sub-sequence. (right) The distribution of sequence lengths. }
	\label{fig:distribution-normal}
	\end{figure*}

\begin{figure*}[htbp!]
	\centering 
	\includegraphics[width=\textwidth]{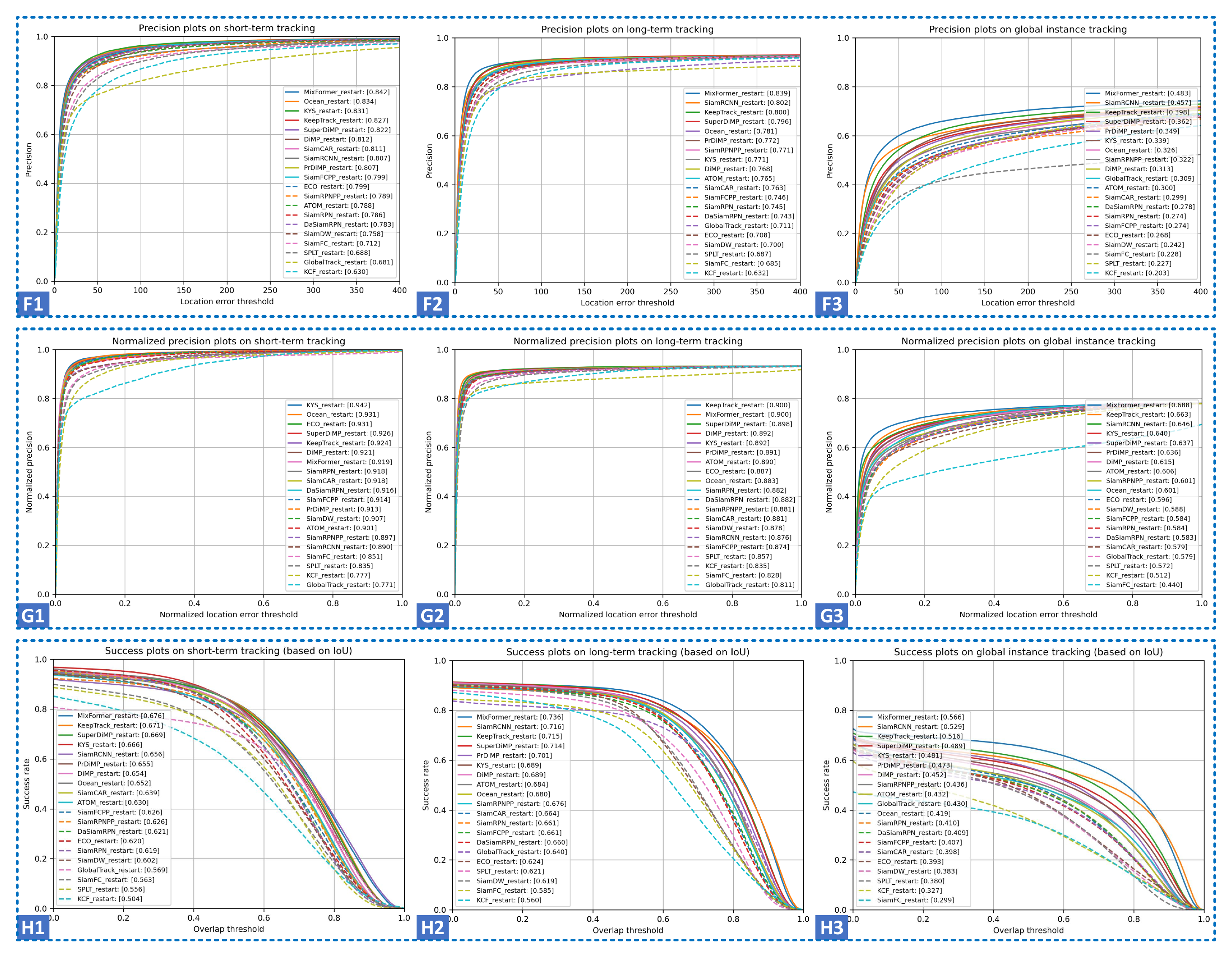}
	\caption{Experiments in normal space with R-OPE mechanisms. Three columns represent the results in the short-term tracking task (left), long-term tracking task (middle), and global instance tracking task (right). Each task is evaluated by precision plots (F1-F3), normalized precision plots (G1-G3), and success plots (H1-H3).}
	\label{fig:rope-normal}
	\end{figure*}

\begin{figure*}[htbp!]
	\centering 
	\includegraphics[width=\textwidth]{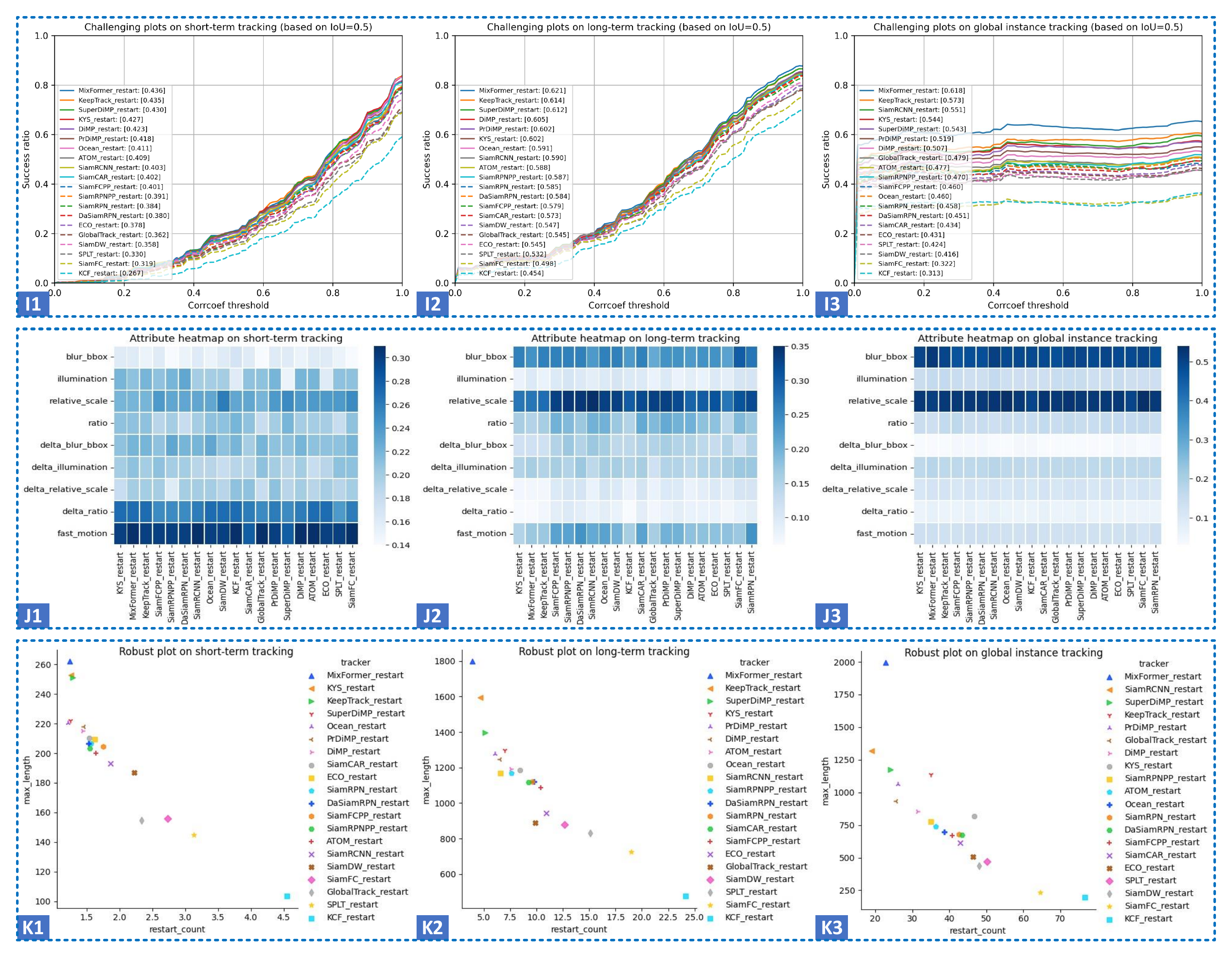}
	\caption{Experiments in normal space with R-OPE mechanisms. Three columns represent the results in the short-term tracking task (left), long-term tracking task (middle), and global instance tracking task (right). Each task is evaluated by challenging plots (I1-I3), attribute plots (J1-J3), and robust plots (K1-K3).}
	\label{fig:rope-challenge}
	\end{figure*}

Figure~\ref{fig:ope-normal} to Figure~\ref{fig:rope-challenge} illustrate the experimental results in normal space. We first conduct experiments on the meta-datasets $e_i$ for each subtask $S_{n_i}$, then average results as the final performance for the current subtask. We add a figure number at the bottom left for each subplot to better illustrate the experimental results.

\subsubsection{Experiments in OPE mechanism.}

Figure~\ref{fig:ope-normal} to Figure~\ref{fig:ope-challenge} show the performance of trackers under the OPE mechanism.

\vspace{0.5em} \noindent
\textbf{\textit{Influence of Task Constraints on Tracking Performance.}}

From the task perspective, Figure~\ref{fig:ope-normal} shows a downward trend in trackers' performance from short-term tracking to global instance tracking. This phenomenon indicates that with the relaxation of task constraints, more challenging factors are occurred and require higher tracking ability.
Especially compared with the first two tasks, performance drops the most on global instance tracking, which indicates that as the SOT task that is closest to the actual application scenario, global instance tracking is still a considerable difficulty to existing methods.

\vspace{0.5em} \noindent
\textbf{\textit{Limitations of Existing Evaluation Metrics.}}

Before conducting analyses based on new metrics, we first illustrate the limitations of existing metrics through an experiment in Figure~\ref{fig:motivation}. Existing benchmarks only evaluate complete sequences but ignore the challenging frames. Taking the OTB \cite{OTB2015} as an example, the traditional success plot (left) for KeepTrack \cite{KeepTrack} indicates it successfully tracks 90\% frames, while the challenging plot (right) proposed in this paper shows that the success rate of challenging frames is only 46.8\%. Obviously, the existing evaluation system ignores the influence of challenging factors. 

\vspace{0.5em} \noindent
\textbf{\textit{Tracking Evaluation via Challenging Plots.}}

Figure~\ref{fig:ope-challenge} shows the performance of the challenging plot and attribute plot. The challenging plots (D1-D3) demonstrate that the algorithm's success rate on challenging frames is basically lower than 50\%.

Observe three points in challenging plots (D1-D3): the inflection point, the point with $\theta_{\rho}=0.75$, and the endpoint on the right ($\theta_{\rho}=1$). Attention that the variation trends of challenging plots are not totally similar, meaning the decisive factors for influencing algorithm performance of various tasks are different:

\begin{itemize}
	\item \textbf{Challenging factors mainly influence short-term tracking task.} For short-term tracking (D1), the success rate of most algorithms increases with the improvement of the correlation coefficient, indicating that challenging factors are the main element affecting the success rate.
	\item \textbf{Challenging factors and sequence length mainly influence long-term tracking task.} For long-term tracking (D2), the performance increases with the addition of the correlation coefficient in the challenging frame area; but when the sequence contains more normal frames, the improvement of performance slows down, revealing that sequence length becomes the main factor affecting the success rate.
	\item \textbf{Shot-switching mainly influences global instance tracking task.} For the global instance tracking task (D3) that includes shot switching, the curve's inflection point occurs when the correlation coefficient is very low, and the slope of the curve to the right of the inflection point is gentle. Thus, the decisive factor for algorithm performance in the GIT task is not challenging factors, but whether the target position can be re-located after each shot switching. 
\end{itemize}

\vspace{0.5em} \noindent
\textbf{\textit{Tracking Evaluation via Attribute Plots.}}

For attribute plots (E1-E3), in the static attributes, the object scale has a greater impact on tracking performance; in the dynamic attributes, the fast motion has a more significant effect on success rates.

\begin{figure*}[htbp!]
	\centering 
	\includegraphics[width=\textwidth]{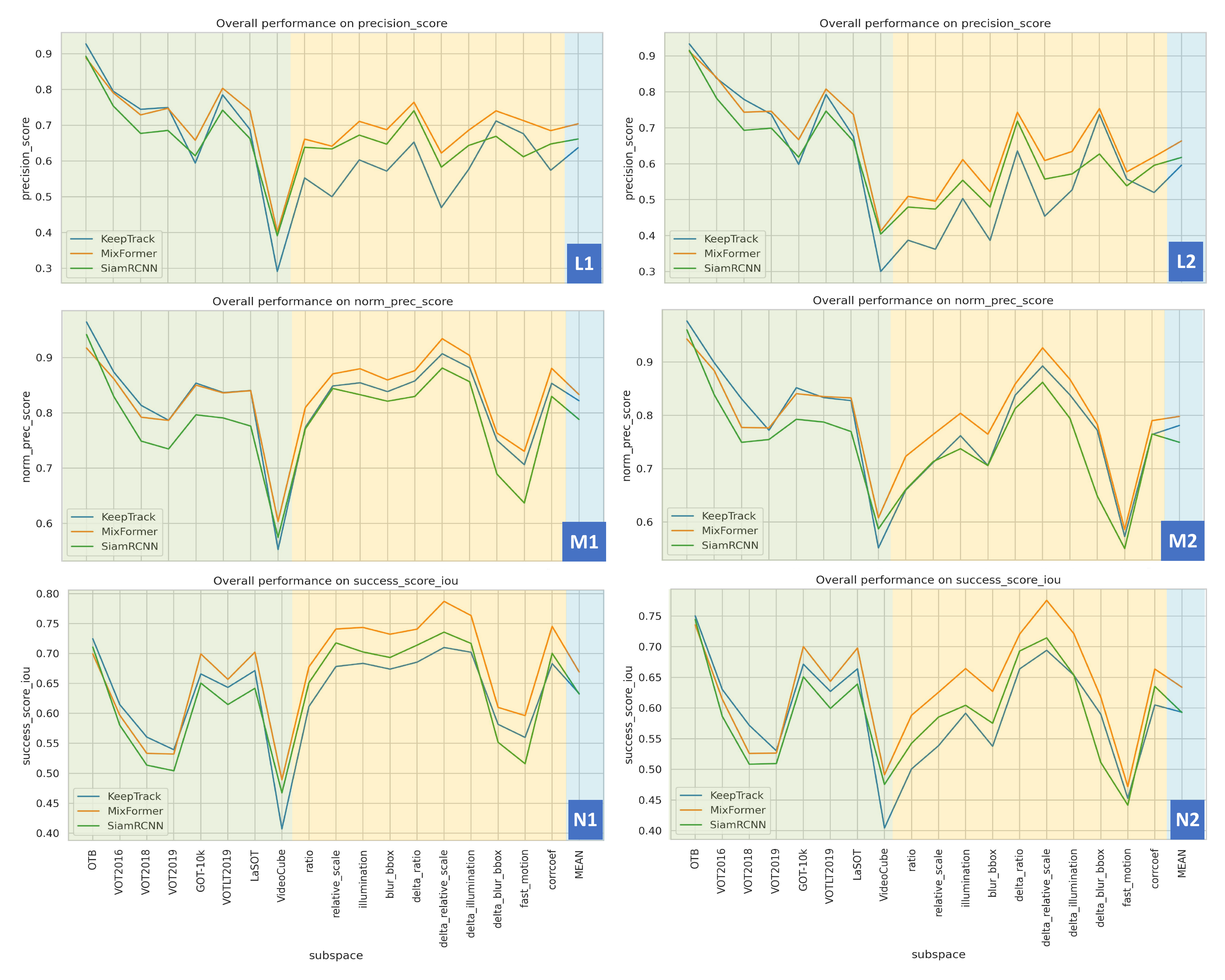}
	\caption{Experiments in all sub-spaces with OPE mechanism, represented by KeepTrack \cite{KeepTrack}, MixFormer \cite{MixFormer} and SiamRCNN \cite{SiamRCNN}. The Green and orange backgrounds represent the performance in normal space and challenge space respectively. The blue background represents the average performance over all sub-spaces. Results are calculated by precision plots (L1-L2), normalized precision plots (M1-M2), and success plots (N1-N2). The left column (L1, M1, N1) is generated by original indicators, while the right column (L2, M2, N2) is weighted by sequences' length.}
	\label{fig:overall-exp}
	\end{figure*}

\subsubsection{Experiments in R-OPE mechanism.}

\vspace{0.5em} \noindent
\textbf{\textit{Comparison of OPE and R-OPE Mechanisms.}}

Figure~\ref{fig:rope-normal} to Figure~\ref{fig:rope-challenge} show the performance of trackers under the R-OPE mechanism.
The R-OPE mechanism allows re-initialization after failure and avoids the wastage of subsequent sequences. Thus, the precision score, normalized precision score, and success score in Figure~\ref{fig:rope-normal} are higher than OPE mechanism (Figure~\ref{fig:ope-normal}). 
But the success score in challenging plots (I1-I3) does not improve significantly, indicating that the improvement in the other three indicators mainly comes from successfully tracking in normal frames rather than challenging frames.

\vspace{0.5em} \noindent
\textbf{\textit{Tracking Evaluation via Robust Plots.}}

The robust plots (K1-K3) visually display the relationship between the longest successfully tracked sub-sequence and the restart times. From these plots, we can summarize the necessity of the R-OPE mechanism for different tracking tasks. The following conclusions also indicate the irrationality of the existing evaluation mechanism:

\begin{itemize}
	\item \textbf{Re-initialization is optional for short-term tracking task.} Most SOTA algorithms can complete tracking with only one restart in the short-term tracking. Furthermore, for GOT-10k \cite{GOT-10k} with a shorter average length, trackers can track the entire video without re-initializations. The rare restart times and slight score differences under the OPE and R-OPE mechanisms illustrate that the restart mechanism is unnecessary for the short-term tracking task.
	\item \textbf{Re-initialization is necessary for long-term tracking and global instance tracking task.} Conversely, the restart mechanism is necessary for long-term tracking and global instance tracking. K2 in Figure~\ref{fig:rope-challenge} demonstrates that most excellent algorithms can continuously track 1000 to 1800 frames and then fail due to the influence of challenging attributes. The longest successfully tracked sub-sequence length is still far from the length interval (1k to 10k frames) of the long-term tracking task represented in Figure~\ref{fig:distribution-normal}.
	The experimental results reveal existing short-term benchmarks (OTB \cite{OTB2015}, VOT series \cite{VOT2013, VOT2014, VOT2015, VOT2016}) allow restarting, while the long-term tracking benchmarks (VOTLT2019 \cite{VOT2019}, LaSOT \cite{LaSOT}) do not design a restart mechanism is unreasonable.
	K3 in Figure~\ref{fig:rope-challenge} shows that even if the algorithm restarts dozens of times, it is still difficult to re-locate the object quickly when the shot switching occurs again in the global instance tracking task.
\end{itemize}

Please refer to Sections~\ref{supsec:st},~\ref{supsec:lt}, and~\ref{supsec:git} of the appendices for the initial experimental results under each benchmark.

\subsection{Experiments in Challenging Space}

We conduct experiments in ten challenge attribute spaces. Three evaluation indicators (precision plot, normalized precision plot, and success plot) are chosen to evaluate performance under OPE mechanism. 

To better compare the performance changes under different sub-spaces, we plot the algorithm scores in Figure~\ref{fig:overall-exp}. 
Since the sequence length is an essential factor affecting the results, while the sequence lengths in challenging spaces are quite diverse, we recalculate the original results (L1, M1, N1) by using sequence length as the weight and generate the weighted result (L2, M2, N2).
Here, we select the top-3 trackers (KeepTrack \cite{KeepTrack}, MixFormer \cite{MixFormer}, and SiamRCNN \cite{SiamRCNN}) as representative models to exhibit the results. For detailed results of the 20 trackers, please refer to Table~\ref{tab:pre} to~\ref{tab:success} (based on original indicators) and Table~\ref{tab:pre-l} to~\ref{tab:success-l} for weighted results.

\vspace{0.5em} \noindent
\textbf{\textit{Influence of Task Constraints.}}

Obviously, all trackers have the worst performance on VideoCube\cite{GIT}, indicating that the GIT task is challenging for even SOTA algorithms. 
Besides, compared to the challenging factors that have widely existed in each subtask, new challenges included by broadening the task constraints (\ie, relocating the target in shot-switching) have a more significant influence on tracking performance. It also indicates that the correct modeling task is vital in domain development.

\vspace{0.5em} \noindent
\textbf{\textit{Influence of Challengeing Factors.}}

Most algorithms perform poorly in challenging spaces (orange area in Figure~\ref{fig:overall-exp}). However, some trackers score relatively high on initial results (L1, M1, N1) -- one possible reason is multiple challenging spaces contain many short sub-sequences extracted from GOT-10k \cite{GOT-10k}. The weighted results decrease (L2, M2, N3) indicates that most challenging factors (\eg, fast motion, abnormal ratio, and abnormal scale) significantly impact the tracking performance; existing algorithms are still required to enhance the tracking robustness under challenging situations.

We also note that the top-3 trackers are based on different model architectures. 
MixFormer \cite{MixFormer} is an end-to-end transformer-based tracker. It performs well in most sub-spaces than other trackers, indicating that the mixed attention module and a straightforward detection head can provide powerful tracking ability. 
The SiamRCNN \cite{SiamRCNN} selects the siamese network as the basic model structure and combines a two-stage scheme with a new trajectory-based dynamic planning algorithm. Besides, the  re-detection mechanism in SiamRCNN helps it to accomplish stable tracking in the abruption of appearance or motion information.
KeepTrack \cite{KeepTrack} model is improved by the DiMP series (\cite{ATOM,DiMP,PrDiMP}), and has an enhanced ability to discriminate interferers in challenging situations. 
The combination of different sub-tasks in SOTVerse can help us comprehensively analyze various executors, rather than only comparing the performance of the algorithms on a single dataset with shallow conclusions.

Section~\ref{supsec:composition-cs} of the appendices shows the detailed distribution of challenge sub-spaces, and Section~\ref{supsec:experiment-cs} shows the specific experimental results. Please refer to Sections~\ref{supsec:composition-cs} and~\ref{supsec:experiment-cs} of the appendices for the initial experimental results.

\section{Conclusion}
\label{sec:conclusion}

This paper first proposes a 3E Paradigm to describe computer vision tasks by three components (\ie, environment, evaluation, and executor), then construct representative benchmarks as SOTVerse with 12.56 million frames. SOTVerse contains a comprehensive and user-defined environment and a thoroughgoing evaluation scheme, allowing users to customize tasks according to research purposes. We also conduct extensive experiments in the SOTVerse and conduct a performance analysis on various executors. 

In future work, we will continue to maintain the platform to support users to continuously enrich the content of the SOTVerse through \textit{interaction} and \textit{expansion} functions, and create new sub-spaces according to their respective research purposes. Additionally, we encourage users to expand into new visual or other domain tasks and create new metaverse spaces following our defined paradigm. We welcome researchers to join our platform with their newly created metaverse spaces accepted by the research community, and together promote the promotion of the metaverse paradigm to form broader research outcomes.

\section*{Declarations}

\begin{itemize}
% \item \textbf{Funding.}
\item \textbf{Conflict of Interest.} All authors declare no conflicts of interest.
% \item \textbf{Ethics approval.} 
% \item \textbf{Consent to participate.}
% \item \textbf{Consent for publication.}
\item \textbf{Availability of data and materials.} All data will be made available on reasonable request.
\item \textbf{Code availability.} The toolkit and experimental results will be made publicly available. 
% \item \textbf{Authors' contributions.}
\end{itemize}

\bibliography{sn-bibliography}

\begin{appendices}
\clearpage
\onecolumn

\section{Comprehensive Experimental results}
\label{supsec:overall}

\begin{table}[h!]
	\begin{center}
	\caption{The model architectures and URLs of open-sourced algorithms used in this work.}
	\small
	\renewcommand{\arraystretch}{1.2}
	\begin{tabular}{lll}
	\hline
	\textbf{Tracker} & \textbf{Architecture} & {\textbf{URL}} \\ \hline
	KCF \cite{KCF}   & CF & https://github.com/uoip/KCFpy                                    \\
	ECO  \cite{ECO}  & CF+CNN & https://github.com/visionml/pytracking                           \\ \hline
	SiamFC \cite{SiamFC} & SNN        & https://github.com/huanglianghua/siamfc-pytorch                  \\
	SiamRPN \cite{SiamRPN}  & SNN        & https://github.com/huanglianghua/siamrpn-pytorch                 \\
	DaSiamRPN  \cite{DaSiamRPN}  & SNN     & https://github.com/foolwood/DaSiamRPN                            \\ 
	SiamRPN++  \cite{SiamRPN++}  & SNN     & https://github.com/PengBoXiangShang/SiamRPN\_plus\_plus\_PyTorch \\
	SPLT  \cite{SPLT} & SNN           & https://github.com/iiau-tracker/SPLT                             \\SiamDW  \cite{SiamDW}   & SNN       & https://github.com/researchmm/TracKit                            \\
	SiamCAR  \cite{SiamCAR} & SNN        & https://github.com/ohhhyeahhh/SiamCAR                            \\
	SiamFC++  \cite{SiamFC++}  & SNN      & https://github.com/MegviiDetection/video\_analyst                \\
	Ocean   \cite{Ocean}  & SNN        & https://github.com/researchmm/TracKit                            \\
	SiamRCNN   \cite{SiamRCNN} & SNN      & https://github.com/VisualComputingInstitute/SiamR-CNN            \\ \hline
	ATOM \cite{ATOM}  & SNN+CF           & https://github.com/visionml/pytracking                           \\
	DiMP \cite{DiMP}   & SNN+CF           & https://github.com/visionml/pytracking                           \\
	PrDiMP  \cite{PrDiMP}  & SNN+CF        & https://github.com/visionml/pytracking                           \\
	SuperDiMP  \cite{PrDiMP}  & SNN+CF      & https://github.com/visionml/pytracking                           \\
	KeepTrack    \cite{KeepTrack} & SNN+CF     & https://github.com/visionml/pytracking                           \\ \hline
	KYS    \cite{KYS}  & Custom networks          & https://github.com/visionml/pytracking                           \\
	GlobalTrack \cite{GlobalTrack}   & Custom networks     & https://github.com/huanglianghua/GlobalTrack                     \\
	MixFormer  \cite{MixFormer}   & Custom networks      & https://github.com/MCG-NJU/MixFormer                             \\ \hline
	\end{tabular}
	\begin{tablenotes}
		\item \textit{Note: CF-Correlation Filter. CNN-Convolutional Neural Network. SNN-Siamese Neural Network.}
		\end{tablenotes}
\end{center}
\label{table:machines}
	\end{table}

All experiments are performed on a server with 4 NVIDIA TITAN RTX GPUs and a 64 Intel(R) Xeon(R) Gold 5218 CPU @ 2.30GHz. We use the parameters provided by the original authors. 

\clearpage
\begin{sidewaystable*}[htbp!]
	\sidewaystablefn%
	\begin{center}
	\caption{Performance of 20 representative trackers on all sub-spaces, based on precision score. 
	}
	\renewcommand{\arraystretch}{1.2}
	\begin{center}
		\tabcolsep=0.1cm
		\small
		\begin{tabular}{llllllllllllllllllllll}
			\toprule
			\multirow{2}{*}{\textbf{Trackers}} & \multicolumn{8}{l}{\textbf{Normal space}} &  & \multicolumn{10}{l}{\textbf{Challenging space}} & & \multirow{2}{*}{\textbf{Mean}} \\ \cline{2-9} \cline{11-20}
				& $e_1$ & $e_2$ & $e_3$ & $e_4$ & $e_5$ & $e_6$ & $e_7$ & $e_8$ &  & $c_1$ & $c_2$ & $c_3$ & $c_4$ & $c_5$ & $c_6$ & $c_7$ & $c_8$ & $c_9$ & $c_{10}$ &  \\ \midrule
				KCF & \cellcolor[HTML]{FDEB84}0.510 & \cellcolor[HTML]{FDC87D}0.374 & \cellcolor[HTML]{FBB179}0.287 & \cellcolor[HTML]{FBA175}0.226 & \cellcolor[HTML]{F98B71}0.141 & \cellcolor[HTML]{FA9673}0.185 & \cellcolor[HTML]{FA9072}0.160 & \cellcolor[HTML]{F8696B}0.012 &  & \cellcolor[HTML]{FA9E75}0.213 & \cellcolor[HTML]{FA9072}0.161 & \cellcolor[HTML]{FA9F75}0.218 & \cellcolor[HTML]{FA9473}0.176 & \cellcolor[HTML]{FBA376}0.234 & \cellcolor[HTML]{F98770}0.128 & \cellcolor[HTML]{FA9774}0.189 & \cellcolor[HTML]{FCC27C}0.349 & \cellcolor[HTML]{FA9773}0.189 & \cellcolor[HTML]{FA9773}0.187 &  & \cellcolor[HTML]{FA9F75}0.219 \\
				ECO & \cellcolor[HTML]{65BF7C}0.922 & \cellcolor[HTML]{A6D27F}0.748 & \cellcolor[HTML]{B5D680}0.706 & \cellcolor[HTML]{C5DB81}0.661 & \cellcolor[HTML]{FCC17B}0.344 & \cellcolor[HTML]{FDD17F}0.407 & \cellcolor[HTML]{FDC77D}0.370 & \cellcolor[HTML]{F86E6B}0.031 &  & \cellcolor[HTML]{FCC37C}0.354 & \cellcolor[HTML]{FBA977}0.255 & \cellcolor[HTML]{FDCE7E}0.397 & \cellcolor[HTML]{FCBA7A}0.319 & \cellcolor[HTML]{FEE282}0.471 & \cellcolor[HTML]{FBA276}0.229 & \cellcolor[HTML]{FCC27C}0.350 & \cellcolor[HTML]{DCE182}0.599 & \cellcolor[HTML]{FDEB84}0.509 & \cellcolor[HTML]{FCC57C}0.361 &  & \cellcolor[HTML]{FEDC81}0.446 \\\hline
				SiamFC & \cellcolor[HTML]{A5D17F}0.750 & \cellcolor[HTML]{E2E383}0.583 & \cellcolor[HTML]{FBEA84}0.515 & \cellcolor[HTML]{FEE482}0.479 & \cellcolor[HTML]{FBAA77}0.259 & \cellcolor[HTML]{FDC97D}0.375 & \cellcolor[HTML]{FCBC7A}0.325 & \cellcolor[HTML]{F86D6B}0.028 &  & \cellcolor[HTML]{FBAA77}0.259 & \cellcolor[HTML]{FA9D75}0.212 & \cellcolor[HTML]{FCBA7A}0.319 & \cellcolor[HTML]{FBAA77}0.258 & \cellcolor[HTML]{FCBF7B}0.337 & \cellcolor[HTML]{F98D72}0.150 & \cellcolor[HTML]{FBAE78}0.274 & \cellcolor[HTML]{FDEB84}0.508 & \cellcolor[HTML]{FDCF7E}0.399 & \cellcolor[HTML]{FBAD78}0.272 &  & \cellcolor[HTML]{FCC27C}0.350 \\
				SiamRPN & \cellcolor[HTML]{7EC67D}0.856 & \cellcolor[HTML]{A4D17F}0.752 & \cellcolor[HTML]{BCD881}0.687 & \cellcolor[HTML]{D0DE82}0.632 & \cellcolor[HTML]{FCC17C}0.346 & \cellcolor[HTML]{FEE983}0.498 & \cellcolor[HTML]{FEE081}0.462 & \cellcolor[HTML]{F9826F}0.108 &  & \cellcolor[HTML]{FDC67D}0.366 & \cellcolor[HTML]{FCB679}0.303 & \cellcolor[HTML]{FDD57F}0.421 & \cellcolor[HTML]{FCC37C}0.354 & \cellcolor[HTML]{FEDB81}0.445 & \cellcolor[HTML]{FBAD78}0.269 & \cellcolor[HTML]{FDCA7D}0.379 & \cellcolor[HTML]{F1E784}0.543 & \cellcolor[HTML]{FDD27F}0.410 & \cellcolor[HTML]{FDC67C}0.363 &  & \cellcolor[HTML]{FEDE81}0.455 \\
				DaSiamRPN & \cellcolor[HTML]{7EC67D}0.856 & \cellcolor[HTML]{A1D07F}0.759 & \cellcolor[HTML]{B7D780}0.701 & \cellcolor[HTML]{CEDD82}0.637 & \cellcolor[HTML]{FCC07B}0.341 & \cellcolor[HTML]{FEEB84}0.506 & \cellcolor[HTML]{FEDE81}0.454 & \cellcolor[HTML]{F9806F}0.101 &  & \cellcolor[HTML]{FDC57C}0.363 & \cellcolor[HTML]{FCB679}0.304 & \cellcolor[HTML]{FDD57F}0.421 & \cellcolor[HTML]{FCC37C}0.355 & \cellcolor[HTML]{FEDB80}0.443 & \cellcolor[HTML]{FBAC78}0.268 & \cellcolor[HTML]{FDCA7D}0.378 & \cellcolor[HTML]{EEE784}0.550 & \cellcolor[HTML]{FDD27F}0.409 & \cellcolor[HTML]{FDC67C}0.364 &  & \cellcolor[HTML]{FEDE81}0.456 \\
				SiamRPNPP & \cellcolor[HTML]{83C87D}0.840 & \cellcolor[HTML]{A5D27F}0.748 & \cellcolor[HTML]{B6D680}0.703 & \cellcolor[HTML]{BED981}0.680 & \cellcolor[HTML]{FDD47F}0.419 & \cellcolor[HTML]{B7D780}0.701 & \cellcolor[HTML]{F5E884}0.533 & \cellcolor[HTML]{FA9774}0.189 &  & \cellcolor[HTML]{FEDA80}0.442 & \cellcolor[HTML]{FDCB7D}0.383 & \cellcolor[HTML]{FEE783}0.489 & \cellcolor[HTML]{FEDA80}0.440 & \cellcolor[HTML]{F6E984}0.530 & \cellcolor[HTML]{FCC07B}0.342 & \cellcolor[HTML]{FEDD81}0.450 & \cellcolor[HTML]{DFE283}0.591 & \cellcolor[HTML]{FEEA83}0.500 & \cellcolor[HTML]{FEDA80}0.441 &  & \cellcolor[HTML]{F8E984}0.523 \\
				SPLT & \cellcolor[HTML]{AAD380}0.734 & \cellcolor[HTML]{E4E483}0.577 & \cellcolor[HTML]{FEE683}0.487 & \cellcolor[HTML]{FDD680}0.425 & \cellcolor[HTML]{FCB87A}0.313 & \cellcolor[HTML]{C5DA81}0.663 & \cellcolor[HTML]{FDD07E}0.402 & \cellcolor[HTML]{F98670}0.122 &  & \cellcolor[HTML]{FCBB7A}0.322 & \cellcolor[HTML]{FBA877}0.251 & \cellcolor[HTML]{FCC17B}0.346 & \cellcolor[HTML]{FCB579}0.302 & \cellcolor[HTML]{FCC07B}0.343 & \cellcolor[HTML]{FBA075}0.223 & \cellcolor[HTML]{FCBA7A}0.321 & \cellcolor[HTML]{FEE382}0.474 & \cellcolor[HTML]{FCC07B}0.344 & \cellcolor[HTML]{FCB479}0.296 &  & \cellcolor[HTML]{FDCC7E}0.386 \\
				SiamDW & \cellcolor[HTML]{7BC57D}0.863 & \cellcolor[HTML]{BDD881}0.683 & \cellcolor[HTML]{E4E383}0.579 & \cellcolor[HTML]{F3E884}0.536 & \cellcolor[HTML]{FCB77A}0.308 & \cellcolor[HTML]{FEDE81}0.455 & \cellcolor[HTML]{FDC87D}0.374 & \cellcolor[HTML]{F87A6E}0.077 &  & \cellcolor[HTML]{FCC17C}0.346 & \cellcolor[HTML]{FBA777}0.250 & \cellcolor[HTML]{FDCA7D}0.382 & \cellcolor[HTML]{FCBA7A}0.319 & \cellcolor[HTML]{FEDE81}0.457 & \cellcolor[HTML]{FBA276}0.228 & \cellcolor[HTML]{FCBE7B}0.336 & \cellcolor[HTML]{EBE583}0.560 & \cellcolor[HTML]{FEE182}0.467 & \cellcolor[HTML]{FCC37C}0.353 &  & \cellcolor[HTML]{FDD57F}0.421 \\
				SiamCAR & \cellcolor[HTML]{7AC57D}0.866 & \cellcolor[HTML]{A9D27F}0.739 & \cellcolor[HTML]{C0D981}0.674 & \cellcolor[HTML]{C4DA81}0.664 & \cellcolor[HTML]{FDD67F}0.424 & \cellcolor[HTML]{F9EA84}0.520 & \cellcolor[HTML]{FEEA83}0.501 & \cellcolor[HTML]{F9806F}0.100 &  & \cellcolor[HTML]{FCC47C}0.358 & \cellcolor[HTML]{FCB97A}0.316 & \cellcolor[HTML]{FEDC81}0.447 & \cellcolor[HTML]{FDCC7E}0.387 & \cellcolor[HTML]{FEDB80}0.443 & \cellcolor[HTML]{FBAE78}0.276 & \cellcolor[HTML]{FDD17F}0.408 & \cellcolor[HTML]{E2E383}0.583 & \cellcolor[HTML]{FEE983}0.498 & \cellcolor[HTML]{FDCF7E}0.399 &  & \cellcolor[HTML]{FEE482}0.478 \\
				SiamFCPP & \cellcolor[HTML]{80C77D}0.850 & \cellcolor[HTML]{A3D17F}0.754 & \cellcolor[HTML]{C3DA81}0.667 & \cellcolor[HTML]{CFDD82}0.636 & \cellcolor[HTML]{FEDC81}0.448 & \cellcolor[HTML]{FEE182}0.466 & \cellcolor[HTML]{FEDE81}0.456 & \cellcolor[HTML]{F9806F}0.102 &  & \cellcolor[HTML]{FEDA80}0.440 & \cellcolor[HTML]{FDCE7E}0.395 & \cellcolor[HTML]{FEE983}0.496 & \cellcolor[HTML]{FEDE81}0.455 & \cellcolor[HTML]{BDD881}0.685 & \cellcolor[HTML]{FED980}0.435 & \cellcolor[HTML]{FEE382}0.476 & \cellcolor[HTML]{E7E483}0.571 & \cellcolor[HTML]{F7E984}0.526 & \cellcolor[HTML]{FEE983}0.498 &  & \cellcolor[HTML]{F9EA84}0.520 \\
				Ocean & \cellcolor[HTML]{82C77D}0.845 & \cellcolor[HTML]{A2D17F}0.756 & \cellcolor[HTML]{B5D680}0.706 & \cellcolor[HTML]{BCD881}0.687 & \cellcolor[HTML]{FEEB84}0.507 & \cellcolor[HTML]{F9EA84}0.519 & \cellcolor[HTML]{FAEA84}0.518 & \cellcolor[HTML]{F98D72}0.150 &  & \cellcolor[HTML]{FEDF81}0.459 & \cellcolor[HTML]{FDD17F}0.407 & \cellcolor[HTML]{FAEA84}0.519 & \cellcolor[HTML]{FEE182}0.469 & \cellcolor[HTML]{C8DC81}0.654 & \cellcolor[HTML]{FEDA80}0.439 & \cellcolor[HTML]{FEEA83}0.499 & \cellcolor[HTML]{DBE182}0.603 & \cellcolor[HTML]{FAEA84}0.519 & \cellcolor[HTML]{FFEB84}0.505 &  & \cellcolor[HTML]{F1E784}0.542 \\
				SiamRCNN & \cellcolor[HTML]{70C27C}0.893 & \cellcolor[HTML]{A3D17F}0.754 & \cellcolor[HTML]{BFD981}0.677 & \cellcolor[HTML]{BCD881}0.685 & \cellcolor[HTML]{D6E082}0.615 & \cellcolor[HTML]{A8D27F}0.742 & \cellcolor[HTML]{C5DB81}0.663 & \cellcolor[HTML]{FDCD7E}0.391 &  & \cellcolor[HTML]{CEDD82}0.638 & \cellcolor[HTML]{CFDE82}0.634 & \cellcolor[HTML]{C1DA81}0.672 & \cellcolor[HTML]{CADC81}0.647 & \cellcolor[HTML]{A8D27F}0.741 & \cellcolor[HTML]{E2E383}0.583 & \cellcolor[HTML]{CCDD82}0.644 & \cellcolor[HTML]{C2DA81}0.669 & \cellcolor[HTML]{D7E082}0.612 & \cellcolor[HTML]{CADC81}0.648 &  & \cellcolor[HTML]{C5DB81}0.661 \\\hline
				ATOM & \cellcolor[HTML]{78C47D}0.872 & \cellcolor[HTML]{B6D680}0.704 & \cellcolor[HTML]{BFD981}0.680 & \cellcolor[HTML]{C5DB81}0.662 & \cellcolor[HTML]{FEDA80}0.441 & \cellcolor[HTML]{FEEB84}0.508 & \cellcolor[HTML]{FEEA83}0.500 & \cellcolor[HTML]{F9826F}0.107 &  & \cellcolor[HTML]{FDCF7E}0.400 & \cellcolor[HTML]{FCBE7B}0.334 & \cellcolor[HTML]{FEE081}0.463 & \cellcolor[HTML]{FDD17F}0.408 & \cellcolor[HTML]{FEE482}0.480 & \cellcolor[HTML]{FCB379}0.294 & \cellcolor[HTML]{FDD680}0.427 & \cellcolor[HTML]{D5DF82}0.620 & \cellcolor[HTML]{EFE784}0.547 & \cellcolor[HTML]{FDD57F}0.420 &  & \cellcolor[HTML]{FEE883}0.492 \\
				DiMP & \cellcolor[HTML]{71C27C}0.890 & \cellcolor[HTML]{96CD7E}0.791 & \cellcolor[HTML]{AED480}0.725 & \cellcolor[HTML]{B6D680}0.703 & \cellcolor[HTML]{FEE883}0.495 & \cellcolor[HTML]{D3DF82}0.624 & \cellcolor[HTML]{EAE583}0.561 & \cellcolor[HTML]{FA9172}0.165 &  & \cellcolor[HTML]{FEE182}0.466 & \cellcolor[HTML]{FDCE7E}0.393 & \cellcolor[HTML]{FAEA84}0.518 & \cellcolor[HTML]{FEE382}0.473 & \cellcolor[HTML]{DAE182}0.604 & \cellcolor[HTML]{FDCB7D}0.382 & \cellcolor[HTML]{FEE983}0.496 & \cellcolor[HTML]{C3DA81}0.667 & \cellcolor[HTML]{D4DF82}0.620 & \cellcolor[HTML]{FEE883}0.494 &  & \cellcolor[HTML]{EBE683}0.559 \\
				PrDiMP & \cellcolor[HTML]{70C27C}0.893 & \cellcolor[HTML]{9FD07F}0.765 & \cellcolor[HTML]{ABD380}0.733 & \cellcolor[HTML]{B7D680}0.701 & \cellcolor[HTML]{F6E984}0.530 & \cellcolor[HTML]{C6DB81}0.661 & \cellcolor[HTML]{E1E383}0.585 & \cellcolor[HTML]{FA9F75}0.218 &  & \cellcolor[HTML]{FEEA83}0.499 & \cellcolor[HTML]{FED980}0.437 & \cellcolor[HTML]{EEE683}0.552 & \cellcolor[HTML]{FCEB84}0.512 & \cellcolor[HTML]{D3DF82}0.623 & \cellcolor[HTML]{FDD57F}0.422 & \cellcolor[HTML]{F5E884}0.532 & \cellcolor[HTML]{BCD881}0.685 & \cellcolor[HTML]{CFDD82}0.635 & \cellcolor[HTML]{F7E984}0.527 &  & \cellcolor[HTML]{E2E383}0.584 \\
				SuperDiMP & \cellcolor[HTML]{68C07C}0.914 & \cellcolor[HTML]{94CC7E}0.795 & \cellcolor[HTML]{9FD07F}0.767 & \cellcolor[HTML]{ABD380}0.732 & \cellcolor[HTML]{E9E583}0.563 & \cellcolor[HTML]{BCD881}0.687 & \cellcolor[HTML]{C9DC81}0.650 & \cellcolor[HTML]{FBA476}0.235 &  & \cellcolor[HTML]{F5E884}0.532 & \cellcolor[HTML]{FEE482}0.477 & \cellcolor[HTML]{E1E383}0.585 & \cellcolor[HTML]{EEE683}0.551 & \cellcolor[HTML]{C9DC81}0.651 & \cellcolor[HTML]{FEE182}0.468 & \cellcolor[HTML]{E7E483}0.569 & \cellcolor[HTML]{B6D680}0.704 & \cellcolor[HTML]{C8DC81}0.653 & \cellcolor[HTML]{E9E583}0.564 &  & \cellcolor[HTML]{D6DF82}0.617 \\
				KeepTrack & \cellcolor[HTML]{63BE7B}0.927 & \cellcolor[HTML]{94CD7E}0.795 & \cellcolor[HTML]{A7D27F}0.745 & \cellcolor[HTML]{A5D17F}0.750 & \cellcolor[HTML]{DEE283}0.594 & \cellcolor[HTML]{98CE7F}0.785 & \cellcolor[HTML]{BBD881}0.688 & \cellcolor[HTML]{FBB379}0.291 &  & \cellcolor[HTML]{EDE683}0.553 & \cellcolor[HTML]{FEEA83}0.500 & \cellcolor[HTML]{DBE182}0.603 & \cellcolor[HTML]{E6E483}0.572 & \cellcolor[HTML]{C8DC81}0.653 & \cellcolor[HTML]{FEE282}0.470 & \cellcolor[HTML]{E4E483}0.577 & \cellcolor[HTML]{B3D580}0.712 & \cellcolor[HTML]{C0D981}0.676 & \cellcolor[HTML]{E5E483}0.574 &  & \cellcolor[HTML]{CEDD82}0.637 \\\hline
				GlobalTrack & \cellcolor[HTML]{99CE7F}0.782 & \cellcolor[HTML]{D3DF82}0.624 & \cellcolor[HTML]{F4E884}0.534 & \cellcolor[HTML]{FAEA84}0.518 & \cellcolor[HTML]{FEE582}0.482 & \cellcolor[HTML]{D5DF82}0.618 & \cellcolor[HTML]{F7E984}0.526 & \cellcolor[HTML]{FBA576}0.240 &  & \cellcolor[HTML]{FEE482}0.477 & \cellcolor[HTML]{FDD67F}0.424 & \cellcolor[HTML]{FDEB84}0.509 & \cellcolor[HTML]{FEE683}0.486 & \cellcolor[HTML]{E5E483}0.575 & \cellcolor[HTML]{FDD07E}0.401 & \cellcolor[HTML]{FEE482}0.477 & \cellcolor[HTML]{EBE683}0.559 & \cellcolor[HTML]{EDE683}0.554 & \cellcolor[HTML]{FEE683}0.485 &  & \cellcolor[HTML]{FBEA84}0.515 \\
				KYS & \cellcolor[HTML]{76C47D}0.878 & \cellcolor[HTML]{93CC7E}0.799 & \cellcolor[HTML]{95CD7E}0.792 & \cellcolor[HTML]{A7D27F}0.744 & \cellcolor[HTML]{FEEB84}0.506 & \cellcolor[HTML]{E6E483}0.573 & \cellcolor[HTML]{F6E984}0.529 & \cellcolor[HTML]{F98971}0.134 &  & \cellcolor[HTML]{FEDF81}0.458 & \cellcolor[HTML]{FDCB7D}0.383 & \cellcolor[HTML]{FBEA84}0.514 & \cellcolor[HTML]{FEE182}0.465 & \cellcolor[HTML]{D9E082}0.609 & \cellcolor[HTML]{FDCA7D}0.382 & \cellcolor[HTML]{FEE983}0.498 & \cellcolor[HTML]{C1D981}0.674 & \cellcolor[HTML]{D2DE82}0.627 & \cellcolor[HTML]{FEE783}0.491 &  & \cellcolor[HTML]{EBE683}0.559 \\
				MixFormer & \cellcolor[HTML]{72C37C}0.888 & \cellcolor[HTML]{96CD7E}0.791 & \cellcolor[HTML]{ACD480}0.729 & \cellcolor[HTML]{A6D27F}0.747 & \cellcolor[HTML]{C6DB81}0.658 & \cellcolor[HTML]{91CC7E}0.803 & \cellcolor[HTML]{A8D27F}0.741 & \cellcolor[HTML]{FDD07E}0.402 &  & \cellcolor[HTML]{C5DB81}0.661 & \cellcolor[HTML]{CDDD82}0.641 & \cellcolor[HTML]{B3D580}0.711 & \cellcolor[HTML]{BCD881}0.687 & \cellcolor[HTML]{9FD07F}0.764 & \cellcolor[HTML]{D4DF82}0.622 & \cellcolor[HTML]{BCD881}0.686 & \cellcolor[HTML]{A8D27F}0.740 & \cellcolor[HTML]{B2D580}0.713 & \cellcolor[HTML]{BDD881}0.685 &  & \cellcolor[HTML]{B6D680}0.704 \\
			\botrule
			\end{tabular}
			\begin{tablenotes}
				\item [1] \textit{
				$e_1$-OTB2015 \cite{OTB2015}. 
				$e_2$-VOT2016 \cite{VOT2016}.
				$e_{3}$-VOT2018 \cite{VOT2018}.
				$e_{4}$-VOT2019 \cite{VOT2019}.
				$e_{5}$-GOT-10k \cite{GOT-10k}.
				$e_{6}$-VOTLT2019 \cite{VOT2019}.
				$e_{7}$-LaSOT \cite{LaSOT}.
				$e_{8}$-VideoCube \cite{GIT}.
				$c_1$-abnormal ratio.
				$c_2$-abnormal scale.
				$c_3$-abnormal illumination.
				$c_4$-blur bbox.
				$c_5$-delta ratio.
				$c_6$-delta scale.
				$c_7$-delta illumination.
				$c_8$-delta blur bbox.
				$c_9$-fast motion.
				$c_{10}$-low correlation coefficient.
				}
				\item[2] \textit{
				The background color of cells indicates the score, with red indicating a low score and green indicating a high score.
				}
				\end{tablenotes}
		\label{tab:pre}
	\end{center}
	\end{center}
	\end{sidewaystable*}

\clearpage
\begin{sidewaystable*}[htbp!]
	\sidewaystablefn%
	\begin{center}
	\caption{Performance of 20 representative trackers on all sub-spaces, based on normalized precision score. 
	}
	\renewcommand{\arraystretch}{1.2}
	\begin{center}
		\tabcolsep=0.1cm
		\small
		\begin{tabular}{llllllllllllllllllllll}
			\toprule
			\multirow{2}{*}{\textbf{Trackers}} & \multicolumn{8}{l}{\textbf{Normal space}} &  & \multicolumn{10}{l}{\textbf{Challenging space}} & & \multirow{2}{*}{\textbf{Mean}} \\ \cline{2-9} \cline{11-20}
				& $e_1$ & $e_2$ & $e_3$ & $e_4$ & $e_5$ & $e_6$ & $e_7$ & $e_8$ &  & $c_1$ & $c_2$ & $c_3$ & $c_4$ & $c_5$ & $c_6$ & $c_7$ & $c_8$ & $c_9$ & $c_{10}$ &  \\ \midrule
				KCF & \cellcolor[HTML]{FDD07E}0.601 & \cellcolor[HTML]{FCB97A}0.488 & \cellcolor[HTML]{FAA075}0.371 & \cellcolor[HTML]{FA8E72}0.286 & \cellcolor[HTML]{FBAE78}0.437 & \cellcolor[HTML]{FA9072}0.292 & \cellcolor[HTML]{FA9E75}0.361 & \cellcolor[HTML]{F8716C}0.146 &  & \cellcolor[HTML]{FBB178}0.450 & \cellcolor[HTML]{FBB178}0.450 & \cellcolor[HTML]{FBB178}0.450 & \cellcolor[HTML]{FBB178}0.450 & \cellcolor[HTML]{FBB178}0.450 & \cellcolor[HTML]{FBB178}0.450 & \cellcolor[HTML]{FBB178}0.450 & \cellcolor[HTML]{FBB178}0.450 & \cellcolor[HTML]{FBB178}0.450 & \cellcolor[HTML]{FBB178}0.450 &  & \cellcolor[HTML]{FBB178}0.450 \\
			ECO & \cellcolor[HTML]{70C27C}0.945 & \cellcolor[HTML]{C2DA81}0.821 & \cellcolor[HTML]{DFE283}0.775 & \cellcolor[HTML]{FEE482}0.696 & \cellcolor[HTML]{F1E784}0.748 & \cellcolor[HTML]{FCBC7B}0.506 & \cellcolor[HTML]{FDD57F}0.622 & \cellcolor[HTML]{F97C6E}0.201 &  & \cellcolor[HTML]{FEE582}0.698 & \cellcolor[HTML]{FEE582}0.698 & \cellcolor[HTML]{FEE582}0.698 & \cellcolor[HTML]{FEE582}0.698 & \cellcolor[HTML]{FEE582}0.698 & \cellcolor[HTML]{FEE582}0.698 & \cellcolor[HTML]{FEE582}0.698 & \cellcolor[HTML]{FEE582}0.698 & \cellcolor[HTML]{FEE582}0.698 & \cellcolor[HTML]{FEE582}0.698 &  & \cellcolor[HTML]{FEE582}0.698 \\\hline
			SiamFC & \cellcolor[HTML]{D2DE82}0.796 & \cellcolor[HTML]{FEE582}0.700 & \cellcolor[HTML]{FDD27F}0.611 & \cellcolor[HTML]{FCC47C}0.543 & \cellcolor[HTML]{FDCB7D}0.574 & \cellcolor[HTML]{FBAB77}0.424 & \cellcolor[HTML]{FCBF7B}0.520 & \cellcolor[HTML]{F8696B}0.106 &  & \cellcolor[HTML]{FCC57C}0.545 & \cellcolor[HTML]{FCC57C}0.545 & \cellcolor[HTML]{FCC57C}0.545 & \cellcolor[HTML]{FCC57C}0.545 & \cellcolor[HTML]{FCC57C}0.545 & \cellcolor[HTML]{FCC57C}0.545 & \cellcolor[HTML]{FCC57C}0.545 & \cellcolor[HTML]{FCC57C}0.545 & \cellcolor[HTML]{FCC57C}0.545 & \cellcolor[HTML]{FCC57C}0.545 &  & \cellcolor[HTML]{FCC57C}0.545 \\
			SiamRPN & \cellcolor[HTML]{88C97E}0.908 & \cellcolor[HTML]{BCD881}0.830 & \cellcolor[HTML]{E8E583}0.762 & \cellcolor[HTML]{FEE282}0.686 & \cellcolor[HTML]{FEE783}0.709 & \cellcolor[HTML]{FCC57C}0.547 & \cellcolor[HTML]{FEDF81}0.671 & \cellcolor[HTML]{FA9E75}0.363 &  & \cellcolor[HTML]{FEE482}0.697 & \cellcolor[HTML]{FEE482}0.697 & \cellcolor[HTML]{FEE482}0.697 & \cellcolor[HTML]{FEE482}0.697 & \cellcolor[HTML]{FEE482}0.697 & \cellcolor[HTML]{FEE482}0.697 & \cellcolor[HTML]{FEE482}0.697 & \cellcolor[HTML]{FEE482}0.697 & \cellcolor[HTML]{FEE482}0.697 & \cellcolor[HTML]{FEE482}0.697 &  & \cellcolor[HTML]{FEE482}0.697 \\
			DaSiamRPN & \cellcolor[HTML]{89C97E}0.906 & \cellcolor[HTML]{B8D780}0.836 & \cellcolor[HTML]{E0E283}0.774 & \cellcolor[HTML]{FEE382}0.690 & \cellcolor[HTML]{FEE583}0.702 & \cellcolor[HTML]{FDC77D}0.557 & \cellcolor[HTML]{FEDF81}0.669 & \cellcolor[HTML]{FA9D75}0.359 &  & \cellcolor[HTML]{FEE582}0.698 & \cellcolor[HTML]{FEE582}0.698 & \cellcolor[HTML]{FEE582}0.698 & \cellcolor[HTML]{FEE582}0.698 & \cellcolor[HTML]{FEE582}0.698 & \cellcolor[HTML]{FEE582}0.698 & \cellcolor[HTML]{FEE582}0.698 & \cellcolor[HTML]{FEE582}0.698 & \cellcolor[HTML]{FEE582}0.698 & \cellcolor[HTML]{FEE582}0.698 &  & \cellcolor[HTML]{FEE582}0.698 \\
			SiamRPNPP & \cellcolor[HTML]{9BCF7F}0.879 & \cellcolor[HTML]{C7DB81}0.812 & \cellcolor[HTML]{EAE583}0.760 & \cellcolor[HTML]{FEE983}0.718 & \cellcolor[HTML]{F2E884}0.746 & \cellcolor[HTML]{EAE583}0.758 & \cellcolor[HTML]{FEE883}0.714 & \cellcolor[HTML]{FCB379}0.460 &  & \cellcolor[HTML]{F6E984}0.740 & \cellcolor[HTML]{F6E984}0.740 & \cellcolor[HTML]{F6E984}0.740 & \cellcolor[HTML]{F6E984}0.740 & \cellcolor[HTML]{F6E984}0.740 & \cellcolor[HTML]{F6E984}0.740 & \cellcolor[HTML]{F6E984}0.740 & \cellcolor[HTML]{F6E984}0.740 & \cellcolor[HTML]{F6E984}0.740 & \cellcolor[HTML]{F6E984}0.740 &  & \cellcolor[HTML]{F6E984}0.740 \\
			SPLT & \cellcolor[HTML]{CBDC81}0.807 & \cellcolor[HTML]{FEDE81}0.666 & \cellcolor[HTML]{FDC97D}0.564 & \cellcolor[HTML]{FCB97A}0.488 & \cellcolor[HTML]{FEDF81}0.673 & \cellcolor[HTML]{F3E884}0.745 & \cellcolor[HTML]{FED980}0.642 & \cellcolor[HTML]{FBAB77}0.425 &  & \cellcolor[HTML]{FEDA80}0.649 & \cellcolor[HTML]{FEDA80}0.649 & \cellcolor[HTML]{FEDA80}0.649 & \cellcolor[HTML]{FEDA80}0.649 & \cellcolor[HTML]{FEDA80}0.649 & \cellcolor[HTML]{FEDA80}0.649 & \cellcolor[HTML]{FEDA80}0.649 & \cellcolor[HTML]{FEDA80}0.649 & \cellcolor[HTML]{FEDA80}0.649 & \cellcolor[HTML]{FEDA80}0.649 &  & \cellcolor[HTML]{FEDA80}0.649 \\
			SiamDW & \cellcolor[HTML]{81C77D}0.920 & \cellcolor[HTML]{CCDD82}0.804 & \cellcolor[HTML]{FEE182}0.679 & \cellcolor[HTML]{FDD17F}0.606 & \cellcolor[HTML]{FEE683}0.703 & \cellcolor[HTML]{FCC27C}0.533 & \cellcolor[HTML]{FEDA80}0.649 & \cellcolor[HTML]{FA9F75}0.365 &  & \cellcolor[HTML]{FEE482}0.695 & \cellcolor[HTML]{FEE482}0.695 & \cellcolor[HTML]{FEE482}0.695 & \cellcolor[HTML]{FEE482}0.695 & \cellcolor[HTML]{FEE482}0.695 & \cellcolor[HTML]{FEE482}0.695 & \cellcolor[HTML]{FEE482}0.695 & \cellcolor[HTML]{FEE482}0.695 & \cellcolor[HTML]{FEE482}0.695 & \cellcolor[HTML]{FEE482}0.695 &  & \cellcolor[HTML]{FEE482}0.695 \\
			SiamCAR & \cellcolor[HTML]{8ECB7E}0.899 & \cellcolor[HTML]{CADC81}0.808 & \cellcolor[HTML]{FAEA84}0.735 & \cellcolor[HTML]{FEE482}0.696 & \cellcolor[HTML]{FEEB84}0.728 & \cellcolor[HTML]{FDCB7D}0.576 & \cellcolor[HTML]{FEE382}0.689 & \cellcolor[HTML]{FA8E72}0.286 &  & \cellcolor[HTML]{FEE482}0.694 & \cellcolor[HTML]{FEE482}0.694 & \cellcolor[HTML]{FEE482}0.694 & \cellcolor[HTML]{FEE482}0.694 & \cellcolor[HTML]{FEE482}0.694 & \cellcolor[HTML]{FEE482}0.694 & \cellcolor[HTML]{FEE482}0.694 & \cellcolor[HTML]{FEE482}0.694 & \cellcolor[HTML]{FEE482}0.694 & \cellcolor[HTML]{FEE482}0.694 &  & \cellcolor[HTML]{FEE482}0.694 \\
			SiamFCPP & \cellcolor[HTML]{90CB7E}0.896 & \cellcolor[HTML]{B3D580}0.843 & \cellcolor[HTML]{F3E884}0.745 & \cellcolor[HTML]{FEE382}0.688 & \cellcolor[HTML]{E7E483}0.764 & \cellcolor[HTML]{FCBE7B}0.513 & \cellcolor[HTML]{FEDA80}0.646 & \cellcolor[HTML]{FA9874}0.332 &  & \cellcolor[HTML]{FEE883}0.713 & \cellcolor[HTML]{FEE883}0.713 & \cellcolor[HTML]{FEE883}0.713 & \cellcolor[HTML]{FEE883}0.713 & \cellcolor[HTML]{FEE883}0.713 & \cellcolor[HTML]{FEE883}0.713 & \cellcolor[HTML]{FEE883}0.713 & \cellcolor[HTML]{FEE883}0.713 & \cellcolor[HTML]{FEE883}0.713 & \cellcolor[HTML]{FEE883}0.713 &  & \cellcolor[HTML]{FEE883}0.713 \\
			Ocean & \cellcolor[HTML]{93CC7E}0.892 & \cellcolor[HTML]{BAD780}0.832 & \cellcolor[HTML]{E2E383}0.771 & \cellcolor[HTML]{FEEA83}0.726 & \cellcolor[HTML]{D3DF82}0.793 & \cellcolor[HTML]{FDCC7E}0.579 & \cellcolor[HTML]{FEE382}0.691 & \cellcolor[HTML]{FBA977}0.415 &  & \cellcolor[HTML]{FBEA84}0.733 & \cellcolor[HTML]{FBEA84}0.733 & \cellcolor[HTML]{FBEA84}0.733 & \cellcolor[HTML]{FBEA84}0.733 & \cellcolor[HTML]{FBEA84}0.733 & \cellcolor[HTML]{FBEA84}0.733 & \cellcolor[HTML]{FBEA84}0.733 & \cellcolor[HTML]{FBEA84}0.733 & \cellcolor[HTML]{FBEA84}0.733 & \cellcolor[HTML]{FBEA84}0.733 &  & \cellcolor[HTML]{FBEA84}0.733 \\
			SiamRCNN & \cellcolor[HTML]{72C37C}0.942 & \cellcolor[HTML]{BCD881}0.829 & \cellcolor[HTML]{F0E784}0.749 & \cellcolor[HTML]{FAEA84}0.735 & \cellcolor[HTML]{D1DE82}0.797 & \cellcolor[HTML]{D5DF82}0.791 & \cellcolor[HTML]{DFE283}0.776 & \cellcolor[HTML]{FDCB7D}0.574 &  & \cellcolor[HTML]{D7E082}0.788 & \cellcolor[HTML]{D7E082}0.788 & \cellcolor[HTML]{D7E082}0.788 & \cellcolor[HTML]{D7E082}0.788 & \cellcolor[HTML]{D7E082}0.788 & \cellcolor[HTML]{D7E082}0.788 & \cellcolor[HTML]{D7E082}0.788 & \cellcolor[HTML]{D7E082}0.788 & \cellcolor[HTML]{D7E082}0.788 & \cellcolor[HTML]{D7E082}0.788 &  & \cellcolor[HTML]{D7E082}0.788 \\\hline
			ATOM & \cellcolor[HTML]{7DC67D}0.926 & \cellcolor[HTML]{D2DE82}0.795 & \cellcolor[HTML]{F5E984}0.741 & \cellcolor[HTML]{FEE683}0.703 & \cellcolor[HTML]{E9E583}0.761 & \cellcolor[HTML]{FDCA7D}0.572 & \cellcolor[HTML]{FEE683}0.706 & \cellcolor[HTML]{FA9C74}0.351 &  & \cellcolor[HTML]{FFEB84}0.726 & \cellcolor[HTML]{FFEB84}0.726 & \cellcolor[HTML]{FFEB84}0.726 & \cellcolor[HTML]{FFEB84}0.726 & \cellcolor[HTML]{FFEB84}0.726 & \cellcolor[HTML]{FFEB84}0.726 & \cellcolor[HTML]{FFEB84}0.726 & \cellcolor[HTML]{FFEB84}0.726 & \cellcolor[HTML]{FFEB84}0.726 & \cellcolor[HTML]{FFEB84}0.726 &  & \cellcolor[HTML]{FFEB84}0.726 \\
			DiMP & \cellcolor[HTML]{76C47D}0.937 & \cellcolor[HTML]{A0D07F}0.872 & \cellcolor[HTML]{D2DE82}0.796 & \cellcolor[HTML]{EAE583}0.760 & \cellcolor[HTML]{CFDE82}0.800 & \cellcolor[HTML]{FEE382}0.691 & \cellcolor[HTML]{F0E784}0.749 & \cellcolor[HTML]{FBAD78}0.432 &  & \cellcolor[HTML]{E0E383}0.774 & \cellcolor[HTML]{E0E383}0.774 & \cellcolor[HTML]{E0E383}0.774 & \cellcolor[HTML]{E0E383}0.774 & \cellcolor[HTML]{E0E383}0.774 & \cellcolor[HTML]{E0E383}0.774 & \cellcolor[HTML]{E0E383}0.774 & \cellcolor[HTML]{E0E383}0.774 & \cellcolor[HTML]{E0E383}0.774 & \cellcolor[HTML]{E0E383}0.774 &  & \cellcolor[HTML]{E0E383}0.774 \\
			PrDiMP & \cellcolor[HTML]{75C37C}0.938 & \cellcolor[HTML]{AFD480}0.849 & \cellcolor[HTML]{D1DE82}0.798 & \cellcolor[HTML]{F3E884}0.745 & \cellcolor[HTML]{C8DB81}0.811 & \cellcolor[HTML]{FEEB84}0.728 & \cellcolor[HTML]{E1E383}0.773 & \cellcolor[HTML]{FCB97A}0.492 &  & \cellcolor[HTML]{D6E082}0.789 & \cellcolor[HTML]{D6E082}0.789 & \cellcolor[HTML]{D6E082}0.789 & \cellcolor[HTML]{D6E082}0.789 & \cellcolor[HTML]{D6E082}0.789 & \cellcolor[HTML]{D6E082}0.789 & \cellcolor[HTML]{D6E082}0.789 & \cellcolor[HTML]{D6E082}0.789 & \cellcolor[HTML]{D6E082}0.789 & \cellcolor[HTML]{D6E082}0.789 &  & \cellcolor[HTML]{D6E082}0.789 \\
			SuperDiMP & \cellcolor[HTML]{65BF7C}0.962 & \cellcolor[HTML]{9CCF7F}0.878 & \cellcolor[HTML]{B7D780}0.837 & \cellcolor[HTML]{D5DF82}0.792 & \cellcolor[HTML]{BED981}0.826 & \cellcolor[HTML]{F1E784}0.749 & \cellcolor[HTML]{CFDE82}0.800 & \cellcolor[HTML]{FCB87A}0.487 &  & \cellcolor[HTML]{CCDD82}0.805 & \cellcolor[HTML]{CCDD82}0.805 & \cellcolor[HTML]{CCDD82}0.805 & \cellcolor[HTML]{CCDD82}0.805 & \cellcolor[HTML]{CCDD82}0.805 & \cellcolor[HTML]{CCDD82}0.805 & \cellcolor[HTML]{CCDD82}0.805 & \cellcolor[HTML]{CCDD82}0.805 & \cellcolor[HTML]{CCDD82}0.805 & \cellcolor[HTML]{CCDD82}0.805 &  & \cellcolor[HTML]{CCDD82}0.805 \\
			KeepTrack & \cellcolor[HTML]{63BE7B}0.965 & \cellcolor[HTML]{9FD07F}0.874 & \cellcolor[HTML]{C6DB81}0.814 & \cellcolor[HTML]{D8E082}0.787 & \cellcolor[HTML]{ACD380}0.854 & \cellcolor[HTML]{B7D780}0.837 & \cellcolor[HTML]{B5D680}0.841 & \cellcolor[HTML]{FDC67D}0.553 &  & \cellcolor[HTML]{C1D981}0.822 & \cellcolor[HTML]{C1D981}0.822 & \cellcolor[HTML]{C1D981}0.822 & \cellcolor[HTML]{C1D981}0.822 & \cellcolor[HTML]{C1D981}0.822 & \cellcolor[HTML]{C1D981}0.822 & \cellcolor[HTML]{C1D981}0.822 & \cellcolor[HTML]{C1D981}0.822 & \cellcolor[HTML]{C1D981}0.822 & \cellcolor[HTML]{C1D981}0.822 &  & \cellcolor[HTML]{C1D981}0.822 \\\hline
			GlobalTrack & \cellcolor[HTML]{BFD981}0.825 & \cellcolor[HTML]{FEE482}0.694 & \cellcolor[HTML]{FDCF7E}0.594 & \cellcolor[HTML]{FDC77D}0.557 & \cellcolor[HTML]{FCEA84}0.731 & \cellcolor[HTML]{FEDD81}0.663 & \cellcolor[HTML]{FEE182}0.679 & \cellcolor[HTML]{FCBC7B}0.506 &  & \cellcolor[HTML]{FEE482}0.695 & \cellcolor[HTML]{FEE482}0.695 & \cellcolor[HTML]{FEE482}0.695 & \cellcolor[HTML]{FEE482}0.695 & \cellcolor[HTML]{FEE482}0.695 & \cellcolor[HTML]{FEE482}0.695 & \cellcolor[HTML]{FEE482}0.695 & \cellcolor[HTML]{FEE482}0.695 & \cellcolor[HTML]{FEE482}0.695 & \cellcolor[HTML]{FEE482}0.695 &  & \cellcolor[HTML]{FEE482}0.695 \\
			KYS & \cellcolor[HTML]{7CC67D}0.927 & \cellcolor[HTML]{97CD7E}0.885 & \cellcolor[HTML]{AAD380}0.857 & \cellcolor[HTML]{CFDD82}0.801 & \cellcolor[HTML]{C7DB81}0.812 & \cellcolor[HTML]{FDD680}0.630 & \cellcolor[HTML]{FEE783}0.709 & \cellcolor[HTML]{FBA075}0.373 &  & \cellcolor[HTML]{E3E383}0.769 & \cellcolor[HTML]{E3E383}0.769 & \cellcolor[HTML]{E3E383}0.769 & \cellcolor[HTML]{E3E383}0.769 & \cellcolor[HTML]{E3E383}0.769 & \cellcolor[HTML]{E3E383}0.769 & \cellcolor[HTML]{E3E383}0.769 & \cellcolor[HTML]{E3E383}0.769 & \cellcolor[HTML]{E3E383}0.769 & \cellcolor[HTML]{E3E383}0.769 &  & \cellcolor[HTML]{E3E383}0.769 \\
			MixFormer & \cellcolor[HTML]{82C77D}0.917 & \cellcolor[HTML]{A7D27F}0.862 & \cellcolor[HTML]{D4DF82}0.792 & \cellcolor[HTML]{D8E082}0.786 & \cellcolor[HTML]{AED480}0.850 & \cellcolor[HTML]{B8D780}0.836 & \cellcolor[HTML]{B5D680}0.840 & \cellcolor[HTML]{FDD17F}0.603 &  & \cellcolor[HTML]{B9D780}0.833 & \cellcolor[HTML]{B9D780}0.833 & \cellcolor[HTML]{B9D780}0.833 & \cellcolor[HTML]{B9D780}0.833 & \cellcolor[HTML]{B9D780}0.833 & \cellcolor[HTML]{B9D780}0.833 & \cellcolor[HTML]{B9D780}0.833 & \cellcolor[HTML]{B9D780}0.833 & \cellcolor[HTML]{B9D780}0.833 & \cellcolor[HTML]{B9D780}0.833 &  & \cellcolor[HTML]{B9D780}0.833 \\
			\botrule
			\end{tabular}
			\begin{tablenotes}
				\item [1] \textit{
				$e_1$-OTB2015 \cite{OTB2015}. 
				$e_2$-VOT2016 \cite{VOT2016}.
				$e_{3}$-VOT2018 \cite{VOT2018}.
				$e_{4}$-VOT2019 \cite{VOT2019}.
				$e_{5}$-GOT-10k \cite{GOT-10k}.
				$e_{6}$-VOTLT2019 \cite{VOT2019}.
				$e_{7}$-LaSOT \cite{LaSOT}.
				$e_{8}$-VideoCube \cite{GIT}.
				$c_1$-abnormal ratio.
				$c_2$-abnormal scale.
				$c_3$-abnormal illumination.
				$c_4$-blur bbox.
				$c_5$-delta ratio.
				$c_6$-delta scale.
				$c_7$-delta illumination.
				$c_8$-delta blur bbox.
				$c_9$-fast motion.
				$c_{10}$-low correlation coefficient.
				}
				\item[2] \textit{
				The background color of cells indicates the score, with red indicating a low score and green indicating a high score.
				}
				\end{tablenotes}
		\label{tab:npre}
	\end{center}
	\end{center}
	\end{sidewaystable*}

\clearpage
\begin{sidewaystable*}[htbp!]
	\sidewaystablefn%
	\begin{center}
	\caption{Performance of 20 representative trackers on all sub-spaces, based on success score. 
	}
	\renewcommand{\arraystretch}{1.2}
	\begin{center}
		\tabcolsep=0.1cm
		\small
		\begin{tabular}{llllllllllllllllllllll}
			\toprule
			\multirow{2}{*}{\textbf{Trackers}} & \multicolumn{8}{l}{\textbf{Normal space}} &  & \multicolumn{10}{l}{\textbf{Challenging space}} & & \multirow{2}{*}{\textbf{Mean}} \\ \cline{2-9} \cline{11-20}
				& $e_1$ & $e_2$ & $e_3$ & $e_4$ & $e_5$ & $e_6$ & $e_7$ & $e_8$ &  & $c_1$ & $c_2$ & $c_3$ & $c_4$ & $c_5$ & $c_6$ & $c_7$ & $c_8$ & $c_9$ & $c_{10}$ &  \\ \midrule
				KCF & \cellcolor[HTML]{FCC37C}0.380 & \cellcolor[HTML]{FBA276}0.262 & \cellcolor[HTML]{FA9172}0.201 & \cellcolor[HTML]{F98971}0.170 & \cellcolor[HTML]{FA9A74}0.231 & \cellcolor[HTML]{F9806F}0.137 & \cellcolor[HTML]{F98A71}0.175 & \cellcolor[HTML]{F86A6B}0.058 &  & \cellcolor[HTML]{FBAD78}0.301 & \cellcolor[HTML]{FCB479}0.329 & \cellcolor[HTML]{FCB579}0.332 & \cellcolor[HTML]{FBAD78}0.301 & \cellcolor[HTML]{FCBA7A}0.349 & \cellcolor[HTML]{FCBC7B}0.357 & \cellcolor[HTML]{FCB579}0.332 & \cellcolor[HTML]{FAA075}0.255 & \cellcolor[HTML]{F98370}0.147 & \cellcolor[HTML]{FBB078}0.312 &  & \cellcolor[HTML]{FBA175}0.257 \\
ECO & \cellcolor[HTML]{9DCF7F}0.690 & \cellcolor[HTML]{FEEA83}0.524 & \cellcolor[HTML]{FEDB80}0.469 & \cellcolor[HTML]{FDCA7D}0.407 & \cellcolor[HTML]{FDCE7E}0.424 & \cellcolor[HTML]{FCB479}0.328 & \cellcolor[HTML]{FCC27C}0.378 & \cellcolor[HTML]{F8766D}0.100 &  & \cellcolor[HTML]{FDCE7E}0.423 & \cellcolor[HTML]{FDD780}0.456 & \cellcolor[HTML]{FEE282}0.497 & \cellcolor[HTML]{FDD680}0.453 & \cellcolor[HTML]{FEEA83}0.525 & \cellcolor[HTML]{FEE783}0.515 & \cellcolor[HTML]{FEE783}0.511 & \cellcolor[HTML]{FDD47F}0.444 & \cellcolor[HTML]{FCC57C}0.388 & \cellcolor[HTML]{FEE282}0.495 &  & \cellcolor[HTML]{FDD57F}0.446 \\ \hline
SiamFC & \cellcolor[HTML]{E0E283}0.579 & \cellcolor[HTML]{FDCD7E}0.419 & \cellcolor[HTML]{FCBB7A}0.352 & \cellcolor[HTML]{FBB178}0.315 & \cellcolor[HTML]{FCB579}0.332 & \cellcolor[HTML]{FBA877}0.285 & \cellcolor[HTML]{FBB279}0.319 & \cellcolor[HTML]{F8696B}0.052 &  & \cellcolor[HTML]{FBAE78}0.306 & \cellcolor[HTML]{FCBB7A}0.351 & \cellcolor[HTML]{FCC17C}0.375 & \cellcolor[HTML]{FCB87A}0.343 & \cellcolor[HTML]{FDD07E}0.429 & \cellcolor[HTML]{FCB679}0.333 & \cellcolor[HTML]{FCBB7A}0.353 & \cellcolor[HTML]{FCC27C}0.379 & \cellcolor[HTML]{FBAD78}0.301 & \cellcolor[HTML]{FCBE7B}0.362 &  & \cellcolor[HTML]{FCB97A}0.344 \\
SiamRPN & \cellcolor[HTML]{B3D580}0.654 & \cellcolor[HTML]{F2E884}0.549 & \cellcolor[HTML]{FEE081}0.486 & \cellcolor[HTML]{FDD17F}0.433 & \cellcolor[HTML]{FDD680}0.451 & \cellcolor[HTML]{FDCA7D}0.406 & \cellcolor[HTML]{FEDD81}0.479 & \cellcolor[HTML]{FA9773}0.222 &  & \cellcolor[HTML]{FED880}0.458 & \cellcolor[HTML]{FEE983}0.520 & \cellcolor[HTML]{FDEB84}0.530 & \cellcolor[HTML]{FEE282}0.497 & \cellcolor[HTML]{F8E984}0.538 & \cellcolor[HTML]{E9E583}0.564 & \cellcolor[HTML]{F6E984}0.543 & \cellcolor[HTML]{FDCD7E}0.420 & \cellcolor[HTML]{FBB078}0.313 & \cellcolor[HTML]{FEE783}0.513 &  & \cellcolor[HTML]{FEDD81}0.476 \\
DaSiamRPN & \cellcolor[HTML]{B3D680}0.653 & \cellcolor[HTML]{EFE784}0.554 & \cellcolor[HTML]{FEE282}0.493 & \cellcolor[HTML]{FDD27F}0.435 & \cellcolor[HTML]{FDD57F}0.447 & \cellcolor[HTML]{FDCC7E}0.414 & \cellcolor[HTML]{FEDC81}0.473 & \cellcolor[HTML]{FA9573}0.215 &  & \cellcolor[HTML]{FDD780}0.456 & \cellcolor[HTML]{FEE983}0.522 & \cellcolor[HTML]{FDEB84}0.530 & \cellcolor[HTML]{FEE382}0.498 & \cellcolor[HTML]{F8E984}0.538 & \cellcolor[HTML]{EAE583}0.562 & \cellcolor[HTML]{F6E984}0.542 & \cellcolor[HTML]{FDCE7E}0.424 & \cellcolor[HTML]{FBB078}0.313 & \cellcolor[HTML]{FEE783}0.514 &  & \cellcolor[HTML]{FEDD81}0.477 \\
SiamRPNPP & \cellcolor[HTML]{B6D680}0.648 & \cellcolor[HTML]{EDE683}0.557 & \cellcolor[HTML]{FEE482}0.501 & \cellcolor[HTML]{FEDB81}0.471 & \cellcolor[HTML]{FEE182}0.490 & \cellcolor[HTML]{EAE583}0.562 & \cellcolor[HTML]{FEEB84}0.528 & \cellcolor[HTML]{FBAE78}0.304 &  & \cellcolor[HTML]{FEE582}0.507 & \cellcolor[HTML]{E6E483}0.569 & \cellcolor[HTML]{E4E483}0.572 & \cellcolor[HTML]{F3E884}0.547 & \cellcolor[HTML]{E3E383}0.573 & \cellcolor[HTML]{D5DF82}0.597 & \cellcolor[HTML]{DFE283}0.581 & \cellcolor[HTML]{FED980}0.463 & \cellcolor[HTML]{FCC47C}0.387 & \cellcolor[HTML]{ECE683}0.558 &  & \cellcolor[HTML]{FEEA83}0.523 \\
SPLT & \cellcolor[HTML]{EBE683}0.560 & \cellcolor[HTML]{FDD17F}0.433 & \cellcolor[HTML]{FCBD7B}0.359 & \cellcolor[HTML]{FBAD78}0.302 & \cellcolor[HTML]{FDD17F}0.434 & \cellcolor[HTML]{FEEA83}0.526 & \cellcolor[HTML]{FDD17F}0.432 & \cellcolor[HTML]{FBA175}0.257 &  & \cellcolor[HTML]{FDCE7E}0.421 & \cellcolor[HTML]{FEE182}0.491 & \cellcolor[HTML]{FEDF81}0.484 & \cellcolor[HTML]{FEDA80}0.466 & \cellcolor[HTML]{FDCE7E}0.422 & \cellcolor[HTML]{FEE683}0.509 & \cellcolor[HTML]{FEE683}0.508 & \cellcolor[HTML]{FCBD7B}0.358 & \cellcolor[HTML]{FBA476}0.270 & \cellcolor[HTML]{FEDA80}0.466 &  & \cellcolor[HTML]{FDD07E}0.428 \\
SiamDW & \cellcolor[HTML]{B1D580}0.657 & \cellcolor[HTML]{FEE382}0.499 & \cellcolor[HTML]{FDCB7D}0.409 & \cellcolor[HTML]{FCBE7B}0.365 & \cellcolor[HTML]{FDCA7D}0.408 & \cellcolor[HTML]{FCB479}0.328 & \cellcolor[HTML]{FCC17B}0.374 & \cellcolor[HTML]{F97D6E}0.127 &  & \cellcolor[HTML]{FDD07E}0.430 & \cellcolor[HTML]{FEDB81}0.470 & \cellcolor[HTML]{FEE282}0.495 & \cellcolor[HTML]{FED980}0.462 & \cellcolor[HTML]{F7E984}0.539 & \cellcolor[HTML]{FEEA83}0.523 & \cellcolor[HTML]{FEE683}0.511 & \cellcolor[HTML]{FDCE7E}0.420 & \cellcolor[HTML]{FCBD7B}0.361 & \cellcolor[HTML]{FEE382}0.499 &  & \cellcolor[HTML]{FDD27F}0.438 \\
SiamCAR & \cellcolor[HTML]{AFD480}0.660 & \cellcolor[HTML]{F6E984}0.542 & \cellcolor[HTML]{FEDA80}0.466 & \cellcolor[HTML]{FDD47F}0.443 & \cellcolor[HTML]{FEE182}0.492 & \cellcolor[HTML]{FDCD7E}0.419 & \cellcolor[HTML]{FEE081}0.487 & \cellcolor[HTML]{F9816F}0.142 &  & \cellcolor[HTML]{FDCE7E}0.422 & \cellcolor[HTML]{FEE482}0.502 & \cellcolor[HTML]{FEEA83}0.524 & \cellcolor[HTML]{FEE182}0.490 & \cellcolor[HTML]{FEE582}0.506 & \cellcolor[HTML]{FEEA83}0.525 & \cellcolor[HTML]{FBEA84}0.533 & \cellcolor[HTML]{FDD67F}0.450 & \cellcolor[HTML]{FCC17C}0.375 & \cellcolor[HTML]{FEE783}0.513 &  & \cellcolor[HTML]{FEDC81}0.472 \\
SiamFCPP & \cellcolor[HTML]{B7D780}0.647 & \cellcolor[HTML]{F1E784}0.551 & \cellcolor[HTML]{FEDB80}0.469 & \cellcolor[HTML]{FDD27F}0.436 & \cellcolor[HTML]{FFEB84}0.526 & \cellcolor[HTML]{FCC37C}0.382 & \cellcolor[HTML]{FEDC81}0.474 & \cellcolor[HTML]{FA9373}0.205 &  & \cellcolor[HTML]{FEE382}0.499 & \cellcolor[HTML]{E8E583}0.566 & \cellcolor[HTML]{E1E383}0.576 & \cellcolor[HTML]{EFE784}0.554 & \cellcolor[HTML]{B1D580}0.657 & \cellcolor[HTML]{B7D780}0.647 & \cellcolor[HTML]{D1DE82}0.604 & \cellcolor[HTML]{FDD57F}0.449 & \cellcolor[HTML]{FDCF7E}0.424 & \cellcolor[HTML]{D3DF82}0.601 &  & \cellcolor[HTML]{FEE783}0.515 \\
Ocean & \cellcolor[HTML]{B4D680}0.652 & \cellcolor[HTML]{EAE583}0.562 & \cellcolor[HTML]{FEE482}0.503 & \cellcolor[HTML]{FEDC81}0.473 & \cellcolor[HTML]{EBE683}0.560 & \cellcolor[HTML]{FDD07E}0.429 & \cellcolor[HTML]{FEE683}0.511 & \cellcolor[HTML]{FBA276}0.262 &  & \cellcolor[HTML]{FEE482}0.503 & \cellcolor[HTML]{E3E383}0.573 & \cellcolor[HTML]{DEE283}0.582 & \cellcolor[HTML]{EBE683}0.560 & \cellcolor[HTML]{BDD881}0.637 & \cellcolor[HTML]{BED981}0.635 & \cellcolor[HTML]{D3DF82}0.600 & \cellcolor[HTML]{FEDB81}0.469 & \cellcolor[HTML]{FDCC7E}0.413 & \cellcolor[HTML]{D8E082}0.592 &  & \cellcolor[HTML]{FEEB84}0.529 \\
SiamRCNN & \cellcolor[HTML]{91CC7E}0.711 & \cellcolor[HTML]{DFE283}0.580 & \cellcolor[HTML]{FEE783}0.514 & \cellcolor[HTML]{FEE582}0.504 & \cellcolor[HTML]{B5D680}0.650 & \cellcolor[HTML]{CADC81}0.615 & \cellcolor[HTML]{BAD881}0.642 & \cellcolor[HTML]{FEDA80}0.467 &  & \cellcolor[HTML]{B5D680}0.651 & \cellcolor[HTML]{8DCA7E}0.718 & \cellcolor[HTML]{96CD7E}0.703 & \cellcolor[HTML]{9BCF7F}0.693 & \cellcolor[HTML]{8FCB7E}0.714 & \cellcolor[HTML]{82C77D}0.736 & \cellcolor[HTML]{8DCB7E}0.717 & \cellcolor[HTML]{F0E784}0.552 & \cellcolor[HTML]{FEE883}0.516 & \cellcolor[HTML]{97CD7E}0.700 &  & \cellcolor[HTML]{C0D981}0.632 \\\hline
ATOM & \cellcolor[HTML]{A6D27F}0.675 & \cellcolor[HTML]{FAEA84}0.536 & \cellcolor[HTML]{FEE182}0.492 & \cellcolor[HTML]{FED880}0.458 & \cellcolor[HTML]{FEEB84}0.528 & \cellcolor[HTML]{FDCD7E}0.419 & \cellcolor[HTML]{FEE683}0.511 & \cellcolor[HTML]{FA9373}0.207 &  & \cellcolor[HTML]{FEDF81}0.483 & \cellcolor[HTML]{F4E884}0.546 & \cellcolor[HTML]{E8E583}0.565 & \cellcolor[HTML]{F8E984}0.539 & \cellcolor[HTML]{EAE583}0.562 & \cellcolor[HTML]{DEE283}0.583 & \cellcolor[HTML]{DDE182}0.584 & \cellcolor[HTML]{FEE182}0.492 & \cellcolor[HTML]{FDD07E}0.429 & \cellcolor[HTML]{EBE683}0.560 &  & \cellcolor[HTML]{FEE683}0.509 \\
DiMP & \cellcolor[HTML]{9ACE7F}0.696 & \cellcolor[HTML]{D4DF82}0.599 & \cellcolor[HTML]{FBEA84}0.534 & \cellcolor[HTML]{FEE482}0.502 & \cellcolor[HTML]{DAE182}0.589 & \cellcolor[HTML]{FEE783}0.514 & \cellcolor[HTML]{E5E483}0.571 & \cellcolor[HTML]{FBAB77}0.296 &  & \cellcolor[HTML]{F8E984}0.539 & \cellcolor[HTML]{D2DE82}0.602 & \cellcolor[HTML]{C8DB81}0.619 & \cellcolor[HTML]{D4DF82}0.599 & \cellcolor[HTML]{B9D780}0.644 & \cellcolor[HTML]{B0D480}0.660 & \cellcolor[HTML]{B7D780}0.647 & \cellcolor[HTML]{FAEA84}0.534 & \cellcolor[HTML]{FEE582}0.504 & \cellcolor[HTML]{C3DA81}0.627 &  & \cellcolor[HTML]{E5E483}0.571 \\
PrDiMP & \cellcolor[HTML]{95CD7E}0.703 & \cellcolor[HTML]{DAE182}0.590 & \cellcolor[HTML]{F5E884}0.544 & \cellcolor[HTML]{FEE582}0.504 & \cellcolor[HTML]{CBDC81}0.614 & \cellcolor[HTML]{F2E884}0.549 & \cellcolor[HTML]{D6DF82}0.596 & \cellcolor[HTML]{FCB679}0.335 &  & \cellcolor[HTML]{E4E483}0.571 & \cellcolor[HTML]{C1DA81}0.630 & \cellcolor[HTML]{BAD780}0.643 & \cellcolor[HTML]{C3DA81}0.628 & \cellcolor[HTML]{ACD380}0.665 & \cellcolor[HTML]{AAD380}0.670 & \cellcolor[HTML]{ABD380}0.668 & \cellcolor[HTML]{EEE683}0.555 & \cellcolor[HTML]{FEEA83}0.526 & \cellcolor[HTML]{B7D780}0.647 &  & \cellcolor[HTML]{D9E082}0.591 \\
SuperDiMP & \cellcolor[HTML]{8CCA7E}0.719 & \cellcolor[HTML]{CEDD82}0.609 & \cellcolor[HTML]{E8E583}0.565 & \cellcolor[HTML]{FDEB84}0.531 & \cellcolor[HTML]{B8D780}0.646 & \cellcolor[HTML]{E4E483}0.571 & \cellcolor[HTML]{BBD881}0.641 & \cellcolor[HTML]{FCBC7B}0.357 &  & \cellcolor[HTML]{D7E082}0.594 & \cellcolor[HTML]{B0D480}0.660 & \cellcolor[HTML]{A9D380}0.670 & \cellcolor[HTML]{B1D580}0.657 & \cellcolor[HTML]{A0D07F}0.685 & \cellcolor[HTML]{91CC7E}0.711 & \cellcolor[HTML]{9ACE7F}0.696 & \cellcolor[HTML]{E2E383}0.576 & \cellcolor[HTML]{F7E984}0.541 & \cellcolor[HTML]{A7D27F}0.675 &  & \cellcolor[HTML]{C9DC81}0.617 \\
KeepTrack & \cellcolor[HTML]{89C97E}0.725 & \cellcolor[HTML]{CBDC81}0.615 & \cellcolor[HTML]{EBE683}0.560 & \cellcolor[HTML]{F8E984}0.539 & \cellcolor[HTML]{ACD380}0.666 & \cellcolor[HTML]{B9D780}0.643 & \cellcolor[HTML]{A9D27F}0.671 & \cellcolor[HTML]{FDCA7D}0.407 &  & \cellcolor[HTML]{CCDD82}0.612 & \cellcolor[HTML]{A4D17F}0.678 & \cellcolor[HTML]{A1D07F}0.684 & \cellcolor[HTML]{A7D27F}0.674 & \cellcolor[HTML]{A0D07F}0.685 & \cellcolor[HTML]{92CC7E}0.710 & \cellcolor[HTML]{96CD7E}0.702 & \cellcolor[HTML]{DEE283}0.582 & \cellcolor[HTML]{EBE683}0.560 & \cellcolor[HTML]{A2D07F}0.683 &  & \cellcolor[HTML]{BFD981}0.633 \\\hline
GlobalTrack & \cellcolor[HTML]{C4DA81}0.625 & \cellcolor[HTML]{FEE182}0.490 & \cellcolor[HTML]{FDCB7D}0.411 & \cellcolor[HTML]{FCC37C}0.381 & \cellcolor[HTML]{EAE583}0.562 & \cellcolor[HTML]{FEE482}0.502 & \cellcolor[HTML]{FFEB84}0.527 & \cellcolor[HTML]{FCBD7B}0.360 &  & \cellcolor[HTML]{F9EA84}0.536 & \cellcolor[HTML]{CEDD82}0.608 & \cellcolor[HTML]{D4DF82}0.598 & \cellcolor[HTML]{D6E082}0.595 & \cellcolor[HTML]{DDE283}0.583 & \cellcolor[HTML]{C3DA81}0.627 & \cellcolor[HTML]{CDDD82}0.611 & \cellcolor[HTML]{FDD780}0.456 & \cellcolor[HTML]{FDD780}0.455 & \cellcolor[HTML]{D6DF82}0.596 &  & \cellcolor[HTML]{FEEB84}0.529 \\
KYS & \cellcolor[HTML]{A0D07F}0.686 & \cellcolor[HTML]{CEDD82}0.608 & \cellcolor[HTML]{E2E383}0.576 & \cellcolor[HTML]{FEEB84}0.528 & \cellcolor[HTML]{D3DF82}0.601 & \cellcolor[HTML]{FEDC81}0.474 & \cellcolor[HTML]{F6E984}0.542 & \cellcolor[HTML]{FA9F75}0.251 &  & \cellcolor[HTML]{FDEB84}0.530 & \cellcolor[HTML]{DAE182}0.589 & \cellcolor[HTML]{CBDC81}0.613 & \cellcolor[HTML]{D9E082}0.590 & \cellcolor[HTML]{B7D680}0.648 & \cellcolor[HTML]{B1D580}0.658 & \cellcolor[HTML]{B7D680}0.648 & \cellcolor[HTML]{F9EA84}0.536 & \cellcolor[HTML]{FEE683}0.508 & \cellcolor[HTML]{C5DB81}0.624 &  & \cellcolor[HTML]{E7E483}0.567 \\
MixFormer & \cellcolor[HTML]{98CE7F}0.699 & \cellcolor[HTML]{D5DF82}0.597 & \cellcolor[HTML]{FBEA84}0.533 & \cellcolor[HTML]{FCEA84}0.532 & \cellcolor[HTML]{98CE7F}0.699 & \cellcolor[HTML]{B1D580}0.657 & \cellcolor[HTML]{96CD7E}0.702 & \cellcolor[HTML]{FEE082}0.489 &  & \cellcolor[HTML]{A5D17F}0.678 & \cellcolor[HTML]{7FC67D}0.741 & \cellcolor[HTML]{7EC67D}0.743 & \cellcolor[HTML]{84C87D}0.732 & \cellcolor[HTML]{7FC67D}0.741 & \cellcolor[HTML]{63BE7B}0.787 & \cellcolor[HTML]{72C37C}0.763 & \cellcolor[HTML]{CDDD82}0.610 & \cellcolor[HTML]{D5DF82}0.596 & \cellcolor[HTML]{7CC67D}0.745 &  & \cellcolor[HTML]{AAD380}0.669 \\
			\botrule
			\end{tabular}
			\begin{tablenotes}
				\item [1] \textit{
				$e_1$-OTB2015 \cite{OTB2015}. 
				$e_2$-VOT2016 \cite{VOT2016}.
				$e_{3}$-VOT2018 \cite{VOT2018}.
				$e_{4}$-VOT2019 \cite{VOT2019}.
				$e_{5}$-GOT-10k \cite{GOT-10k}.
				$e_{6}$-VOTLT2019 \cite{VOT2019}.
				$e_{7}$-LaSOT \cite{LaSOT}.
				$e_{8}$-VideoCube \cite{GIT}.
				$c_1$-abnormal ratio.
				$c_2$-abnormal scale.
				$c_3$-abnormal illumination.
				$c_4$-blur bbox.
				$c_5$-delta ratio.
				$c_6$-delta scale.
				$c_7$-delta illumination.
				$c_8$-delta blur bbox.
				$c_9$-fast motion.
				$c_{10}$-low correlation coefficient.
				}
				\item[2] \textit{
				The background color of cells indicates the score, with red indicating a low score and green indicating a high score.
				}
				\end{tablenotes}
		\label{tab:success}
	\end{center}
	\end{center}
	\end{sidewaystable*}

\clearpage
\begin{sidewaystable*}[htbp!]
	\sidewaystablefn%
	\begin{center}
	\caption{Performance of 20 representative trackers on all sub-spaces, based on precision score, weighted by sequences' length. 
	}
	\renewcommand{\arraystretch}{1.2}
	\begin{center}
		\tabcolsep=0.1cm
		\small
		\begin{tabular}{llllllllllllllllllllll}
			\toprule
			\multirow{2}{*}{\textbf{Trackers}} & \multicolumn{8}{l}{\textbf{Normal space}} &  & \multicolumn{10}{l}{\textbf{Challenging space}} & & \multirow{2}{*}{\textbf{Mean}} \\ \cline{2-9} \cline{11-20}
				& $e_1$ & $e_2$ & $e_3$ & $e_4$ & $e_5$ & $e_6$ & $e_7$ & $e_8$ &  & $c_1$ & $c_2$ & $c_3$ & $c_4$ & $c_5$ & $c_6$ & $c_7$ & $c_8$ & $c_9$ & $c_{10}$ &  \\ \midrule
				KCF & \cellcolor[HTML]{E6E483}0.519 & \cellcolor[HTML]{FDD780}0.373 & \cellcolor[HTML]{FCBE7B}0.290 & \cellcolor[HTML]{FBB279}0.253 & \cellcolor[HTML]{F98A71}0.121 & \cellcolor[HTML]{F98A71}0.120 & \cellcolor[HTML]{FA9172}0.144 & \cellcolor[HTML]{F8696B}0.011 &  & \cellcolor[HTML]{F97D6E}0.077 & \cellcolor[HTML]{F8736D}0.047 & \cellcolor[HTML]{F98370}0.099 & \cellcolor[HTML]{F8736C}0.045 & \cellcolor[HTML]{FBA476}0.208 & \cellcolor[HTML]{F98770}0.110 & \cellcolor[HTML]{F98D71}0.129 & \cellcolor[HTML]{FBA877}0.220 & \cellcolor[HTML]{F98370}0.099 & \cellcolor[HTML]{F98871}0.114 &  & \cellcolor[HTML]{FA9874}0.165 \\
ECO & \cellcolor[HTML]{66BF7C}0.926 & \cellcolor[HTML]{93CC7E}0.782 & \cellcolor[HTML]{A7D27F}0.719 & \cellcolor[HTML]{C0D981}0.640 & \cellcolor[HTML]{FDCC7E}0.337 & \cellcolor[HTML]{FDCE7E}0.345 & \cellcolor[HTML]{FDD37F}0.360 & \cellcolor[HTML]{F86F6C}0.032 &  & \cellcolor[HTML]{FA9172}0.142 & \cellcolor[HTML]{F9816F}0.091 & \cellcolor[HTML]{FBA776}0.215 & \cellcolor[HTML]{F9826F}0.096 & \cellcolor[HTML]{FEE983}0.431 & \cellcolor[HTML]{FBA576}0.210 & \cellcolor[HTML]{FCB97A}0.274 & \cellcolor[HTML]{ECE683}0.500 & \cellcolor[HTML]{FCC27C}0.304 & \cellcolor[HTML]{FBB279}0.251 &  & \cellcolor[HTML]{FDD680}0.370 \\ \hline
SiamFC & \cellcolor[HTML]{94CC7E}0.781 & \cellcolor[HTML]{C4DA81}0.628 & \cellcolor[HTML]{E0E283}0.539 & \cellcolor[HTML]{E8E583}0.513 & \cellcolor[HTML]{FCB579}0.261 & \cellcolor[HTML]{FCBD7B}0.289 & \cellcolor[HTML]{FDC67D}0.319 & \cellcolor[HTML]{F86F6C}0.031 &  & \cellcolor[HTML]{F98670}0.109 & \cellcolor[HTML]{F97E6F}0.081 & \cellcolor[HTML]{FA9D75}0.183 & \cellcolor[HTML]{F97F6F}0.085 & \cellcolor[HTML]{FCC07B}0.297 & \cellcolor[HTML]{FA9072}0.140 & \cellcolor[HTML]{FBA777}0.217 & \cellcolor[HTML]{FEE282}0.411 & \cellcolor[HTML]{FBAC78}0.233 & \cellcolor[HTML]{FA9E75}0.186 &  & \cellcolor[HTML]{FCBF7B}0.295 \\
SiamRPN & \cellcolor[HTML]{79C57D}0.867 & \cellcolor[HTML]{94CC7E}0.780 & \cellcolor[HTML]{ADD480}0.699 & \cellcolor[HTML]{BDD881}0.648 & \cellcolor[HTML]{FDCC7E}0.339 & \cellcolor[HTML]{E3E383}0.528 & \cellcolor[HTML]{FAEA84}0.456 & \cellcolor[HTML]{F98871}0.116 &  & \cellcolor[HTML]{FBA175}0.196 & \cellcolor[HTML]{FA9874}0.166 & \cellcolor[HTML]{FCBB7A}0.283 & \cellcolor[HTML]{FA9C74}0.179 & \cellcolor[HTML]{FEE382}0.412 & \cellcolor[HTML]{FBB279}0.252 & \cellcolor[HTML]{FDC77D}0.320 & \cellcolor[HTML]{F8E984}0.461 & \cellcolor[HTML]{FCB379}0.256 & \cellcolor[HTML]{FCBC7B}0.286 &  & \cellcolor[HTML]{FEE081}0.402 \\
DaSiamRPN & \cellcolor[HTML]{7CC57D}0.857 & \cellcolor[HTML]{93CC7E}0.782 & \cellcolor[HTML]{AAD380}0.710 & \cellcolor[HTML]{B8D780}0.665 & \cellcolor[HTML]{FDCB7D}0.334 & \cellcolor[HTML]{E1E383}0.535 & \cellcolor[HTML]{FDEB84}0.445 & \cellcolor[HTML]{F98770}0.110 &  & \cellcolor[HTML]{FBA075}0.194 & \cellcolor[HTML]{FA9874}0.168 & \cellcolor[HTML]{FCBB7A}0.283 & \cellcolor[HTML]{FA9B74}0.177 & \cellcolor[HTML]{FEE282}0.410 & \cellcolor[HTML]{FBB179}0.250 & \cellcolor[HTML]{FDC67C}0.316 & \cellcolor[HTML]{F2E884}0.481 & \cellcolor[HTML]{FCB379}0.255 & \cellcolor[HTML]{FCBA7A}0.279 &  & \cellcolor[HTML]{FEE081}0.403 \\
SiamRPNPP & \cellcolor[HTML]{7AC57D}0.863 & \cellcolor[HTML]{8DCA7E}0.802 & \cellcolor[HTML]{A2D07F}0.737 & \cellcolor[HTML]{A9D380}0.713 & \cellcolor[HTML]{FEDF81}0.399 & \cellcolor[HTML]{ABD380}0.706 & \cellcolor[HTML]{E2E383}0.532 & \cellcolor[HTML]{FBA276}0.199 &  & \cellcolor[HTML]{FCB87A}0.271 & \cellcolor[HTML]{FBAD78}0.235 & \cellcolor[HTML]{FDD17F}0.353 & \cellcolor[HTML]{FCB379}0.256 & \cellcolor[HTML]{E7E483}0.517 & \cellcolor[HTML]{FDC87D}0.325 & \cellcolor[HTML]{FEDD81}0.393 & \cellcolor[HTML]{EBE683}0.502 & \cellcolor[HTML]{FDD37F}0.359 & \cellcolor[HTML]{FDD780}0.373 &  & \cellcolor[HTML]{F4E884}0.474 \\
SPLT & \cellcolor[HTML]{95CD7E}0.776 & \cellcolor[HTML]{C2DA81}0.633 & \cellcolor[HTML]{E3E383}0.527 & \cellcolor[HTML]{EBE683}0.502 & \cellcolor[HTML]{FCBF7B}0.296 & \cellcolor[HTML]{B2D580}0.684 & \cellcolor[HTML]{FEDF81}0.400 & \cellcolor[HTML]{FA8E72}0.134 &  & \cellcolor[HTML]{FBA175}0.195 & \cellcolor[HTML]{FA9473}0.155 & \cellcolor[HTML]{FCB379}0.256 & \cellcolor[HTML]{FA9B74}0.178 & \cellcolor[HTML]{FDCB7D}0.334 & \cellcolor[HTML]{FBA576}0.209 & \cellcolor[HTML]{FCBB7A}0.281 & \cellcolor[HTML]{F9EA84}0.458 & \cellcolor[HTML]{FCB379}0.257 & \cellcolor[HTML]{FCB679}0.266 &  & \cellcolor[HTML]{FDD47F}0.363 \\
SiamDW & \cellcolor[HTML]{6FC27C}0.899 & \cellcolor[HTML]{9ACE7F}0.762 & \cellcolor[HTML]{BFD981}0.644 & \cellcolor[HTML]{C9DC81}0.611 & \cellcolor[HTML]{FCBE7B}0.292 & \cellcolor[HTML]{E9E583}0.509 & \cellcolor[HTML]{FDD780}0.372 & \cellcolor[HTML]{F9806F}0.087 &  & \cellcolor[HTML]{FA9773}0.163 & \cellcolor[HTML]{F98971}0.119 & \cellcolor[HTML]{FBAC78}0.234 & \cellcolor[HTML]{FA8E72}0.135 & \cellcolor[HTML]{FEE482}0.415 & \cellcolor[HTML]{FBA576}0.210 & \cellcolor[HTML]{FCBA7A}0.278 & \cellcolor[HTML]{F9EA84}0.457 & \cellcolor[HTML]{FCBA7A}0.279 & \cellcolor[HTML]{FCB579}0.261 &  & \cellcolor[HTML]{FDD780}0.374 \\
SiamCAR & \cellcolor[HTML]{7CC67D}0.856 & \cellcolor[HTML]{A0D07F}0.742 & \cellcolor[HTML]{BAD881}0.658 & \cellcolor[HTML]{BAD881}0.658 & \cellcolor[HTML]{FEE683}0.424 & \cellcolor[HTML]{EAE583}0.505 & \cellcolor[HTML]{EFE784}0.492 & \cellcolor[HTML]{F98570}0.103 &  & \cellcolor[HTML]{FA9D75}0.184 & \cellcolor[HTML]{FA9473}0.155 & \cellcolor[HTML]{FCBD7B}0.288 & \cellcolor[HTML]{FA9A74}0.174 & \cellcolor[HTML]{FEE282}0.409 & \cellcolor[HTML]{FBB279}0.252 & \cellcolor[HTML]{FDCD7E}0.342 & \cellcolor[HTML]{F3E884}0.476 & \cellcolor[HTML]{FCC27C}0.305 & \cellcolor[HTML]{FCC17C}0.301 &  & \cellcolor[HTML]{FEE182}0.407 \\
SiamFCPP & \cellcolor[HTML]{79C57D}0.864 & \cellcolor[HTML]{8CCA7E}0.805 & \cellcolor[HTML]{A2D07F}0.737 & \cellcolor[HTML]{A9D27F}0.714 & \cellcolor[HTML]{FEEA83}0.435 & \cellcolor[HTML]{FEDF81}0.399 & \cellcolor[HTML]{FDEB84}0.446 & \cellcolor[HTML]{F98570}0.105 &  & \cellcolor[HTML]{FBA977}0.222 & \cellcolor[HTML]{FA9E75}0.185 & \cellcolor[HTML]{FCBF7B}0.295 & \cellcolor[HTML]{FAA075}0.193 & \cellcolor[HTML]{BBD881}0.654 & \cellcolor[HTML]{FEE482}0.415 & \cellcolor[HTML]{FDD680}0.372 & \cellcolor[HTML]{FDEB84}0.446 & \cellcolor[HTML]{FCBF7B}0.296 & \cellcolor[HTML]{FDCE7E}0.346 &  & \cellcolor[HTML]{FFEB84}0.440 \\
Ocean & \cellcolor[HTML]{82C77D}0.836 & \cellcolor[HTML]{8ACA7E}0.810 & \cellcolor[HTML]{A5D17F}0.727 & \cellcolor[HTML]{ABD380}0.705 & \cellcolor[HTML]{EFE784}0.490 & \cellcolor[HTML]{DAE182}0.557 & \cellcolor[HTML]{E5E483}0.521 & \cellcolor[HTML]{FA9473}0.155 &  & \cellcolor[HTML]{FCB479}0.259 & \cellcolor[HTML]{FBAD78}0.236 & \cellcolor[HTML]{FDD27F}0.356 & \cellcolor[HTML]{FBB279}0.251 & \cellcolor[HTML]{C5DB81}0.622 & \cellcolor[HTML]{FEE683}0.423 & \cellcolor[HTML]{FEE482}0.417 & \cellcolor[HTML]{EDE683}0.498 & \cellcolor[HTML]{FCC47C}0.313 & \cellcolor[HTML]{FEDA80}0.382 &  & \cellcolor[HTML]{F4E884}0.476 \\
SiamRCNN & \cellcolor[HTML]{69C07C}0.916 & \cellcolor[HTML]{93CC7E}0.782 & \cellcolor[HTML]{AFD480}0.693 & \cellcolor[HTML]{ADD480}0.699 & \cellcolor[HTML]{C7DB81}0.618 & \cellcolor[HTML]{9ECF7F}0.747 & \cellcolor[HTML]{B9D780}0.663 & \cellcolor[HTML]{FEE082}0.404 &  & \cellcolor[HTML]{F2E884}0.480 & \cellcolor[HTML]{F4E884}0.474 & \cellcolor[HTML]{DBE182}0.554 & \cellcolor[HTML]{F2E884}0.480 & \cellcolor[HTML]{A7D27F}0.718 & \cellcolor[HTML]{DAE182}0.557 & \cellcolor[HTML]{D5DF82}0.572 & \cellcolor[HTML]{C4DA81}0.627 & \cellcolor[HTML]{E0E283}0.539 & \cellcolor[HTML]{CEDD82}0.596 &  & \cellcolor[HTML]{C7DB81}0.618 \\ \hline
ATOM & \cellcolor[HTML]{75C47D}0.878 & \cellcolor[HTML]{A1D07F}0.739 & \cellcolor[HTML]{9ECF7F}0.749 & \cellcolor[HTML]{A8D27F}0.716 & \cellcolor[HTML]{FAEA84}0.455 & \cellcolor[HTML]{EDE683}0.495 & \cellcolor[HTML]{F2E884}0.480 & \cellcolor[HTML]{F98770}0.112 &  & \cellcolor[HTML]{FBA476}0.205 & \cellcolor[HTML]{FA9A74}0.174 & \cellcolor[HTML]{FCC37C}0.308 & \cellcolor[HTML]{FA9F75}0.189 & \cellcolor[HTML]{FEEB84}0.442 & \cellcolor[HTML]{FCB87A}0.273 & \cellcolor[HTML]{FDD27F}0.357 & \cellcolor[HTML]{D8E082}0.562 & \cellcolor[HTML]{FDD37F}0.360 & \cellcolor[HTML]{FDCC7E}0.337 &  & \cellcolor[HTML]{FEEA83}0.435 \\
DiMP & \cellcolor[HTML]{74C37C}0.883 & \cellcolor[HTML]{88C97E}0.818 & \cellcolor[HTML]{A4D17F}0.729 & \cellcolor[HTML]{A5D17F}0.724 & \cellcolor[HTML]{EAE583}0.507 & \cellcolor[HTML]{BFD981}0.643 & \cellcolor[HTML]{DBE182}0.553 & \cellcolor[HTML]{FA9B74}0.177 &  & \cellcolor[HTML]{FCB579}0.261 & \cellcolor[HTML]{FBAB77}0.228 & \cellcolor[HTML]{FDD67F}0.369 & \cellcolor[HTML]{FBB078}0.246 & \cellcolor[HTML]{D3DF82}0.579 & \cellcolor[HTML]{FDD47F}0.365 & \cellcolor[HTML]{FEE683}0.421 & \cellcolor[HTML]{BFD981}0.642 & \cellcolor[HTML]{FEEA83}0.437 & \cellcolor[HTML]{FEDF81}0.399 &  & \cellcolor[HTML]{ECE683}0.499 \\
PrDiMP & \cellcolor[HTML]{6FC27C}0.898 & \cellcolor[HTML]{91CC7E}0.788 & \cellcolor[HTML]{A5D17F}0.726 & \cellcolor[HTML]{A9D380}0.712 & \cellcolor[HTML]{E2E383}0.533 & \cellcolor[HTML]{C6DB81}0.622 & \cellcolor[HTML]{D3DF82}0.578 & \cellcolor[HTML]{FBAA77}0.227 &  & \cellcolor[HTML]{FDC67C}0.317 & \cellcolor[HTML]{FCBC7B}0.285 & \cellcolor[HTML]{FEE883}0.428 & \cellcolor[HTML]{FCC47C}0.312 & \cellcolor[HTML]{CADC81}0.607 & \cellcolor[HTML]{FEE182}0.406 & \cellcolor[HTML]{F6E984}0.468 & \cellcolor[HTML]{B8D780}0.665 & \cellcolor[HTML]{F7E984}0.463 & \cellcolor[HTML]{FDEB84}0.446 &  & \cellcolor[HTML]{E3E383}0.527 \\
SuperDiMP & \cellcolor[HTML]{67BF7C}0.924 & \cellcolor[HTML]{85C87D}0.827 & \cellcolor[HTML]{99CE7F}0.763 & \cellcolor[HTML]{A1D07F}0.739 & \cellcolor[HTML]{D4DF82}0.574 & \cellcolor[HTML]{ABD380}0.707 & \cellcolor[HTML]{C0D981}0.641 & \cellcolor[HTML]{FBAF78}0.242 &  & \cellcolor[HTML]{FDCE7E}0.345 & \cellcolor[HTML]{FDC87D}0.324 & \cellcolor[HTML]{F6E984}0.467 & \cellcolor[HTML]{FDCF7E}0.347 & \cellcolor[HTML]{C3DA81}0.631 & \cellcolor[HTML]{FBEA84}0.453 & \cellcolor[HTML]{E8E583}0.514 & \cellcolor[HTML]{ABD380}0.708 & \cellcolor[HTML]{F0E784}0.488 & \cellcolor[HTML]{F4E884}0.474 &  & \cellcolor[HTML]{D8E082}0.565 \\
KeepTrack & \cellcolor[HTML]{63BE7B}0.934 & \cellcolor[HTML]{82C77D}0.838 & \cellcolor[HTML]{94CD7E}0.779 & \cellcolor[HTML]{A1D07F}0.738 & \cellcolor[HTML]{CDDD82}0.598 & \cellcolor[HTML]{90CB7E}0.793 & \cellcolor[HTML]{B4D680}0.676 & \cellcolor[HTML]{FCC17B}0.300 &  & \cellcolor[HTML]{FEDB81}0.387 & \cellcolor[HTML]{FDD37F}0.362 & \cellcolor[HTML]{EBE683}0.503 & \cellcolor[HTML]{FEDB81}0.387 & \cellcolor[HTML]{C1DA81}0.636 & \cellcolor[HTML]{FAEA84}0.454 & \cellcolor[HTML]{E3E383}0.527 & \cellcolor[HTML]{A1D07F}0.737 & \cellcolor[HTML]{DAE182}0.557 & \cellcolor[HTML]{E6E483}0.520 &  & \cellcolor[HTML]{CEDD82}0.596 \\ \hline
GlobalTrack & \cellcolor[HTML]{8ECB7E}0.798 & \cellcolor[HTML]{BFD981}0.644 & \cellcolor[HTML]{DBE182}0.553 & \cellcolor[HTML]{DEE283}0.543 & \cellcolor[HTML]{F3E884}0.478 & \cellcolor[HTML]{C2DA81}0.635 & \cellcolor[HTML]{E3E383}0.529 & \cellcolor[HTML]{FBB279}0.253 &  & \cellcolor[HTML]{FCC57C}0.313 & \cellcolor[HTML]{FCB97A}0.274 & \cellcolor[HTML]{FEDC81}0.390 & \cellcolor[HTML]{FCC17C}0.301 & \cellcolor[HTML]{D8E082}0.563 & \cellcolor[HTML]{FEDC81}0.391 & \cellcolor[HTML]{FEE382}0.415 & \cellcolor[HTML]{E6E483}0.518 & \cellcolor[HTML]{F7E984}0.463 & \cellcolor[HTML]{FFEB84}0.438 &  & \cellcolor[HTML]{F5E884}0.472 \\
KYS & \cellcolor[HTML]{7FC77D}0.845 & \cellcolor[HTML]{88C97E}0.818 & \cellcolor[HTML]{88C97E}0.817 & \cellcolor[HTML]{A2D17F}0.734 & \cellcolor[HTML]{EAE583}0.506 & \cellcolor[HTML]{DCE182}0.551 & \cellcolor[HTML]{E5E483}0.523 & \cellcolor[HTML]{FA9172}0.142 &  & \cellcolor[HTML]{FBB078}0.245 & \cellcolor[HTML]{FBA676}0.213 & \cellcolor[HTML]{FDD07E}0.350 & \cellcolor[HTML]{FBAA77}0.227 & \cellcolor[HTML]{D3DF82}0.579 & \cellcolor[HTML]{FDD47F}0.363 & \cellcolor[HTML]{FEE482}0.418 & \cellcolor[HTML]{C4DA81}0.626 & \cellcolor[HTML]{FEE783}0.428 & \cellcolor[HTML]{FEDB80}0.386 &  & \cellcolor[HTML]{F0E784}0.487 \\
MixFormer & \cellcolor[HTML]{6AC07C}0.913 & \cellcolor[HTML]{81C77D}0.840 & \cellcolor[HTML]{9FD07F}0.744 & \cellcolor[HTML]{9ED07F}0.746 & \cellcolor[HTML]{B7D780}0.667 & \cellcolor[HTML]{8BCA7E}0.808 & \cellcolor[HTML]{A1D07F}0.737 & \cellcolor[HTML]{FEE382}0.412 &  & \cellcolor[HTML]{E9E583}0.509 & \cellcolor[HTML]{EDE683}0.496 & \cellcolor[HTML]{C9DC81}0.612 & \cellcolor[HTML]{E5E483}0.522 & \cellcolor[HTML]{9FD07F}0.743 & \cellcolor[HTML]{CADC81}0.609 & \cellcolor[HTML]{C2DA81}0.634 & \cellcolor[HTML]{9CCF7F}0.754 & \cellcolor[HTML]{D3DF82}0.578 & \cellcolor[HTML]{C6DB81}0.620 &  & \cellcolor[HTML]{B8D780}0.664 \\
			\botrule
			\end{tabular}
			\begin{tablenotes}
				\item [1] \textit{
				$e_1$-OTB2015 \cite{OTB2015}. 
				$e_2$-VOT2016 \cite{VOT2016}.
				$e_{3}$-VOT2018 \cite{VOT2018}.
				$e_{4}$-VOT2019 \cite{VOT2019}.
				$e_{5}$-GOT-10k \cite{GOT-10k}.
				$e_{6}$-VOTLT2019 \cite{VOT2019}.
				$e_{7}$-LaSOT \cite{LaSOT}.
				$e_{8}$-VideoCube \cite{GIT}.
				$c_1$-abnormal ratio.
				$c_2$-abnormal scale.
				$c_3$-abnormal illumination.
				$c_4$-blur bbox.
				$c_5$-delta ratio.
				$c_6$-delta scale.
				$c_7$-delta illumination.
				$c_8$-delta blur bbox.
				$c_9$-fast motion.
				$c_{10}$-low correlation coefficient.
				}
				\item[2] \textit{
				The background color of cells indicates the score, with red indicating a low score and green indicating a high score.
				}
				\end{tablenotes}
		\label{tab:pre-l}
	\end{center}
	\end{center}
	\end{sidewaystable*}

\clearpage
\begin{sidewaystable*}[htbp!]
	\sidewaystablefn%
	\begin{center}
	\caption{Performance of 20 representative trackers on all sub-spaces, based on normalized precision score, weighted by sequences' length. 
	}
	\renewcommand{\arraystretch}{1.2}
	\begin{center}
		\tabcolsep=0.1cm
		\small
		\begin{tabular}{llllllllllllllllllllll}
			\toprule
			\multirow{2}{*}{\textbf{Trackers}} & \multicolumn{8}{l}{\textbf{Normal space}} &  & \multicolumn{10}{l}{\textbf{Challenging space}} & & \multirow{2}{*}{\textbf{Mean}} \\ \cline{2-9} \cline{11-20}
				& $e_1$ & $e_2$ & $e_3$ & $e_4$ & $e_5$ & $e_6$ & $e_7$ & $e_8$ &  & $c_1$ & $c_2$ & $c_3$ & $c_4$ & $c_5$ & $c_6$ & $c_7$ & $c_8$ & $c_9$ & $c_{10}$ &  \\ \midrule
				KCF & \cellcolor[HTML]{FEE082}0.629 & \cellcolor[HTML]{FCC07B}0.487 & \cellcolor[HTML]{FBA576}0.371 & \cellcolor[HTML]{FA9B74}0.326 & \cellcolor[HTML]{FCB579}0.441 & \cellcolor[HTML]{F97E6F}0.202 & \cellcolor[HTML]{FA9E75}0.342 & \cellcolor[HTML]{F86F6C}0.138 &  & \cellcolor[HTML]{FA8E72}0.270 & \cellcolor[HTML]{FA9974}0.321 & \cellcolor[HTML]{FA9A74}0.325 & \cellcolor[HTML]{F98D71}0.266 & \cellcolor[HTML]{FDC77D}0.520 & \cellcolor[HTML]{FED980}0.597 & \cellcolor[HTML]{FCB77A}0.447 & \cellcolor[HTML]{F98C71}0.263 & \cellcolor[HTML]{F86B6B}0.118 & \cellcolor[HTML]{FBA276}0.360 &  & \cellcolor[HTML]{FBA276}0.357 \\
ECO & \cellcolor[HTML]{74C37C}0.944 & \cellcolor[HTML]{B0D580}0.828 & \cellcolor[HTML]{D8E082}0.750 & \cellcolor[HTML]{FEE883}0.662 & \cellcolor[HTML]{D6E082}0.753 & \cellcolor[HTML]{FCB679}0.444 & \cellcolor[HTML]{FEDE81}0.617 & \cellcolor[HTML]{F97C6E}0.193 &  & \cellcolor[HTML]{FBA877}0.382 & \cellcolor[HTML]{FBB178}0.421 & \cellcolor[HTML]{FCC17C}0.493 & \cellcolor[HTML]{FBA877}0.385 & \cellcolor[HTML]{CDDD82}0.771 & \cellcolor[HTML]{B3D680}0.821 & \cellcolor[HTML]{F7E984}0.689 & \cellcolor[HTML]{FDCE7E}0.551 & \cellcolor[HTML]{FA9D75}0.335 & \cellcolor[HTML]{FDCF7E}0.553 &  & \cellcolor[HTML]{FDD780}0.589 \\ \hline
SiamFC & \cellcolor[HTML]{AFD480}0.830 & \cellcolor[HTML]{E1E383}0.733 & \cellcolor[HTML]{FEDE81}0.619 & \cellcolor[HTML]{FDD680}0.586 & \cellcolor[HTML]{FDD780}0.587 & \cellcolor[HTML]{FA9B74}0.329 & \cellcolor[HTML]{FCC57C}0.510 & \cellcolor[HTML]{F8696B}0.108 &  & \cellcolor[HTML]{F98570}0.231 & \cellcolor[HTML]{F98770}0.240 & \cellcolor[HTML]{FA9F75}0.344 & \cellcolor[HTML]{F98370}0.222 & \cellcolor[HTML]{FEE182}0.633 & \cellcolor[HTML]{FDC97D}0.529 & \cellcolor[HTML]{FCB87A}0.453 & \cellcolor[HTML]{FCB87A}0.452 & \cellcolor[HTML]{F98B71}0.257 & \cellcolor[HTML]{FBA977}0.389 &  & \cellcolor[HTML]{FCB77A}0.447 \\
SiamRPN & \cellcolor[HTML]{84C87D}0.913 & \cellcolor[HTML]{ACD380}0.836 & \cellcolor[HTML]{DAE182}0.747 & \cellcolor[HTML]{F4E884}0.695 & \cellcolor[HTML]{F1E784}0.700 & \cellcolor[HTML]{FDD37F}0.569 & \cellcolor[HTML]{FEE883}0.661 & \cellcolor[HTML]{FBA276}0.359 &  & \cellcolor[HTML]{FCBF7B}0.485 & \cellcolor[HTML]{FDD07E}0.557 & \cellcolor[HTML]{FDD680}0.584 & \cellcolor[HTML]{FDC87D}0.524 & \cellcolor[HTML]{DDE182}0.741 & \cellcolor[HTML]{B3D580}0.822 & \cellcolor[HTML]{F4E884}0.696 & \cellcolor[HTML]{FCC47C}0.506 & \cellcolor[HTML]{FA9272}0.286 & \cellcolor[HTML]{FED880}0.594 &  & \cellcolor[HTML]{FEE081}0.627 \\
DaSiamRPN & \cellcolor[HTML]{8ACA7E}0.902 & \cellcolor[HTML]{ABD380}0.837 & \cellcolor[HTML]{D5DF82}0.756 & \cellcolor[HTML]{ECE683}0.711 & \cellcolor[HTML]{F7E984}0.690 & \cellcolor[HTML]{FDD47F}0.577 & \cellcolor[HTML]{FEE783}0.658 & \cellcolor[HTML]{FBA276}0.356 &  & \cellcolor[HTML]{FCBF7B}0.484 & \cellcolor[HTML]{FDD17F}0.561 & \cellcolor[HTML]{FDD680}0.585 & \cellcolor[HTML]{FDC97D}0.526 & \cellcolor[HTML]{DCE182}0.743 & \cellcolor[HTML]{B5D680}0.818 & \cellcolor[HTML]{F5E984}0.693 & \cellcolor[HTML]{FDC97D}0.527 & \cellcolor[HTML]{FA9172}0.285 & \cellcolor[HTML]{FDD67F}0.582 &  & \cellcolor[HTML]{FEE081}0.627 \\
SiamRPNPP & \cellcolor[HTML]{8ACA7E}0.902 & \cellcolor[HTML]{A4D17F}0.851 & \cellcolor[HTML]{CBDC81}0.775 & \cellcolor[HTML]{D8E082}0.751 & \cellcolor[HTML]{E5E483}0.725 & \cellcolor[HTML]{D4DF82}0.759 & \cellcolor[HTML]{F0E784}0.704 & \cellcolor[HTML]{FCB97A}0.460 &  & \cellcolor[HTML]{FDCE7E}0.550 & \cellcolor[HTML]{FEDA80}0.602 & \cellcolor[HTML]{FEE282}0.638 & \cellcolor[HTML]{FDD780}0.587 & \cellcolor[HTML]{C5DB81}0.787 & \cellcolor[HTML]{A9D27F}0.842 & \cellcolor[HTML]{DEE283}0.739 & \cellcolor[HTML]{FDCE7E}0.549 & \cellcolor[HTML]{FBA977}0.388 & \cellcolor[HTML]{FEE883}0.660 &  & \cellcolor[HTML]{FBEA84}0.682 \\
SPLT & \cellcolor[HTML]{A1D07F}0.856 & \cellcolor[HTML]{E6E483}0.724 & \cellcolor[HTML]{FEDB80}0.604 & \cellcolor[HTML]{FDD67F}0.582 & \cellcolor[HTML]{FEE983}0.668 & \cellcolor[HTML]{DAE182}0.747 & \cellcolor[HTML]{FEE182}0.631 & \cellcolor[HTML]{FBB178}0.422 &  & \cellcolor[HTML]{FDC87D}0.521 & \cellcolor[HTML]{FED880}0.592 & \cellcolor[HTML]{FEDB80}0.605 & \cellcolor[HTML]{FDD57F}0.578 & \cellcolor[HTML]{FEE683}0.653 & \cellcolor[HTML]{C5DB81}0.787 & \cellcolor[HTML]{F3E884}0.698 & \cellcolor[HTML]{FCC37C}0.499 & \cellcolor[HTML]{FA9172}0.286 & \cellcolor[HTML]{FEDB81}0.605 &  & \cellcolor[HTML]{FEDD81}0.614 \\
SiamDW & \cellcolor[HTML]{70C27C}0.952 & \cellcolor[HTML]{A6D27F}0.847 & \cellcolor[HTML]{EDE683}0.709 & \cellcolor[HTML]{FEEA83}0.671 & \cellcolor[HTML]{F5E884}0.694 & \cellcolor[HTML]{FDD57F}0.580 & \cellcolor[HTML]{FEE382}0.640 & \cellcolor[HTML]{FBA376}0.363 &  & \cellcolor[HTML]{FCC17C}0.492 & \cellcolor[HTML]{FDD07E}0.558 & \cellcolor[HTML]{FDD680}0.585 & \cellcolor[HTML]{FDCA7D}0.533 & \cellcolor[HTML]{C7DB81}0.784 & \cellcolor[HTML]{B0D580}0.827 & \cellcolor[HTML]{E9E583}0.717 & \cellcolor[HTML]{FCC57C}0.508 & \cellcolor[HTML]{FA9874}0.314 & \cellcolor[HTML]{FDD780}0.588 &  & \cellcolor[HTML]{FEE182}0.631 \\
SiamCAR & \cellcolor[HTML]{8ECB7E}0.895 & \cellcolor[HTML]{C6DB81}0.786 & \cellcolor[HTML]{F7E984}0.689 & \cellcolor[HTML]{FAEA84}0.684 & \cellcolor[HTML]{E3E383}0.729 & \cellcolor[HTML]{FDD17F}0.561 & \cellcolor[HTML]{FDEB84}0.679 & \cellcolor[HTML]{FA8F72}0.275 &  & \cellcolor[HTML]{FBAE78}0.411 & \cellcolor[HTML]{FCB579}0.440 & \cellcolor[HTML]{FDCA7D}0.532 & \cellcolor[HTML]{FBB178}0.421 & \cellcolor[HTML]{E6E483}0.722 & \cellcolor[HTML]{C9DC81}0.779 & \cellcolor[HTML]{FAEA84}0.683 & \cellcolor[HTML]{FDC77D}0.521 & \cellcolor[HTML]{FA9A74}0.321 & \cellcolor[HTML]{FDD37F}0.569 &  & \cellcolor[HTML]{FED880}0.594 \\
SiamFCPP & \cellcolor[HTML]{86C87D}0.909 & \cellcolor[HTML]{A0D07F}0.859 & \cellcolor[HTML]{C9DC81}0.779 & \cellcolor[HTML]{D6E082}0.754 & \cellcolor[HTML]{D8E082}0.749 & \cellcolor[HTML]{FCB579}0.439 & \cellcolor[HTML]{FEE182}0.633 & \cellcolor[HTML]{FA9974}0.320 &  & \cellcolor[HTML]{FCC07B}0.487 & \cellcolor[HTML]{FDCB7D}0.536 & \cellcolor[HTML]{FDD27F}0.567 & \cellcolor[HTML]{FCC37C}0.503 & \cellcolor[HTML]{B6D680}0.816 & \cellcolor[HTML]{97CD7E}0.877 & \cellcolor[HTML]{EBE683}0.713 & \cellcolor[HTML]{FCBD7B}0.474 & \cellcolor[HTML]{FA9673}0.305 & \cellcolor[HTML]{FDD780}0.587 &  & \cellcolor[HTML]{FEE082}0.628 \\
Ocean & \cellcolor[HTML]{90CB7E}0.890 & \cellcolor[HTML]{9FD07F}0.861 & \cellcolor[HTML]{D0DE82}0.767 & \cellcolor[HTML]{DCE182}0.742 & \cellcolor[HTML]{CBDC81}0.775 & \cellcolor[HTML]{FEDB80}0.604 & \cellcolor[HTML]{F8E984}0.689 & \cellcolor[HTML]{FBAC78}0.403 &  & \cellcolor[HTML]{FDCB7D}0.536 & \cellcolor[HTML]{FEDB81}0.606 & \cellcolor[HTML]{FEE182}0.631 & \cellcolor[HTML]{FDD57F}0.580 & \cellcolor[HTML]{BAD881}0.808 & \cellcolor[HTML]{97CD7E}0.876 & \cellcolor[HTML]{D9E082}0.747 & \cellcolor[HTML]{FDCB7D}0.534 & \cellcolor[HTML]{FA9974}0.320 & \cellcolor[HTML]{FEE382}0.641 &  & \cellcolor[HTML]{FEE983}0.667 \\
SiamRCNN & \cellcolor[HTML]{6CC17C}0.960 & \cellcolor[HTML]{ABD380}0.838 & \cellcolor[HTML]{D8E082}0.749 & \cellcolor[HTML]{D6DF82}0.755 & \cellcolor[HTML]{C2DA81}0.792 & \cellcolor[HTML]{C5DB81}0.787 & \cellcolor[HTML]{CEDD82}0.769 & \cellcolor[HTML]{FDD780}0.587 &  & \cellcolor[HTML]{FEE883}0.661 & \cellcolor[HTML]{EBE683}0.713 & \cellcolor[HTML]{DFE283}0.737 & \cellcolor[HTML]{EFE784}0.706 & \cellcolor[HTML]{B8D780}0.813 & \cellcolor[HTML]{9FD07F}0.862 & \cellcolor[HTML]{C1DA81}0.794 & \cellcolor[HTML]{FEE582}0.648 & \cellcolor[HTML]{FDCE7E}0.550 & \cellcolor[HTML]{D0DE82}0.765 &  & \cellcolor[HTML]{D8E082}0.749 \\\hline
ATOM & \cellcolor[HTML]{7BC57D}0.931 & \cellcolor[HTML]{B9D780}0.810 & \cellcolor[HTML]{C4DA81}0.790 & \cellcolor[HTML]{D2DE82}0.762 & \cellcolor[HTML]{CDDD82}0.771 & \cellcolor[HTML]{FDCF7E}0.555 & \cellcolor[HTML]{F9E984}0.687 & \cellcolor[HTML]{FA9D75}0.336 &  & \cellcolor[HTML]{FCBC7B}0.472 & \cellcolor[HTML]{FDC97D}0.529 & \cellcolor[HTML]{FED980}0.597 & \cellcolor[HTML]{FCC37C}0.502 & \cellcolor[HTML]{CFDD82}0.768 & \cellcolor[HTML]{ACD480}0.835 & \cellcolor[HTML]{E3E383}0.728 & \cellcolor[HTML]{FEDC81}0.609 & \cellcolor[HTML]{FBAA77}0.391 & \cellcolor[HTML]{FEE081}0.626 &  & \cellcolor[HTML]{FEE582}0.650 \\
DiMP & \cellcolor[HTML]{78C47D}0.937 & \cellcolor[HTML]{98CE7F}0.875 & \cellcolor[HTML]{CEDD82}0.769 & \cellcolor[HTML]{D2DE82}0.762 & \cellcolor[HTML]{BED981}0.801 & \cellcolor[HTML]{F2E884}0.699 & \cellcolor[HTML]{DDE283}0.740 & \cellcolor[HTML]{FCB379}0.433 &  & \cellcolor[HTML]{FDCC7E}0.541 & \cellcolor[HTML]{FEDB80}0.604 & \cellcolor[HTML]{FEE883}0.662 & \cellcolor[HTML]{FDD680}0.583 & \cellcolor[HTML]{B4D680}0.821 & \cellcolor[HTML]{9ACE7F}0.871 & \cellcolor[HTML]{CCDD82}0.773 & \cellcolor[HTML]{FAEA84}0.683 & \cellcolor[HTML]{FCB97A}0.457 & \cellcolor[HTML]{FEEB84}0.676 &  & \cellcolor[HTML]{EFE784}0.705 \\
PrDiMP & \cellcolor[HTML]{73C37C}0.947 & \cellcolor[HTML]{A6D27F}0.847 & \cellcolor[HTML]{D0DE82}0.766 & \cellcolor[HTML]{D7E082}0.751 & \cellcolor[HTML]{BED981}0.801 & \cellcolor[HTML]{FFEB84}0.674 & \cellcolor[HTML]{D2DE82}0.763 & \cellcolor[HTML]{FCC07B}0.486 &  & \cellcolor[HTML]{FEDC81}0.612 & \cellcolor[HTML]{FEE983}0.667 & \cellcolor[HTML]{EBE683}0.713 & \cellcolor[HTML]{FEE783}0.657 & \cellcolor[HTML]{B2D580}0.825 & \cellcolor[HTML]{8FCB7E}0.891 & \cellcolor[HTML]{B7D780}0.814 & \cellcolor[HTML]{F1E784}0.702 & \cellcolor[HTML]{FCBE7B}0.478 & \cellcolor[HTML]{E9E583}0.716 &  & \cellcolor[HTML]{E3E383}0.728 \\
SuperDiMP & \cellcolor[HTML]{64BF7C}0.976 & \cellcolor[HTML]{92CC7E}0.886 & \cellcolor[HTML]{BAD780}0.809 & \cellcolor[HTML]{C8DB81}0.782 & \cellcolor[HTML]{AED480}0.831 & \cellcolor[HTML]{D3DF82}0.760 & \cellcolor[HTML]{C4DA81}0.790 & \cellcolor[HTML]{FCBF7B}0.486 &  & \cellcolor[HTML]{FEDB81}0.607 & \cellcolor[HTML]{FEE883}0.663 & \cellcolor[HTML]{E9E583}0.718 & \cellcolor[HTML]{FEE683}0.654 & \cellcolor[HTML]{ABD380}0.837 & \cellcolor[HTML]{8ECB7E}0.894 & \cellcolor[HTML]{B3D580}0.823 & \cellcolor[HTML]{DBE182}0.744 & \cellcolor[HTML]{FCC47C}0.504 & \cellcolor[HTML]{E9E583}0.716 &  & \cellcolor[HTML]{D9E082}0.749 \\
KeepTrack & \cellcolor[HTML]{63BE7B}0.977 & \cellcolor[HTML]{8CCA7E}0.898 & \cellcolor[HTML]{AFD480}0.831 & \cellcolor[HTML]{CDDD82}0.772 & \cellcolor[HTML]{A4D17F}0.852 & \cellcolor[HTML]{ADD480}0.833 & \cellcolor[HTML]{B0D580}0.827 & \cellcolor[HTML]{FDCF7E}0.551 &  & \cellcolor[HTML]{FEE783}0.660 & \cellcolor[HTML]{ECE683}0.711 & \cellcolor[HTML]{D2DE82}0.762 & \cellcolor[HTML]{EFE784}0.706 & \cellcolor[HTML]{ABD380}0.838 & \cellcolor[HTML]{8FCB7E}0.892 & \cellcolor[HTML]{ABD380}0.838 & \cellcolor[HTML]{CDDD82}0.772 & \cellcolor[HTML]{FDD37F}0.573 & \cellcolor[HTML]{D1DE82}0.764 &  & \cellcolor[HTML]{C8DC81}0.781 \\\hline
GlobalTrack & \cellcolor[HTML]{A9D380}0.841 & \cellcolor[HTML]{F4E884}0.696 & \cellcolor[HTML]{FED980}0.595 & \cellcolor[HTML]{FDD680}0.584 & \cellcolor[HTML]{E1E383}0.732 & \cellcolor[HTML]{FEEA83}0.672 & \cellcolor[HTML]{FDEB84}0.678 & \cellcolor[HTML]{FCC47C}0.505 &  & \cellcolor[HTML]{FDD780}0.587 & \cellcolor[HTML]{FEE583}0.651 & \cellcolor[HTML]{FEE983}0.666 & \cellcolor[HTML]{FEE282}0.638 & \cellcolor[HTML]{EFE784}0.704 & \cellcolor[HTML]{B8D780}0.813 & \cellcolor[HTML]{E0E383}0.734 & \cellcolor[HTML]{FDCF7E}0.552 & \cellcolor[HTML]{FCBD7B}0.476 & \cellcolor[HTML]{F8E984}0.687 &  & \cellcolor[HTML]{FEE783}0.656 \\
KYS & \cellcolor[HTML]{8CCA7E}0.899 & \cellcolor[HTML]{98CE7F}0.874 & \cellcolor[HTML]{A1D07F}0.857 & \cellcolor[HTML]{CDDD82}0.772 & \cellcolor[HTML]{C0D981}0.797 & \cellcolor[HTML]{FED980}0.598 & \cellcolor[HTML]{F2E884}0.699 & \cellcolor[HTML]{FBA576}0.370 &  & \cellcolor[HTML]{FCC57C}0.511 & \cellcolor[HTML]{FDD37F}0.572 & \cellcolor[HTML]{FEE081}0.626 & \cellcolor[HTML]{FDCD7E}0.546 & \cellcolor[HTML]{B3D580}0.822 & \cellcolor[HTML]{9CCF7F}0.866 & \cellcolor[HTML]{D1DE82}0.764 & \cellcolor[HTML]{FEE883}0.662 & \cellcolor[HTML]{FCB679}0.446 & \cellcolor[HTML]{FEE783}0.656 &  & \cellcolor[HTML]{F9EA84}0.685 \\
MixFormer & \cellcolor[HTML]{75C47D}0.943 & \cellcolor[HTML]{93CC7E}0.884 & \cellcolor[HTML]{CADC81}0.777 & \cellcolor[HTML]{CADC81}0.776 & \cellcolor[HTML]{AAD380}0.841 & \cellcolor[HTML]{ACD480}0.835 & \cellcolor[HTML]{AED480}0.833 & \cellcolor[HTML]{FEDB81}0.608 &  & \cellcolor[HTML]{E6E483}0.723 & \cellcolor[HTML]{D1DE82}0.764 & \cellcolor[HTML]{BCD881}0.804 & \cellcolor[HTML]{D1DE82}0.765 & \cellcolor[HTML]{A0D07F}0.859 & \cellcolor[HTML]{7DC67D}0.926 & \cellcolor[HTML]{9CCF7F}0.867 & \cellcolor[HTML]{C7DB81}0.783 & \cellcolor[HTML]{FDD680}0.586 & \cellcolor[HTML]{C4DA81}0.790 &  & \cellcolor[HTML]{BFD981}0.798 \\
			\botrule
			\end{tabular}
			\begin{tablenotes}
				\item [1] \textit{
				$e_1$-OTB2015 \cite{OTB2015}. 
				$e_2$-VOT2016 \cite{VOT2016}.
				$e_{3}$-VOT2018 \cite{VOT2018}.
				$e_{4}$-VOT2019 \cite{VOT2019}.
				$e_{5}$-GOT-10k \cite{GOT-10k}.
				$e_{6}$-VOTLT2019 \cite{VOT2019}.
				$e_{7}$-LaSOT \cite{LaSOT}.
				$e_{8}$-VideoCube \cite{GIT}.
				$c_1$-abnormal ratio.
				$c_2$-abnormal scale.
				$c_3$-abnormal illumination.
				$c_4$-blur bbox.
				$c_5$-delta ratio.
				$c_6$-delta scale.
				$c_7$-delta illumination.
				$c_8$-delta blur bbox.
				$c_9$-fast motion.
				$c_{10}$-low correlation coefficient.
				}
				\item[2] \textit{
				The background color of cells indicates the score, with red indicating a low score and green indicating a high score.
				}
				\end{tablenotes}
		\label{tab:npre-l}
	\end{center}
	\end{center}
	\end{sidewaystable*}

\clearpage
\begin{sidewaystable*}[htbp!]
	\sidewaystablefn%
	\begin{center}
	\caption{Performance of 20 representative trackers on all sub-spaces, based on success score, weighted by sequences' length. 
	}
	\renewcommand{\arraystretch}{1.2}
	\begin{center}
		\tabcolsep=0.1cm
		\small
		\begin{tabular}{llllllllllllllllllllll}
			\toprule
			\multirow{2}{*}{\textbf{Trackers}} & \multicolumn{8}{l}{\textbf{Normal space}} &  & \multicolumn{10}{l}{\textbf{Challenging space}} & & \multirow{2}{*}{\textbf{Mean}} \\ \cline{2-9} \cline{11-20}
				& $e_1$ & $e_2$ & $e_3$ & $e_4$ & $e_5$ & $e_6$ & $e_7$ & $e_8$ &  & $c_1$ & $c_2$ & $c_3$ & $c_4$ & $c_5$ & $c_6$ & $c_7$ & $c_8$ & $c_9$ & $c_{10}$ &  \\ \midrule
				KCF & \cellcolor[HTML]{FDD17F}0.386 & \cellcolor[HTML]{FBAC77}0.267 & \cellcolor[HTML]{FA9A74}0.210 & \cellcolor[HTML]{FA9473}0.192 & \cellcolor[HTML]{FA9E75}0.223 & \cellcolor[HTML]{F8766D}0.095 & \cellcolor[HTML]{F98B71}0.162 & \cellcolor[HTML]{F8696B}0.054 &  & \cellcolor[HTML]{F98570}0.144 & \cellcolor[HTML]{F98670}0.146 & \cellcolor[HTML]{F98D72}0.170 & \cellcolor[HTML]{F97F6F}0.124 & \cellcolor[HTML]{FCBB7A}0.317 & \cellcolor[HTML]{FCBE7B}0.326 & \cellcolor[HTML]{FBA376}0.240 & \cellcolor[HTML]{F98770}0.149 & \cellcolor[HTML]{F86F6C}0.072 & \cellcolor[HTML]{FA9673}0.197 &  & \cellcolor[HTML]{FA9573}0.193 \\
ECO & \cellcolor[HTML]{8BCA7E}0.698 & \cellcolor[HTML]{E1E383}0.527 & \cellcolor[HTML]{FEE983}0.462 & \cellcolor[HTML]{FDD57F}0.398 & \cellcolor[HTML]{FEDE81}0.427 & \cellcolor[HTML]{FBAE78}0.273 & \cellcolor[HTML]{FDCB7D}0.367 & \cellcolor[HTML]{F8756D}0.094 &  & \cellcolor[HTML]{FA9E75}0.223 & \cellcolor[HTML]{FA9E75}0.222 & \cellcolor[HTML]{FCB379}0.290 & \cellcolor[HTML]{FA9874}0.205 & \cellcolor[HTML]{F6E984}0.486 & \cellcolor[HTML]{F4E884}0.489 & \cellcolor[HTML]{FEDB81}0.418 & \cellcolor[HTML]{FCC57C}0.347 & \cellcolor[HTML]{FA9D75}0.221 & \cellcolor[HTML]{FCC27C}0.339 &  & \cellcolor[HTML]{FDC97D}0.360 \\ \hline
SiamFC & \cellcolor[HTML]{B2D580}0.621 & \cellcolor[HTML]{FEE182}0.438 & \cellcolor[HTML]{FDCB7D}0.367 & \cellcolor[HTML]{FDC67C}0.350 & \cellcolor[HTML]{FCC47C}0.345 & \cellcolor[HTML]{FA9D75}0.219 & \cellcolor[HTML]{FCBA7A}0.313 & \cellcolor[HTML]{F8696B}0.052 &  & \cellcolor[HTML]{F98370}0.138 & \cellcolor[HTML]{F98370}0.136 & \cellcolor[HTML]{FA9974}0.208 & \cellcolor[HTML]{F97F6F}0.125 & \cellcolor[HTML]{FDD37F}0.392 & \cellcolor[HTML]{FCBA7A}0.314 & \cellcolor[HTML]{FBAE78}0.274 & \cellcolor[HTML]{FCB379}0.291 & \cellcolor[HTML]{F98D71}0.168 & \cellcolor[HTML]{FBA476}0.242 &  & \cellcolor[HTML]{FBAF78}0.278 \\
SiamRPN & \cellcolor[HTML]{96CD7E}0.676 & \cellcolor[HTML]{CFDD82}0.563 & \cellcolor[HTML]{F5E984}0.487 & \cellcolor[HTML]{FEE683}0.453 & \cellcolor[HTML]{FEE683}0.454 & \cellcolor[HTML]{FDD880}0.408 & \cellcolor[HTML]{FDEB84}0.471 & \cellcolor[HTML]{FA9E75}0.222 &  & \cellcolor[HTML]{FCBD7B}0.322 & \cellcolor[HTML]{FDC97D}0.361 & \cellcolor[HTML]{FDD47F}0.394 & \cellcolor[HTML]{FCC37C}0.340 & \cellcolor[HTML]{EDE683}0.505 & \cellcolor[HTML]{DAE182}0.542 & \cellcolor[HTML]{FEEB84}0.471 & \cellcolor[HTML]{FDC67D}0.352 & \cellcolor[HTML]{FA9473}0.192 & \cellcolor[HTML]{FDD680}0.403 &  & \cellcolor[HTML]{FEDD81}0.423 \\
DaSiamRPN & \cellcolor[HTML]{9ACE7F}0.668 & \cellcolor[HTML]{CDDD82}0.566 & \cellcolor[HTML]{F2E884}0.493 & \cellcolor[HTML]{FEE983}0.462 & \cellcolor[HTML]{FEE482}0.448 & \cellcolor[HTML]{FEDA80}0.415 & \cellcolor[HTML]{FEE983}0.464 & \cellcolor[HTML]{FA9C74}0.216 &  & \cellcolor[HTML]{FCBD7B}0.322 & \cellcolor[HTML]{FDCA7D}0.365 & \cellcolor[HTML]{FDD47F}0.395 & \cellcolor[HTML]{FCC37C}0.341 & \cellcolor[HTML]{ECE683}0.505 & \cellcolor[HTML]{DBE182}0.540 & \cellcolor[HTML]{FFEB84}0.469 & \cellcolor[HTML]{FDCA7D}0.362 & \cellcolor[HTML]{FA9473}0.191 & \cellcolor[HTML]{FDD47F}0.396 &  & \cellcolor[HTML]{FEDD81}0.423 \\
SiamRPNPP & \cellcolor[HTML]{92CC7E}0.684 & \cellcolor[HTML]{C1DA81}0.590 & \cellcolor[HTML]{E6E483}0.518 & \cellcolor[HTML]{EEE683}0.502 & \cellcolor[HTML]{F6E984}0.487 & \cellcolor[HTML]{D8E082}0.545 & \cellcolor[HTML]{E2E383}0.525 & \cellcolor[HTML]{FCB87A}0.307 &  & \cellcolor[HTML]{FDCF7E}0.378 & \cellcolor[HTML]{FDD780}0.407 & \cellcolor[HTML]{FEE582}0.448 & \cellcolor[HTML]{FDD47F}0.397 & \cellcolor[HTML]{D4DF82}0.554 & \cellcolor[HTML]{C6DB81}0.581 & \cellcolor[HTML]{E5E483}0.519 & \cellcolor[HTML]{FDD37F}0.392 & \cellcolor[HTML]{FBAD78}0.271 & \cellcolor[HTML]{FEEA83}0.467 &  & \cellcolor[HTML]{FBEA84}0.476 \\
SPLT & \cellcolor[HTML]{BBD881}0.603 & \cellcolor[HTML]{FEEA83}0.467 & \cellcolor[HTML]{FDD17F}0.385 & \cellcolor[HTML]{FDCB7D}0.366 & \cellcolor[HTML]{FEDF81}0.432 & \cellcolor[HTML]{F3E884}0.493 & \cellcolor[HTML]{FEDE81}0.427 & \cellcolor[HTML]{FBA977}0.257 &  & \cellcolor[HTML]{FCC07B}0.333 & \cellcolor[HTML]{FDCB7D}0.367 & \cellcolor[HTML]{FDD37F}0.392 & \cellcolor[HTML]{FDC97D}0.362 & \cellcolor[HTML]{FDD780}0.405 & \cellcolor[HTML]{F4E884}0.489 & \cellcolor[HTML]{FEE683}0.453 & \cellcolor[HTML]{FCBF7B}0.330 & \cellcolor[HTML]{FA9473}0.191 & \cellcolor[HTML]{FDD37F}0.392 &  & \cellcolor[HTML]{FDD47F}0.397 \\
SiamDW & \cellcolor[HTML]{8ACA7E}0.699 & \cellcolor[HTML]{DFE283}0.531 & \cellcolor[HTML]{FEE082}0.435 & \cellcolor[HTML]{FED980}0.413 & \cellcolor[HTML]{FDD680}0.403 & \cellcolor[HTML]{FCBF7B}0.327 & \cellcolor[HTML]{FDCA7D}0.363 & \cellcolor[HTML]{F97F6F}0.124 &  & \cellcolor[HTML]{FBA376}0.238 & \cellcolor[HTML]{FBA376}0.239 & \cellcolor[HTML]{FCB77A}0.302 & \cellcolor[HTML]{FA9E75}0.224 & \cellcolor[HTML]{EDE683}0.505 & \cellcolor[HTML]{EFE784}0.501 & \cellcolor[HTML]{FEDC81}0.420 & \cellcolor[HTML]{FCBA7A}0.313 & \cellcolor[HTML]{FA9773}0.201 & \cellcolor[HTML]{FCC47C}0.345 &  & \cellcolor[HTML]{FDCB7D}0.366 \\
SiamCAR & \cellcolor[HTML]{99CE7F}0.669 & \cellcolor[HTML]{E1E383}0.527 & \cellcolor[HTML]{FEE182}0.436 & \cellcolor[HTML]{FEE081}0.433 & \cellcolor[HTML]{EFE784}0.500 & \cellcolor[HTML]{FDD27F}0.388 & \cellcolor[HTML]{F9EA84}0.480 & \cellcolor[HTML]{F98470}0.139 &  & \cellcolor[HTML]{FBA275}0.235 & \cellcolor[HTML]{FBA476}0.243 & \cellcolor[HTML]{FCC27C}0.337 & \cellcolor[HTML]{FBA376}0.239 & \cellcolor[HTML]{FFEB84}0.469 & \cellcolor[HTML]{F6E984}0.486 & \cellcolor[HTML]{FEE282}0.442 & \cellcolor[HTML]{FDC87D}0.356 & \cellcolor[HTML]{FA9F75}0.225 & \cellcolor[HTML]{FDCD7E}0.374 &  & \cellcolor[HTML]{FDD27F}0.388 \\
SiamFCPP & \cellcolor[HTML]{9ACE7F}0.669 & \cellcolor[HTML]{CEDD82}0.565 & \cellcolor[HTML]{F3E884}0.492 & \cellcolor[HTML]{FBEA84}0.476 & \cellcolor[HTML]{E2E383}0.526 & \cellcolor[HTML]{FCBB7A}0.316 & \cellcolor[HTML]{FFEB84}0.468 & \cellcolor[HTML]{FA9673}0.198 &  & \cellcolor[HTML]{FCBF7B}0.329 & \cellcolor[HTML]{FDC87D}0.359 & \cellcolor[HTML]{FDD37F}0.393 & \cellcolor[HTML]{FCC27C}0.337 & \cellcolor[HTML]{B0D580}0.623 & \cellcolor[HTML]{ADD480}0.630 & \cellcolor[HTML]{EDE683}0.504 & \cellcolor[HTML]{FCC37C}0.340 & \cellcolor[HTML]{FBA175}0.234 & \cellcolor[HTML]{FEE081}0.432 &  & \cellcolor[HTML]{FEE182}0.438 \\
Ocean & \cellcolor[HTML]{9FD07F}0.657 & \cellcolor[HTML]{C8DB81}0.577 & \cellcolor[HTML]{EFE784}0.500 & \cellcolor[HTML]{F8E984}0.481 & \cellcolor[HTML]{D4DF82}0.554 & \cellcolor[HTML]{FEDA80}0.414 & \cellcolor[HTML]{E9E583}0.511 & \cellcolor[HTML]{FBA977}0.257 &  & \cellcolor[HTML]{FDC97D}0.359 & \cellcolor[HTML]{FDD680}0.401 & \cellcolor[HTML]{FEE182}0.437 & \cellcolor[HTML]{FDD27F}0.388 & \cellcolor[HTML]{BAD780}0.605 & \cellcolor[HTML]{B3D580}0.618 & \cellcolor[HTML]{E3E383}0.523 & \cellcolor[HTML]{FDD07E}0.382 & \cellcolor[HTML]{FBA376}0.241 & \cellcolor[HTML]{FEE983}0.463 &  & \cellcolor[HTML]{FEEA83}0.465 \\
SiamRCNN & \cellcolor[HTML]{73C37C}0.744 & \cellcolor[HTML]{C3DA81}0.587 & \cellcolor[HTML]{EBE583}0.508 & \cellcolor[HTML]{EAE583}0.510 & \cellcolor[HTML]{A3D17F}0.651 & \cellcolor[HTML]{BDD881}0.599 & \cellcolor[HTML]{A9D27F}0.639 & \cellcolor[HTML]{FBEA84}0.476 &  & \cellcolor[HTML]{D9E182}0.542 & \cellcolor[HTML]{C4DA81}0.585 & \cellcolor[HTML]{BAD780}0.604 & \cellcolor[HTML]{C9DC81}0.575 & \cellcolor[HTML]{8DCB7E}0.693 & \cellcolor[HTML]{82C77D}0.714 & \cellcolor[HTML]{A1D07F}0.654 & \cellcolor[HTML]{E9E583}0.511 & \cellcolor[HTML]{FEE382}0.442 & \cellcolor[HTML]{ABD380}0.635 &  & \cellcolor[HTML]{C0D981}0.593 \\\hline
ATOM & \cellcolor[HTML]{8BCA7E}0.697 & \cellcolor[HTML]{D8E082}0.546 & \cellcolor[HTML]{E0E283}0.530 & \cellcolor[HTML]{ECE683}0.505 & \cellcolor[HTML]{DAE182}0.541 & \cellcolor[HTML]{FDD07E}0.383 & \cellcolor[HTML]{F1E784}0.496 & \cellcolor[HTML]{FA9773}0.201 &  & \cellcolor[HTML]{FCB97A}0.308 & \cellcolor[HTML]{FCC07B}0.333 & \cellcolor[HTML]{FDD680}0.403 & \cellcolor[HTML]{FCBD7B}0.322 & \cellcolor[HTML]{E2E383}0.526 & \cellcolor[HTML]{D3DF82}0.556 & \cellcolor[HTML]{EEE784}0.501 & \cellcolor[HTML]{FEE282}0.442 & \cellcolor[HTML]{FBB078}0.280 & \cellcolor[HTML]{FEE081}0.433 &  & \cellcolor[HTML]{FEE382}0.445 \\
DiMP & \cellcolor[HTML]{87C97E}0.706 & \cellcolor[HTML]{BDD881}0.599 & \cellcolor[HTML]{E6E483}0.518 & \cellcolor[HTML]{E8E583}0.514 & \cellcolor[HTML]{BED981}0.596 & \cellcolor[HTML]{F1E784}0.497 & \cellcolor[HTML]{CEDD82}0.565 & \cellcolor[HTML]{FCB579}0.298 &  & \cellcolor[HTML]{FDD27F}0.388 & \cellcolor[HTML]{FEDE81}0.429 & \cellcolor[HTML]{F7E984}0.484 & \cellcolor[HTML]{FEDA80}0.415 & \cellcolor[HTML]{B4D680}0.616 & \cellcolor[HTML]{A9D380}0.638 & \cellcolor[HTML]{CCDD82}0.569 & \cellcolor[HTML]{EDE683}0.504 & \cellcolor[HTML]{FDC67C}0.349 & \cellcolor[HTML]{ECE683}0.506 &  & \cellcolor[HTML]{EAE583}0.511 \\
PrDiMP & \cellcolor[HTML]{7DC67D}0.725 & \cellcolor[HTML]{C3DA81}0.587 & \cellcolor[HTML]{E3E383}0.524 & \cellcolor[HTML]{EAE583}0.510 & \cellcolor[HTML]{B4D680}0.617 & \cellcolor[HTML]{F3E884}0.492 & \cellcolor[HTML]{C1D981}0.591 & \cellcolor[HTML]{FCC17B}0.334 &  & \cellcolor[HTML]{FEE182}0.436 & \cellcolor[HTML]{FEEA83}0.466 & \cellcolor[HTML]{E3E383}0.524 & \cellcolor[HTML]{FEEA83}0.466 & \cellcolor[HTML]{A8D27F}0.641 & \cellcolor[HTML]{A2D07F}0.652 & \cellcolor[HTML]{BAD881}0.604 & \cellcolor[HTML]{E1E383}0.527 & \cellcolor[HTML]{FDCE7E}0.376 & \cellcolor[HTML]{DAE182}0.541 &  & \cellcolor[HTML]{DEE283}0.534 \\
SuperDiMP & \cellcolor[HTML]{74C37C}0.743 & \cellcolor[HTML]{B6D680}0.613 & \cellcolor[HTML]{D7E082}0.547 & \cellcolor[HTML]{DEE283}0.533 & \cellcolor[HTML]{A1D07F}0.654 & \cellcolor[HTML]{CDDD82}0.567 & \cellcolor[HTML]{ABD380}0.634 & \cellcolor[HTML]{FDC87D}0.356 &  & \cellcolor[HTML]{FEE883}0.460 & \cellcolor[HTML]{EEE683}0.502 & \cellcolor[HTML]{D2DE82}0.558 & \cellcolor[HTML]{F0E784}0.498 & \cellcolor[HTML]{9DCF7F}0.661 & \cellcolor[HTML]{8CCA7E}0.696 & \cellcolor[HTML]{A7D27F}0.643 & \cellcolor[HTML]{CBDC81}0.570 & \cellcolor[HTML]{FDD57F}0.399 & \cellcolor[HTML]{CEDD82}0.564 &  & \cellcolor[HTML]{CDDD82}0.566 \\
KeepTrack & \cellcolor[HTML]{70C27C}0.750 & \cellcolor[HTML]{ADD480}0.630 & \cellcolor[HTML]{CBDC81}0.572 & \cellcolor[HTML]{E0E283}0.530 & \cellcolor[HTML]{98CE7F}0.671 & \cellcolor[HTML]{AFD480}0.627 & \cellcolor[HTML]{9CCF7F}0.664 & \cellcolor[HTML]{FDD780}0.405 &  & \cellcolor[HTML]{EFE784}0.501 & \cellcolor[HTML]{DBE182}0.539 & \cellcolor[HTML]{C1D981}0.591 & \cellcolor[HTML]{DCE182}0.538 & \cellcolor[HTML]{9CCF7F}0.664 & \cellcolor[HTML]{8DCA7E}0.694 & \cellcolor[HTML]{A1D07F}0.654 & \cellcolor[HTML]{C1DA81}0.590 & \cellcolor[HTML]{FEE683}0.453 & \cellcolor[HTML]{BAD780}0.605 &  & \cellcolor[HTML]{C0D981}0.593 \\\hline
GlobalTrack & \cellcolor[HTML]{A4D17F}0.648 & \cellcolor[HTML]{F3E884}0.493 & \cellcolor[HTML]{FEDA80}0.414 & \cellcolor[HTML]{FDD680}0.402 & \cellcolor[HTML]{CEDD82}0.566 & \cellcolor[HTML]{F9EA84}0.480 & \cellcolor[HTML]{E1E383}0.528 & \cellcolor[HTML]{FDC97D}0.359 &  & \cellcolor[HTML]{FEDF81}0.431 & \cellcolor[HTML]{FEEA83}0.464 & \cellcolor[HTML]{F1E784}0.496 & \cellcolor[HTML]{FEE883}0.460 & \cellcolor[HTML]{CDDD82}0.568 & \cellcolor[HTML]{B6D680}0.613 & \cellcolor[HTML]{D6DF82}0.550 & \cellcolor[HTML]{FED980}0.411 & \cellcolor[HTML]{FDCC7E}0.369 & \cellcolor[HTML]{E1E383}0.527 &  & \cellcolor[HTML]{F5E984}0.488 \\
KYS & \cellcolor[HTML]{95CD7E}0.678 & \cellcolor[HTML]{BBD881}0.602 & \cellcolor[HTML]{C8DB81}0.578 & \cellcolor[HTML]{E7E483}0.516 & \cellcolor[HTML]{BED981}0.597 & \cellcolor[HTML]{FEDF81}0.431 & \cellcolor[HTML]{DDE283}0.535 & \cellcolor[HTML]{FBA676}0.249 &  & \cellcolor[HTML]{FDCA7D}0.365 & \cellcolor[HTML]{FDD680}0.403 & \cellcolor[HTML]{FEE883}0.459 & \cellcolor[HTML]{FDD17F}0.387 & \cellcolor[HTML]{B5D680}0.615 & \cellcolor[HTML]{ABD380}0.635 & \cellcolor[HTML]{CEDD82}0.564 & \cellcolor[HTML]{F5E884}0.488 & \cellcolor[HTML]{FCC37C}0.341 & \cellcolor[HTML]{F4E884}0.490 &  & \cellcolor[HTML]{F1E784}0.496 \\
MixFormer & \cellcolor[HTML]{78C47D}0.736 & \cellcolor[HTML]{B5D680}0.615 & \cellcolor[HTML]{E2E383}0.526 & \cellcolor[HTML]{E1E383}0.527 & \cellcolor[HTML]{8ACA7E}0.700 & \cellcolor[HTML]{A6D27F}0.644 & \cellcolor[HTML]{8BCA7E}0.698 & \cellcolor[HTML]{F3E884}0.491 &  & \cellcolor[HTML]{C2DA81}0.588 & \cellcolor[HTML]{AFD480}0.626 & \cellcolor[HTML]{9CCF7F}0.664 & \cellcolor[HTML]{AFD480}0.627 & \cellcolor[HTML]{7FC77D}0.721 & \cellcolor[HTML]{63BE7B}0.776 & \cellcolor[HTML]{7FC67D}0.722 & \cellcolor[HTML]{B3D680}0.618 & \cellcolor[HTML]{FDEB84}0.472 & \cellcolor[HTML]{9CCF7F}0.664 &  & \cellcolor[HTML]{ABD380}0.634\\
			\botrule
			\end{tabular}
			\begin{tablenotes}
				\item [1] \textit{
				$e_1$-OTB2015 \cite{OTB2015}. 
				$e_2$-VOT2016 \cite{VOT2016}.
				$e_{3}$-VOT2018 \cite{VOT2018}.
				$e_{4}$-VOT2019 \cite{VOT2019}.
				$e_{5}$-GOT-10k \cite{GOT-10k}.
				$e_{6}$-VOTLT2019 \cite{VOT2019}.
				$e_{7}$-LaSOT \cite{LaSOT}.
				$e_{8}$-VideoCube \cite{GIT}.
				$c_1$-abnormal ratio.
				$c_2$-abnormal scale.
				$c_3$-abnormal illumination.
				$c_4$-blur bbox.
				$c_5$-delta ratio.
				$c_6$-delta scale.
				$c_7$-delta illumination.
				$c_8$-delta blur bbox.
				$c_9$-fast motion.
				$c_{10}$-low correlation coefficient.
				}
				\item[2] \textit{
				The background color of cells indicates the score, with red indicating a low score and green indicating a high score.
				}
				\end{tablenotes}
		\label{tab:success-l}
	\end{center}
	\end{center}
	\end{sidewaystable*}

\clearpage
\onecolumn
\section{Experiments in Short-term Tracking}
\label{supsec:st}

\subsection{Experiments in OTB \cite{OTB2015}}

\begin{figure*}[h!]
  \centering
  \includegraphics[width=0.95\linewidth]{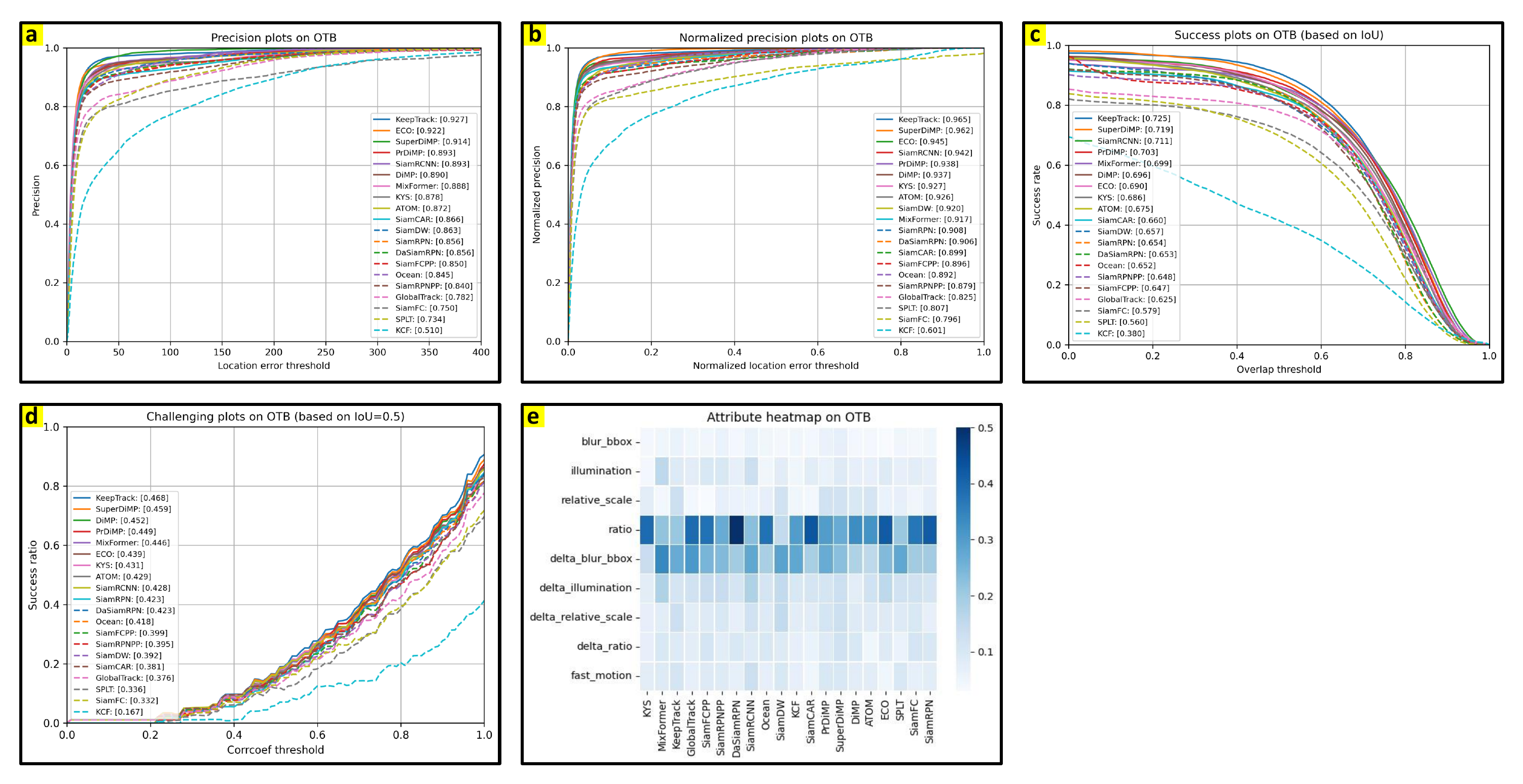}
  \caption{Experiments in OTB \cite{OTB2015} with OPE mechanisms, evaluated by (a) precision plot, (b) normalized precision plot, (c) success plot, (d) challenging plot, and (e)attribute plot. }
  \label{fig:otb-ope}
  \end{figure*}

\begin{figure*}[h!]
  \centering
  \includegraphics[width=0.95\linewidth]{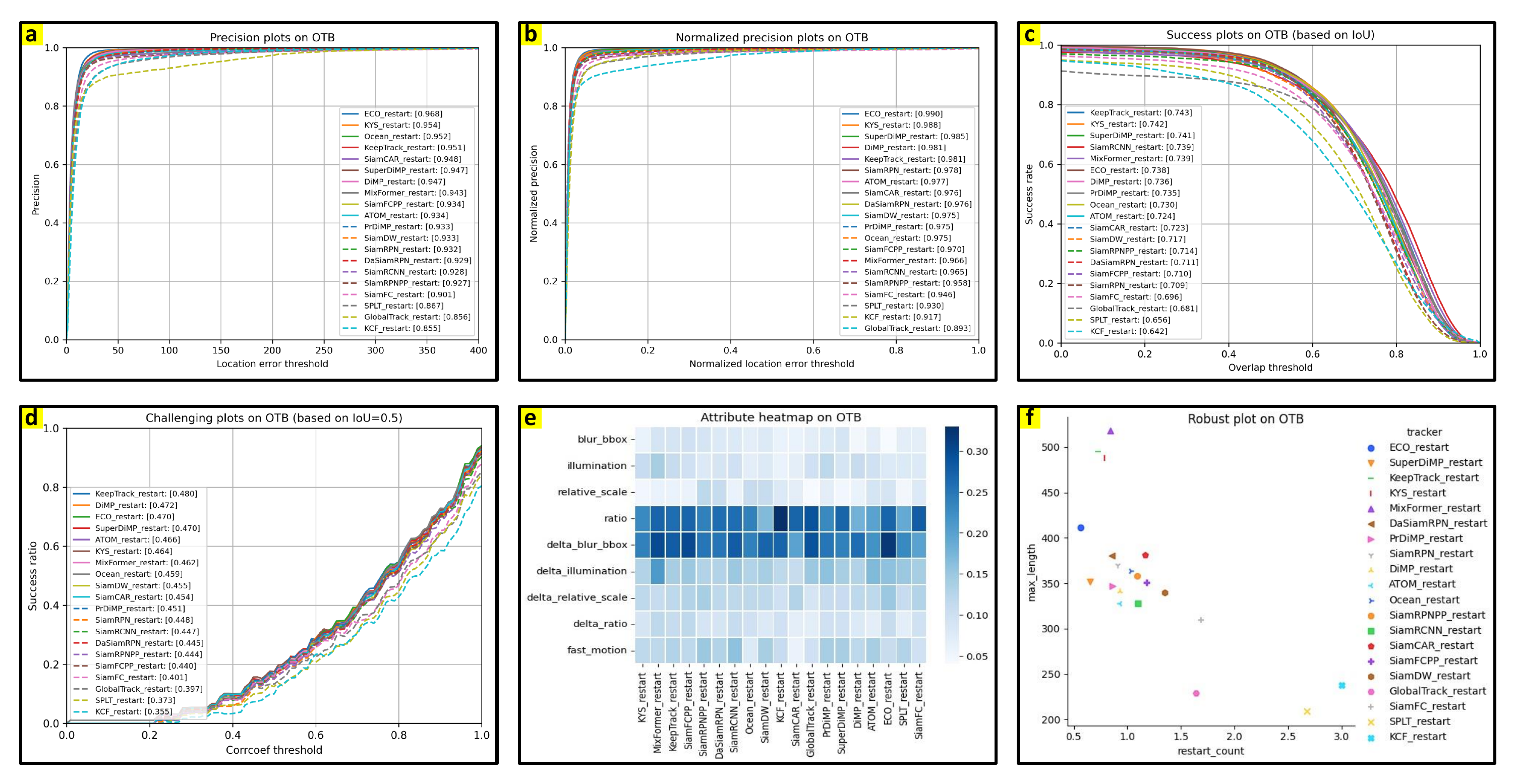}
  \caption{Experiments in OTB \cite{OTB2015} with R-OPE mechanisms, evaluated by (a) precision plot, (b) normalized precision plot, (c) success plot, (d) challenging plot, (e)attribute plot, and (f) robust plot. }
  \label{fig:otb-rope}
  \end{figure*}

\clearpage
\onecolumn

\subsection{Experiments in VOT2016 \cite{VOT2016}}

\begin{figure}[h!]
  \centering
  \includegraphics[width=0.95\linewidth]{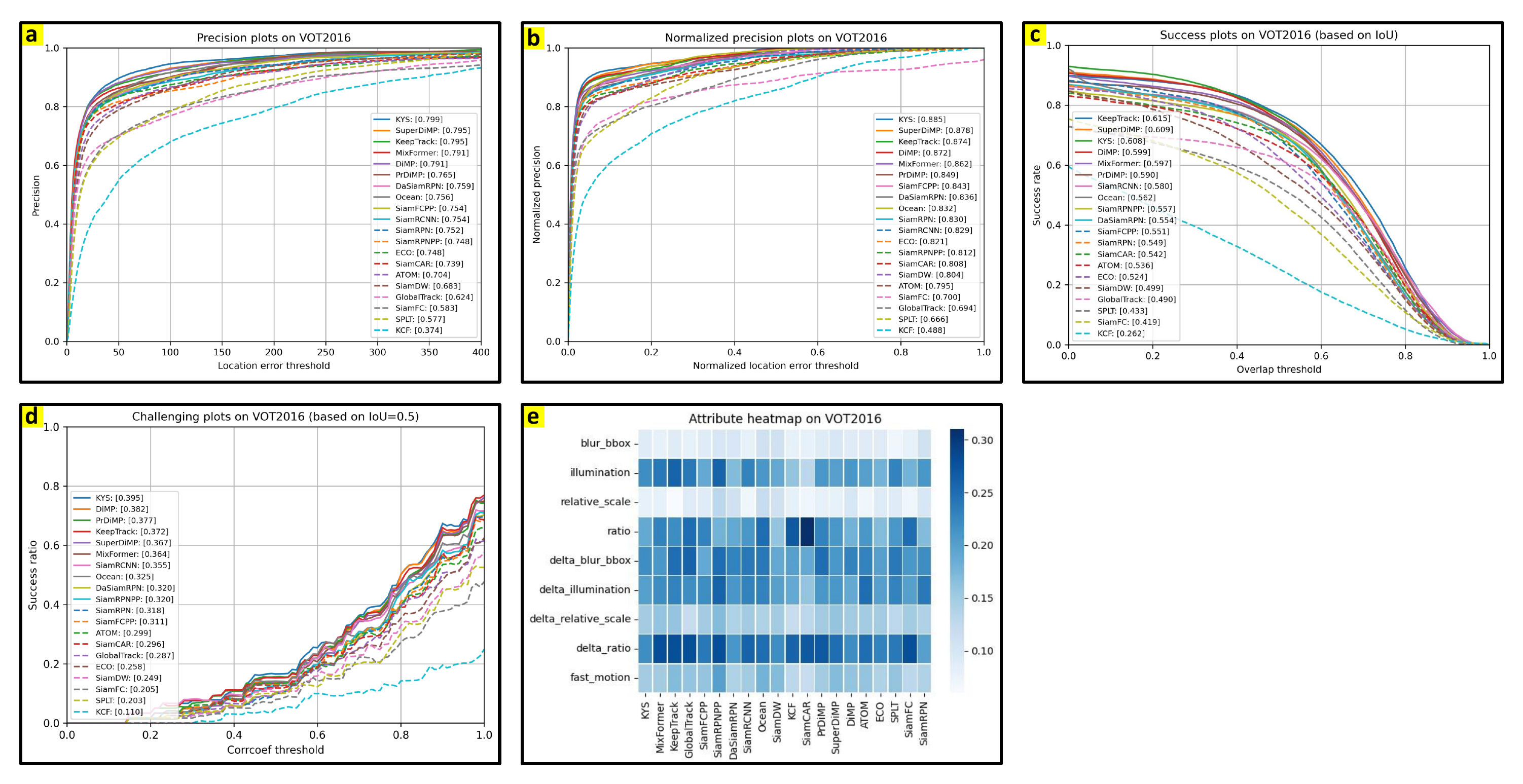}
  \caption{Experiments in VOT2016 \cite{VOT2016} with OPE mechanisms, evaluated by (a) precision plot, (b) normalized precision plot, (c) success plot, (d) challenging plot, and (e)attribute plot. }
  \label{fig:vot2016-ope}
  \end{figure}

\begin{figure}[h!]
  \centering
  \includegraphics[width=0.95\linewidth]{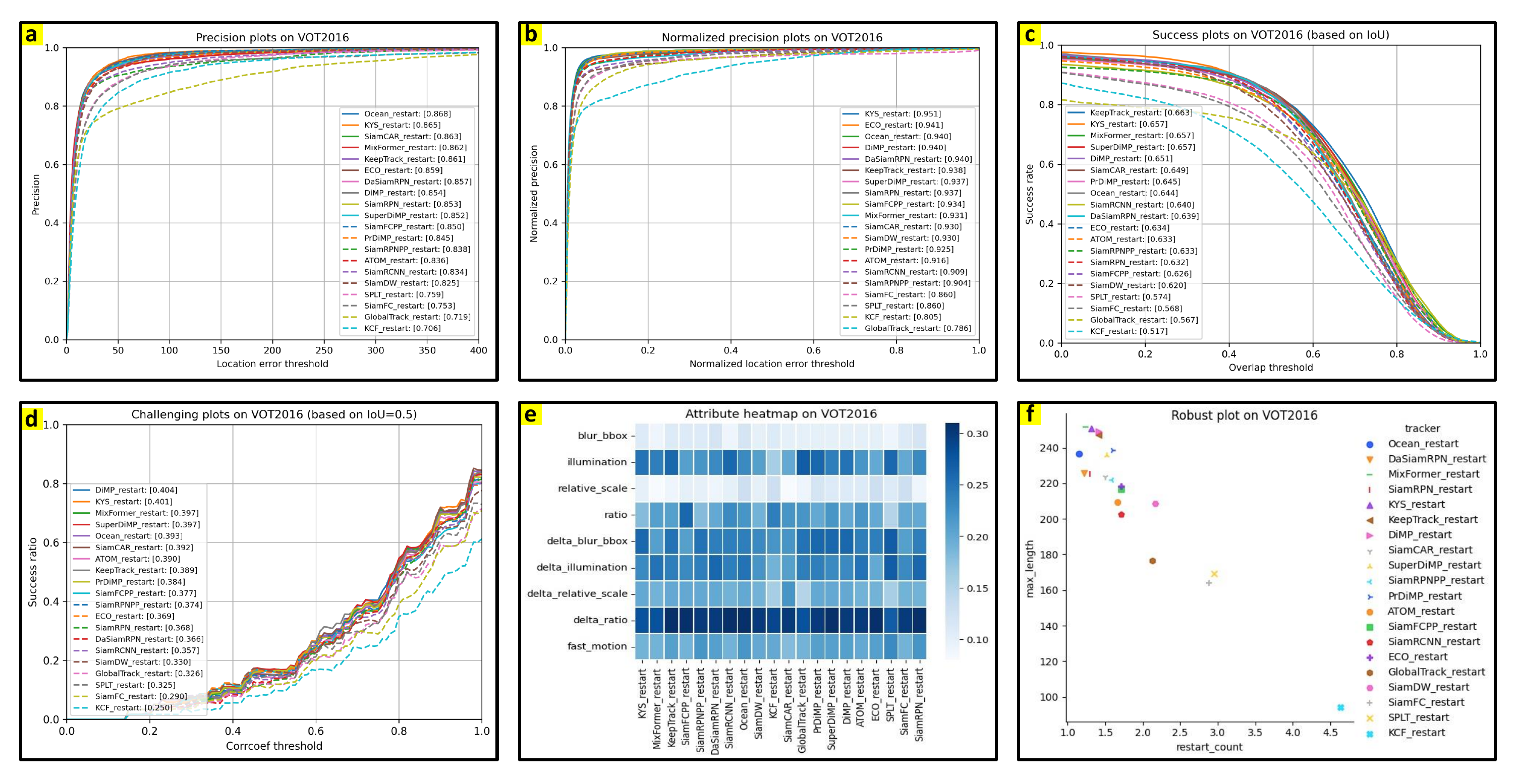}
  \caption{Experiments in VOT2016 \cite{VOT2016} with R-OPE mechanisms, evaluated by (a) precision plot, (b) normalized precision plot, (c) success plot, (d) challenging plot, (e)attribute plot, and (f) robust plot. }
  \label{fig:vot2016-rope}
  \end{figure}

\clearpage
\onecolumn
\subsection{Experiments in VOT2018 \cite{VOT2018}}

\begin{figure}[h!]
  \centering
  \includegraphics[width=0.95\linewidth]{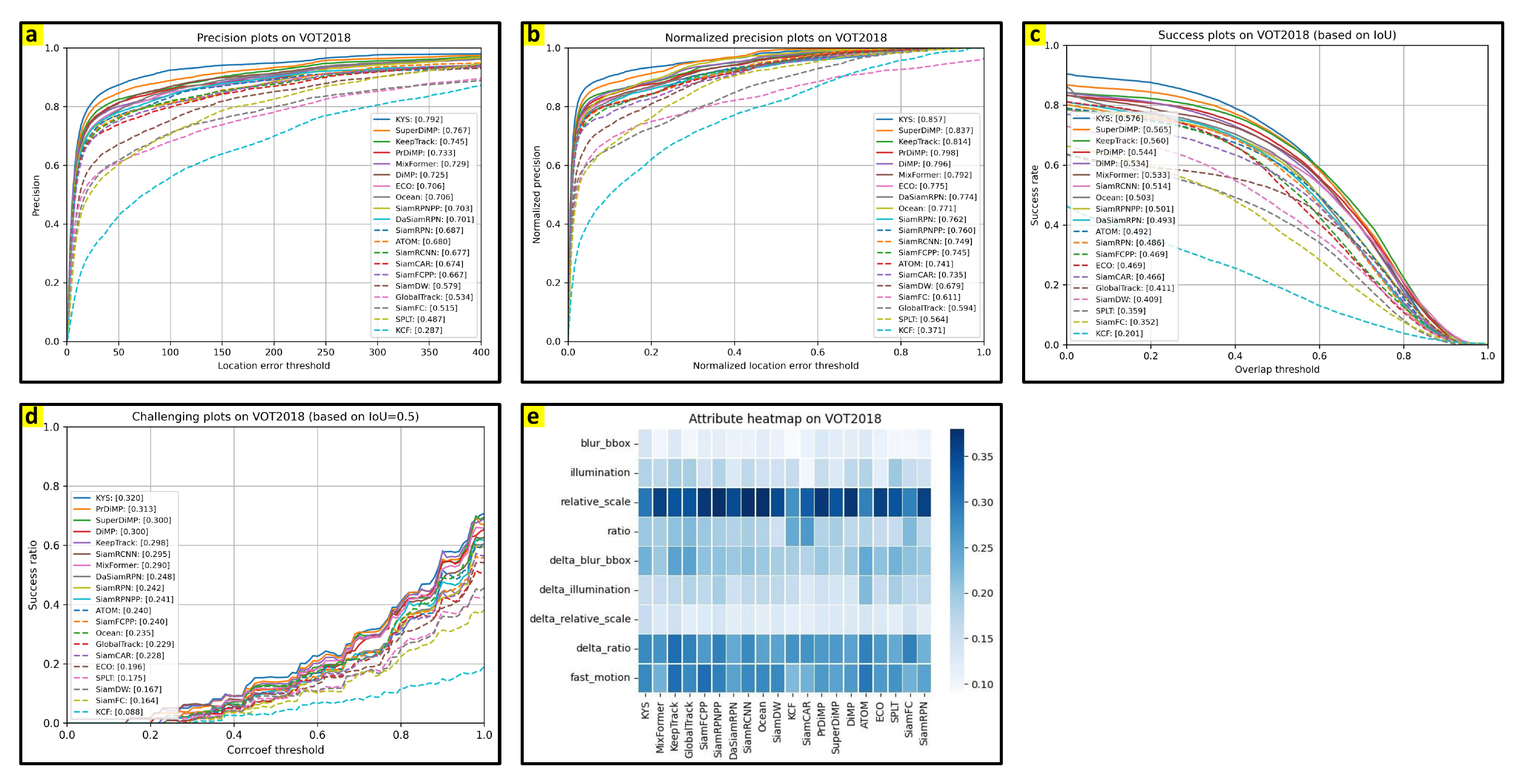}
  \caption{Experiments in VOT2018 \cite{VOT2018} with OPE mechanisms, evaluated by (a) precision plot, (b) normalized precision plot, (c) success plot, (d) challenging plot, and (e)attribute plot. }
  \label{fig:vot2018-ope}
  \end{figure}

\begin{figure}[h!]
  \centering
  \includegraphics[width=0.95\linewidth]{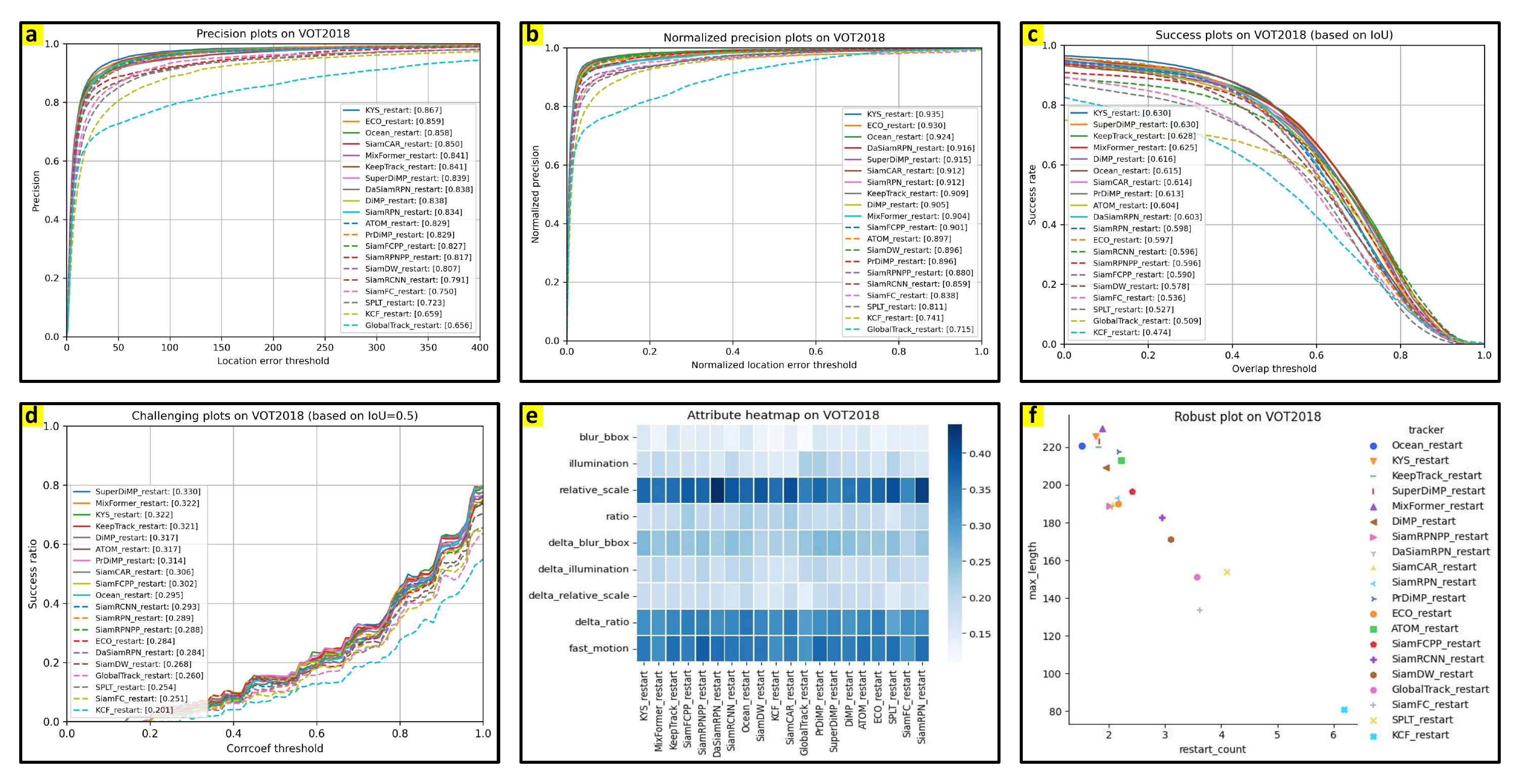}
  \caption{Experiments in VOT2018 \cite{VOT2018} with R-OPE mechanisms, evaluated by (a) precision plot, (b) normalized precision plot, (c) success plot, (d) challenging plot, (e)attribute plot, and (f) robust plot. }
  \label{fig:vot2018-rope}
  \end{figure}

\clearpage
\onecolumn
\subsection{Experiments in VOT2019 \cite{VOT2019}}

\begin{figure}[h!]
  \centering
  \includegraphics[width=0.95\linewidth]{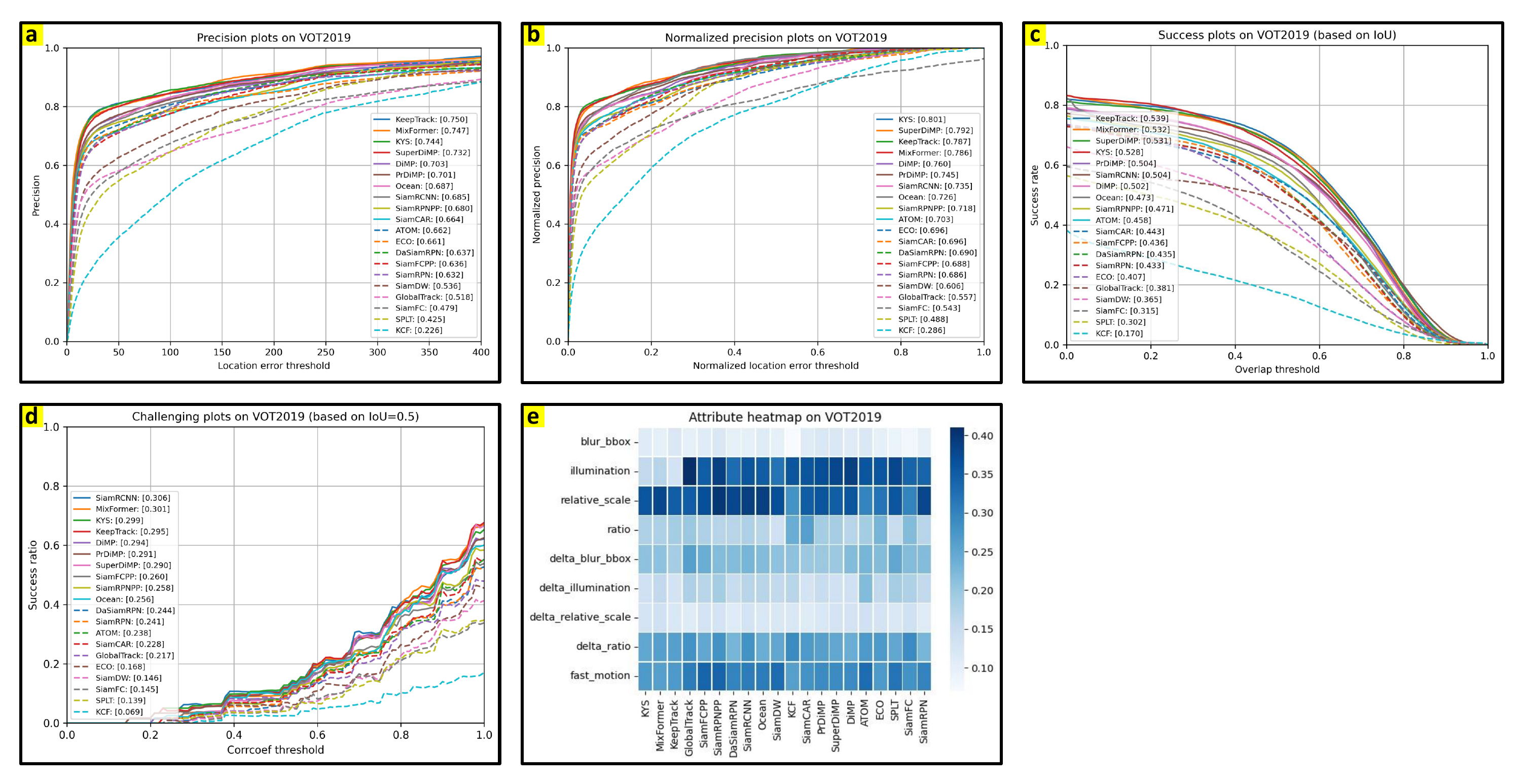}
  \caption{Experiments in VOT2019 \cite{VOT2019} with OPE mechanisms, evaluated by (a) precision plot, (b) normalized precision plot, (c) success plot, (d) challenging plot, and (e)attribute plot. }
  \label{fig:vot2019-ope}
  \end{figure}

\begin{figure}[h!]
  \centering
  \includegraphics[width=0.95\linewidth]{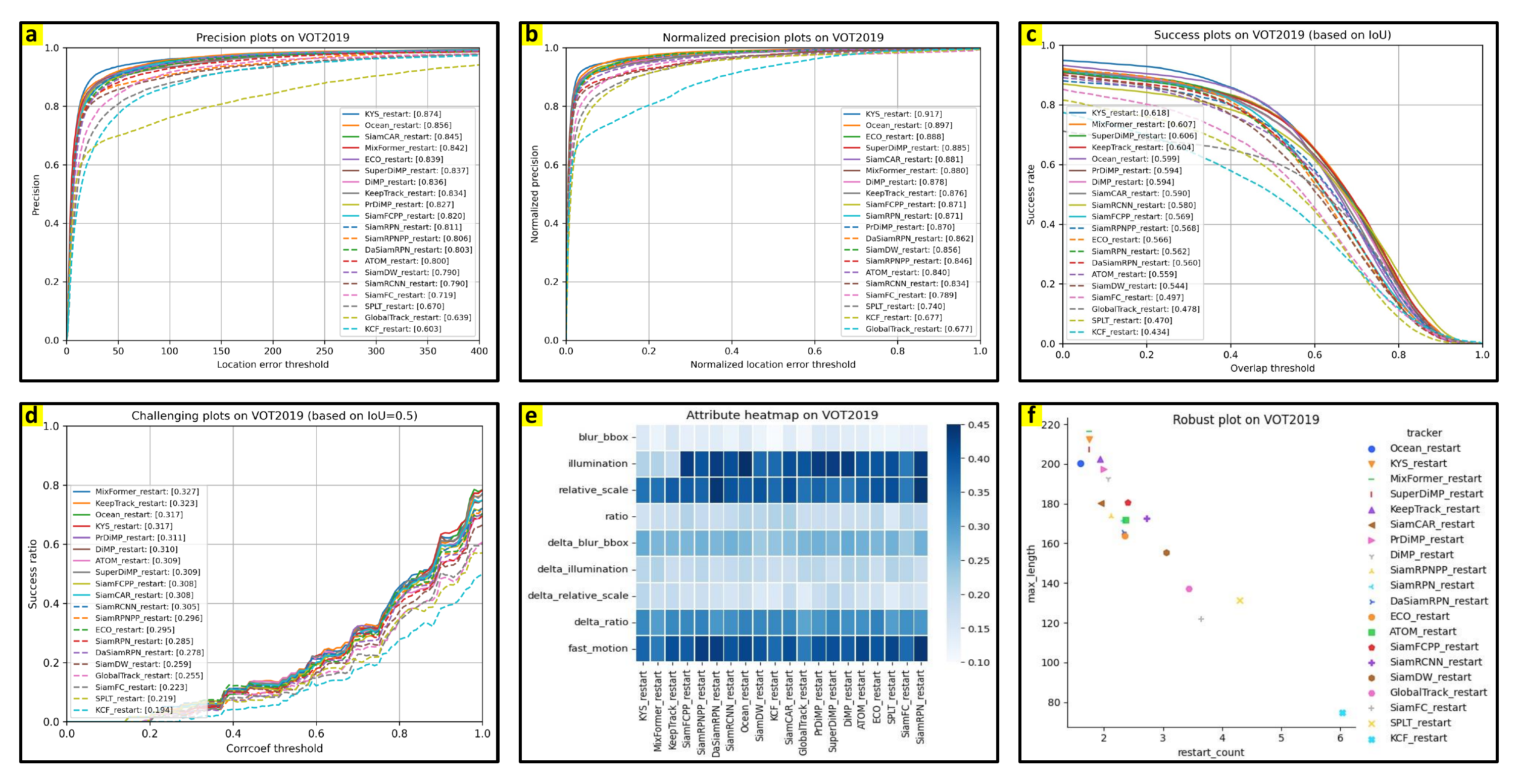}
  \caption{Experiments in VOT2019 \cite{VOT2019} with R-OPE mechanisms, evaluated by (a) precision plot, (b) normalized precision plot, (c) success plot, (d) challenging plot, (e)attribute plot, and (f) robust plot. }
  \label{fig:vot2019-rope}
  \end{figure}

\clearpage
\onecolumn
\subsection{Experiments in GOT-10k \cite{GOT-10k}}

\begin{figure}[h!]
  \centering
  \includegraphics[width=0.95\linewidth]{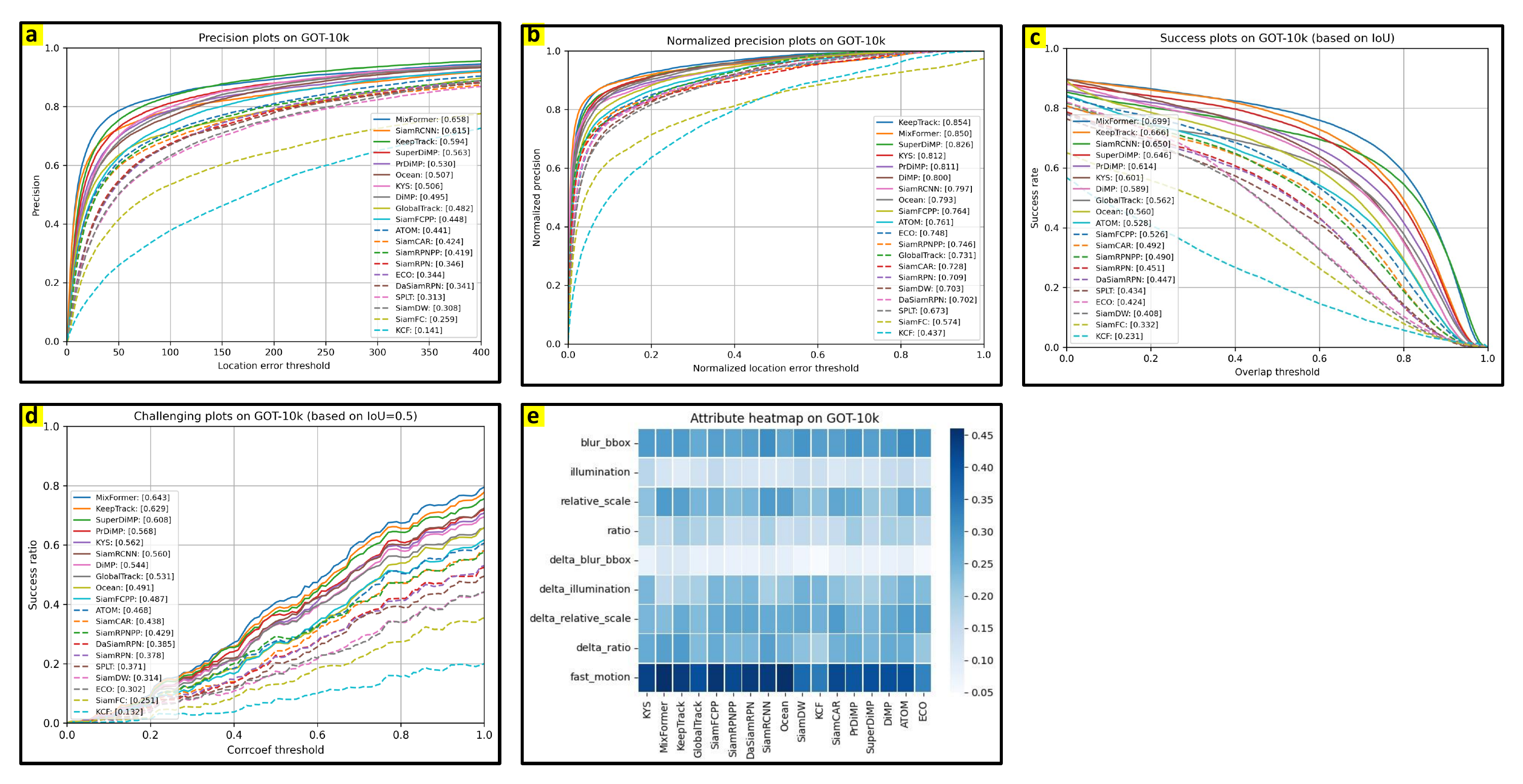}
  \caption{Experiments in GOT-10k \cite{GOT-10k} with OPE mechanisms, evaluated by (a) precision plot, (b) normalized precision plot, (c) success plot, (d) challenging plot, and (e)attribute plot. }
  \label{fig:got10k-ope}
  \end{figure}

\begin{figure}[h!]
  \centering
  \includegraphics[width=0.95\linewidth]{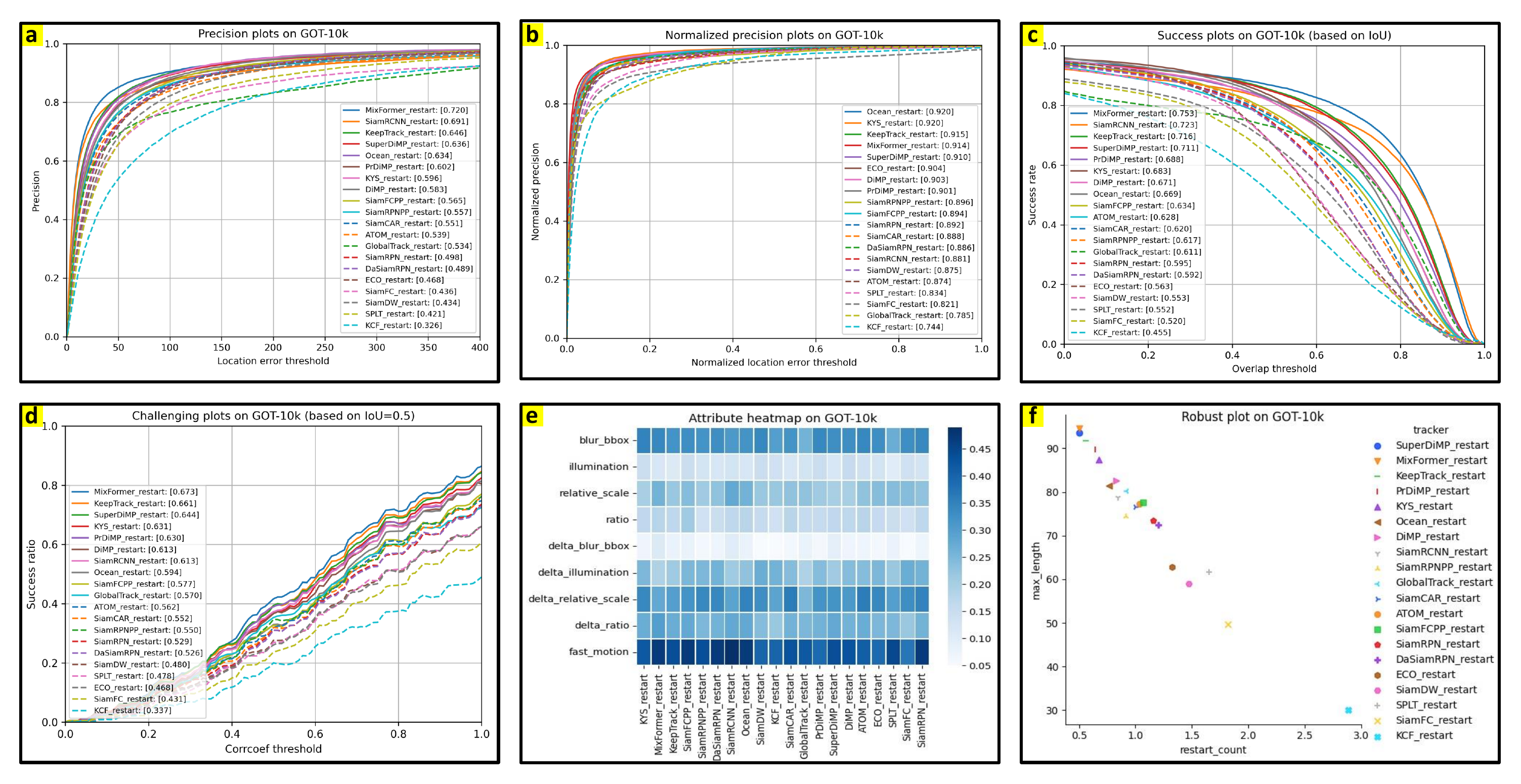}
  \caption{Experiments in GOT-10k \cite{GOT-10k} with R-OPE mechanisms, evaluated by (a) precision plot, (b) normalized precision plot, (c) success plot, (d) challenging plot, (e)attribute plot, and (f) robust plot. }
  \label{fig:got10k-rope}
  \end{figure}

\clearpage
\onecolumn

\section{Experiments in Long-term Tracking}
\label{supsec:lt}
\subsection{Experiments in VOTLT2019 \cite{VOT2019}}

\begin{figure}[h!]
\centering
\includegraphics[width=0.95\linewidth]{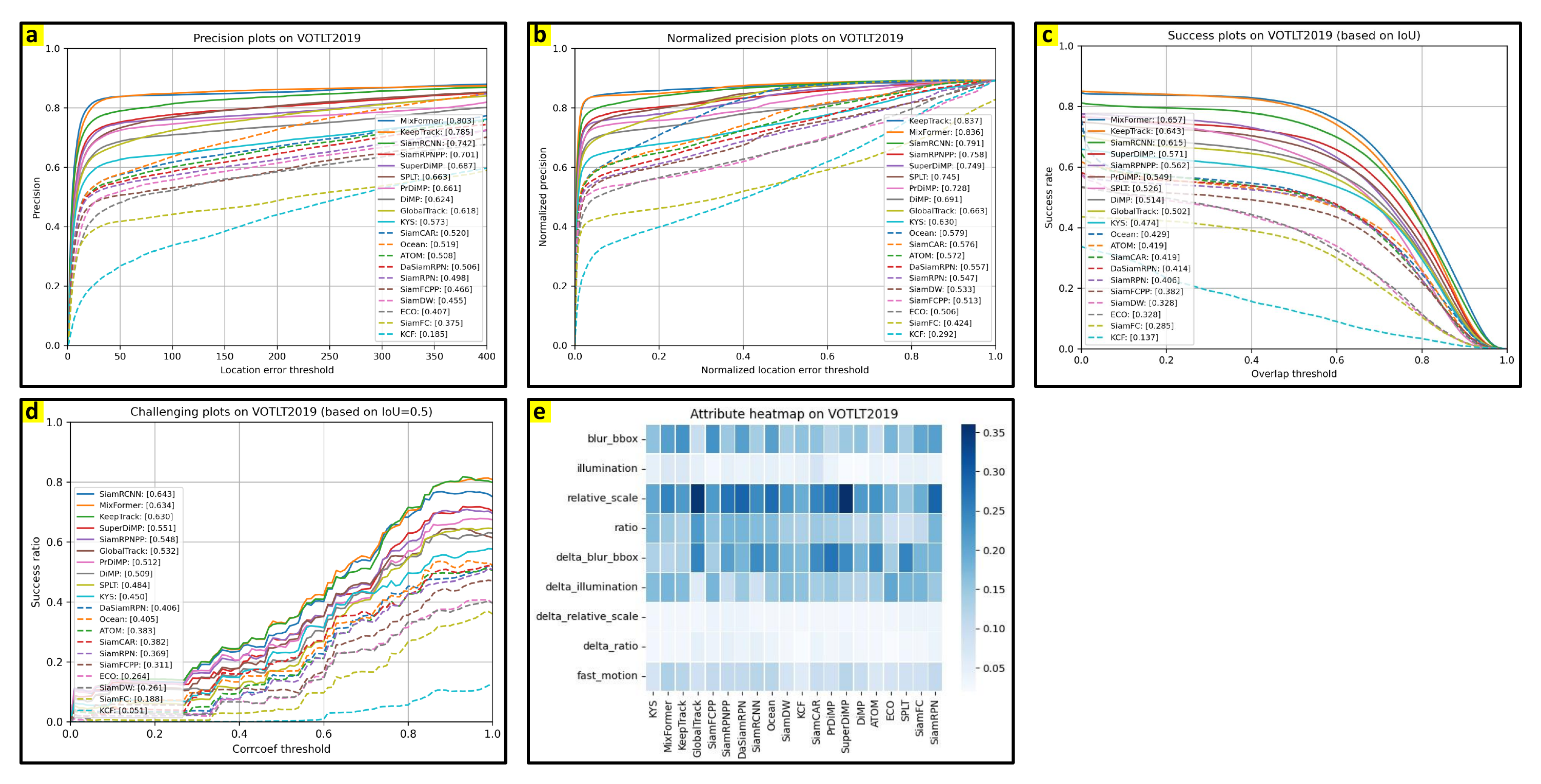}
\caption{Experiments in VOTLT2019 \cite{VOT2019} with OPE mechanisms, evaluated by (a) precision plot, (b) normalized precision plot, (c) success plot, (d) challenging plot, and (e)attribute plot. }
\label{fig:votlt2019-ope}
\end{figure}

\begin{figure}[h!]
\centering
\includegraphics[width=0.95\linewidth]{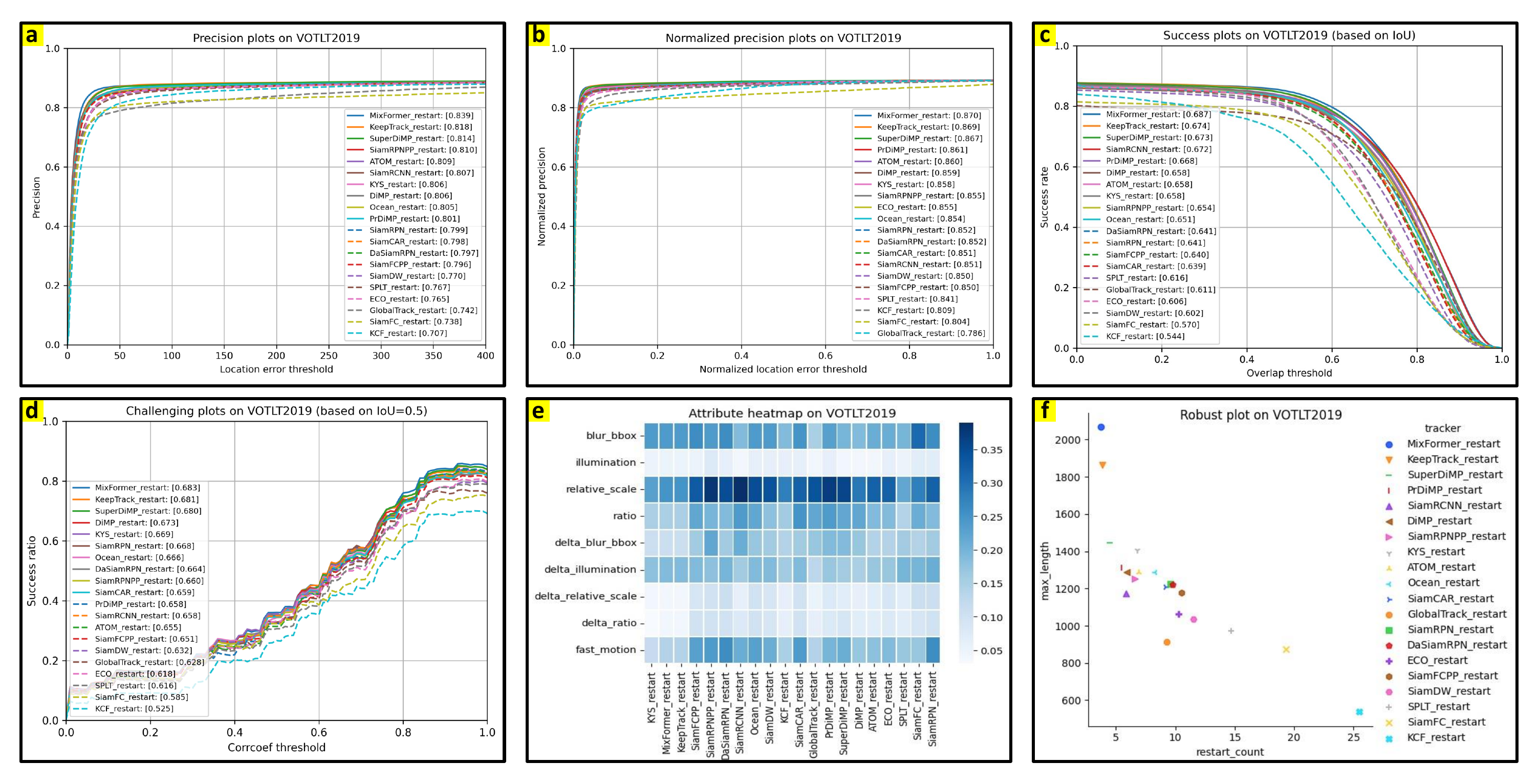}
\caption{Experiments in VOTLT2019 \cite{VOT2019} with R-OPE mechanisms, evaluated by (a) precision plot, (b) normalized precision plot, (c) success plot, (d) challenging plot, (e)attribute plot, and (f) robust plot. }
\label{fig:votlt2019-rope}
\end{figure}

\clearpage
\onecolumn
\subsection{Experiments in LaSOT \cite{LaSOT}}

\begin{figure}[h!]
\centering
\includegraphics[width=0.95\linewidth]{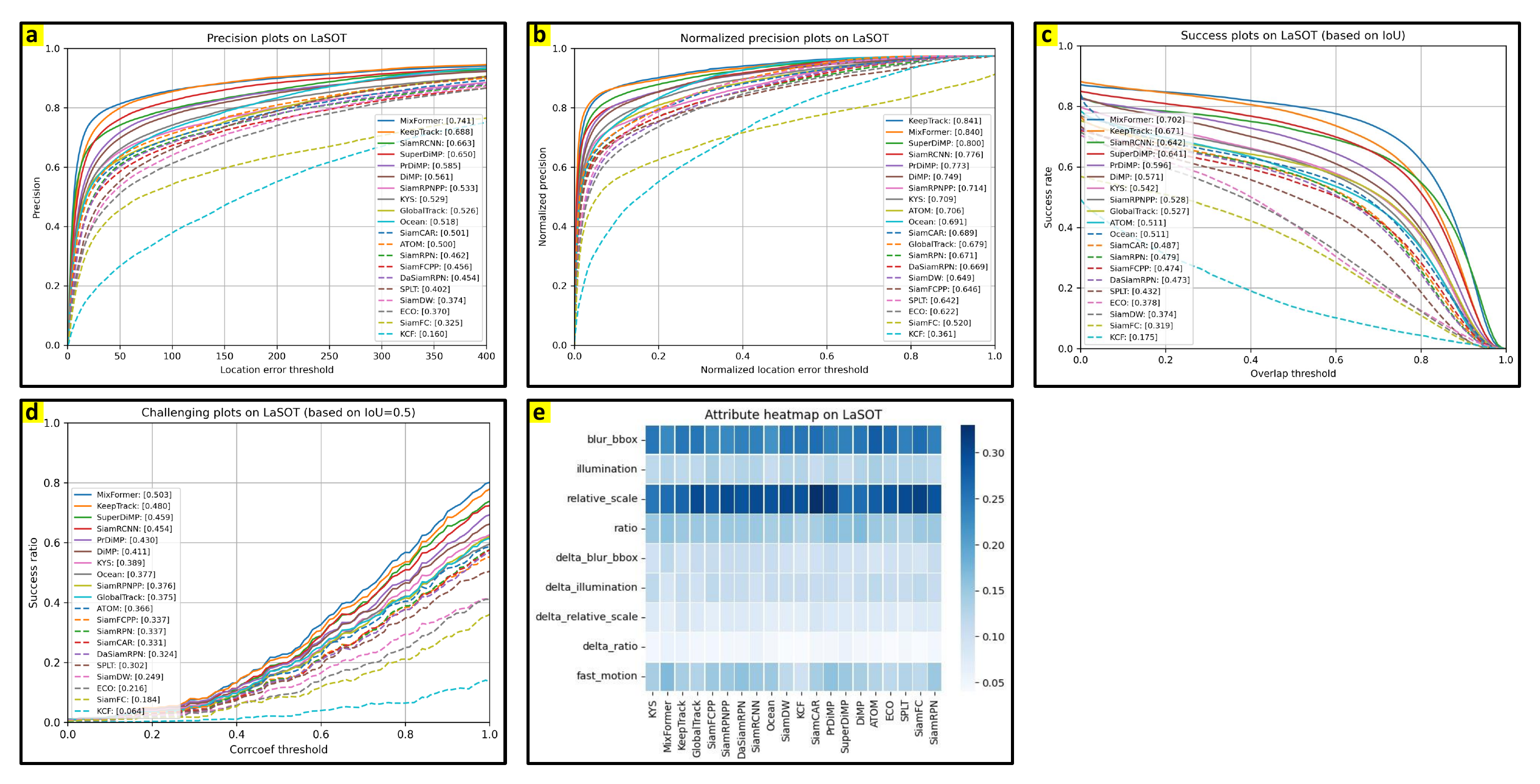}
\caption{Experiments in LaSOT \cite{LaSOT} with OPE mechanisms, evaluated by (a) precision plot, (b) normalized precision plot, (c) success plot, (d) challenging plot, and (e)attribute plot. }
\label{fig:lasot-ope}
\end{figure}

\begin{figure}[h!]
\centering
\includegraphics[width=0.95\linewidth]{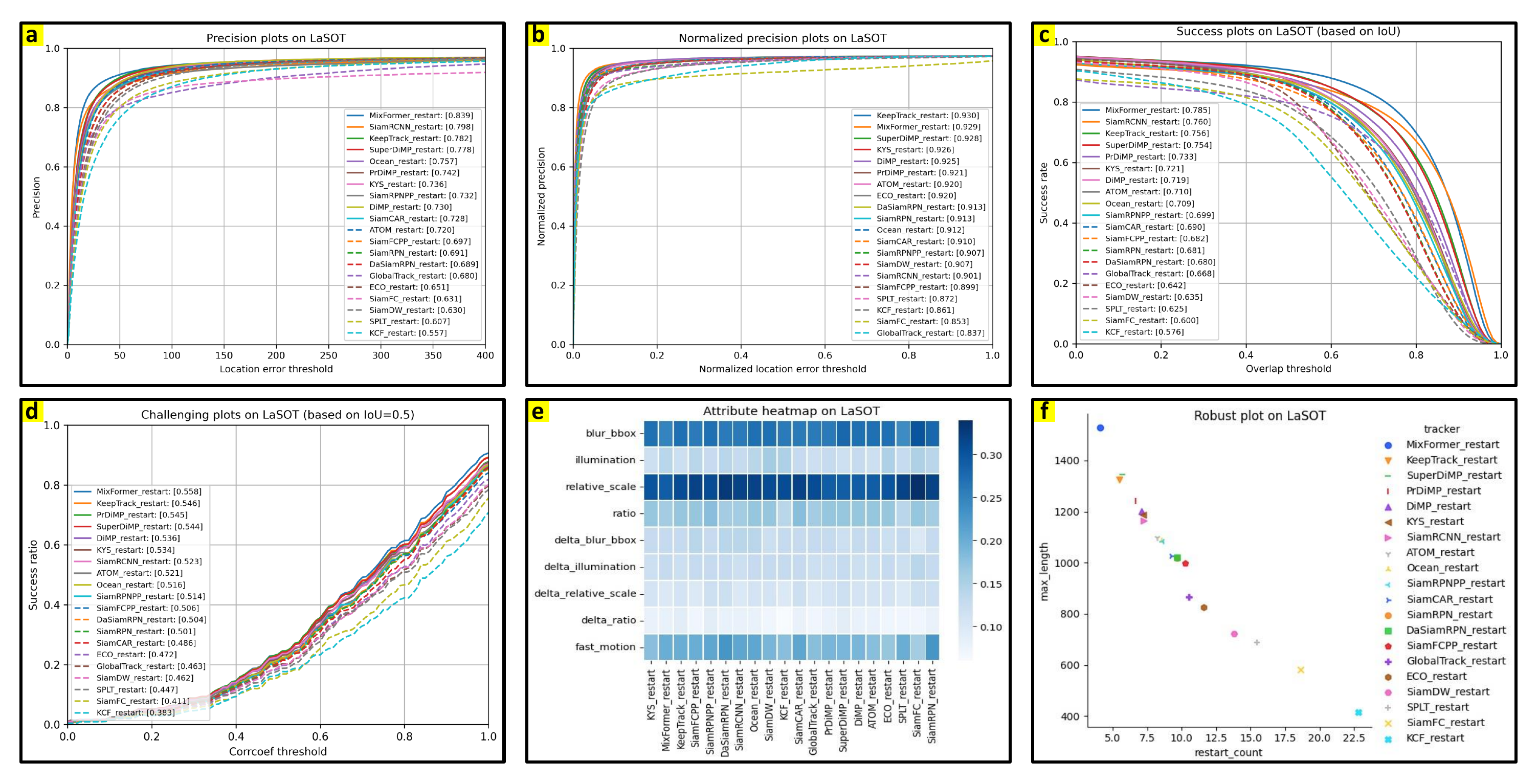}
\caption{Experiments in LaSOT \cite{LaSOT} with R-OPE mechanisms, evaluated by (a) precision plot, (b) normalized precision plot, (c) success plot, (d) challenging plot, (e)attribute plot, and (f) robust plot. }
\label{fig:lasot-rope}
\end{figure}

\clearpage
\onecolumn
\section{Experiments in Global Instance Tracking}
\label{supsec:git}
\subsection{Experiments in VideoCube \cite{GIT}}

\begin{figure}[h!]
\centering
\includegraphics[width=0.95\linewidth]{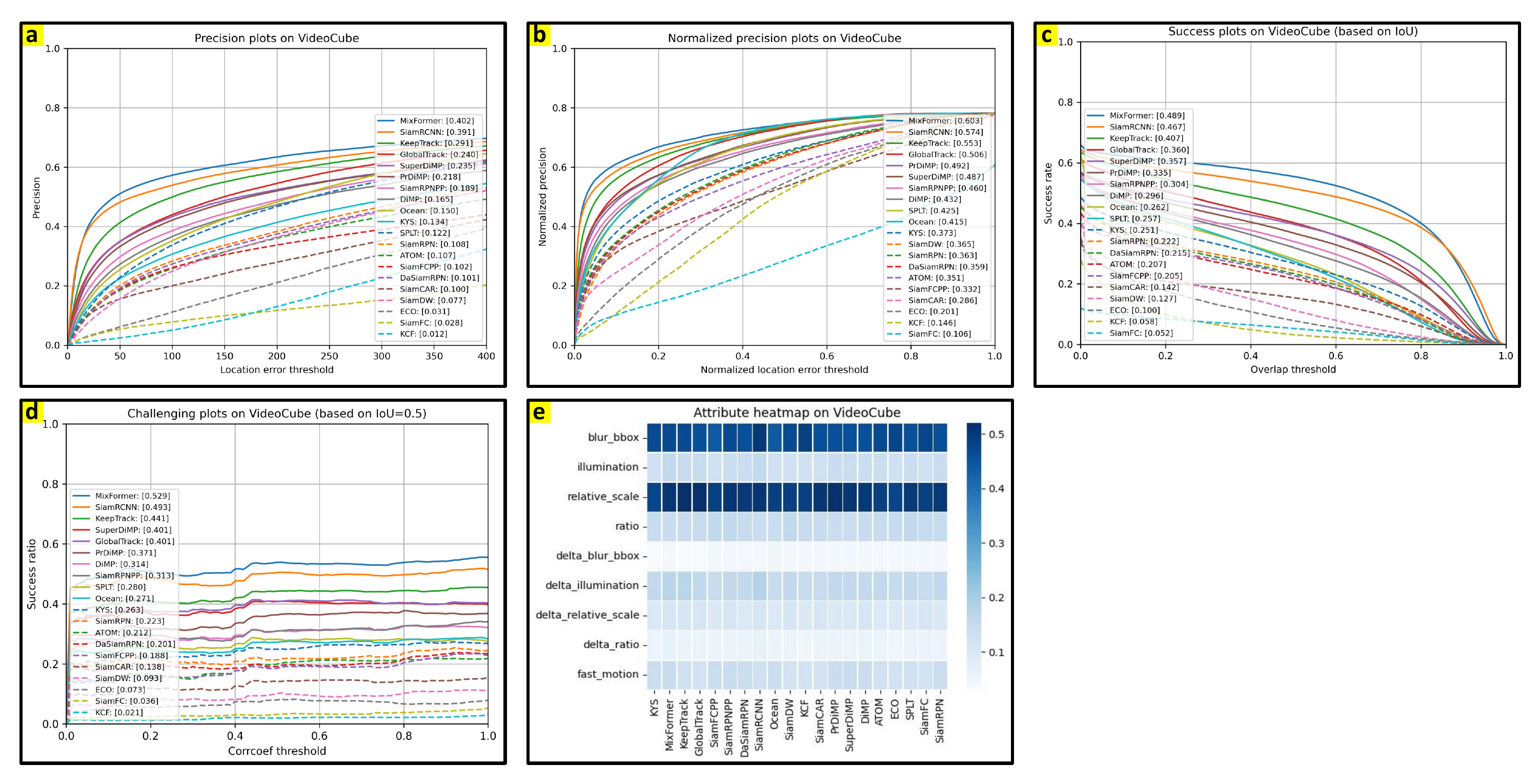}
\caption{Experiments in VideoCube \cite{GIT} with OPE mechanisms, evaluated by (a) precision plot, (b) normalized precision plot, (c) success plot, (d) challenging plot, and (e)attribute plot. }
\label{fig:videocube-ope}
\end{figure}

\begin{figure}[h!]
\centering
\includegraphics[width=0.95\linewidth]{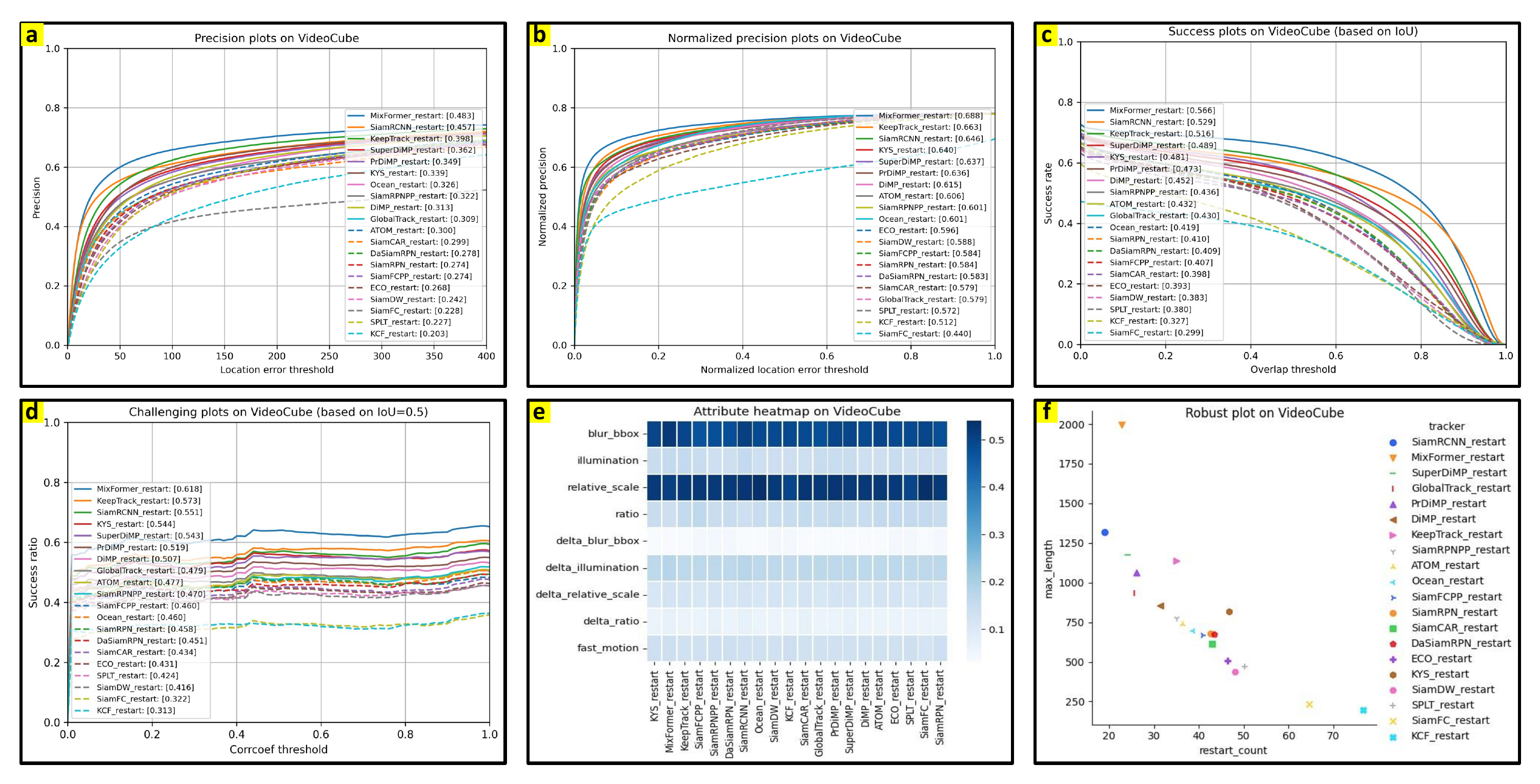}
\caption{Experiments in VideoCube \cite{GIT} with R-OPE mechanisms, evaluated by (a) precision plot, (b) normalized precision plot, (c) success plot, (d) challenging plot, (e)attribute plot, and (f) robust plot. }
\label{fig:videocube-rope}
\end{figure}

\clearpage
\onecolumn
\section{The Composition of Challenging Space}
\label{supsec:composition-cs}

\subsection{Abnormal Ratio}

\begin{figure}[h!]
\centering
\includegraphics[width=0.95\linewidth]{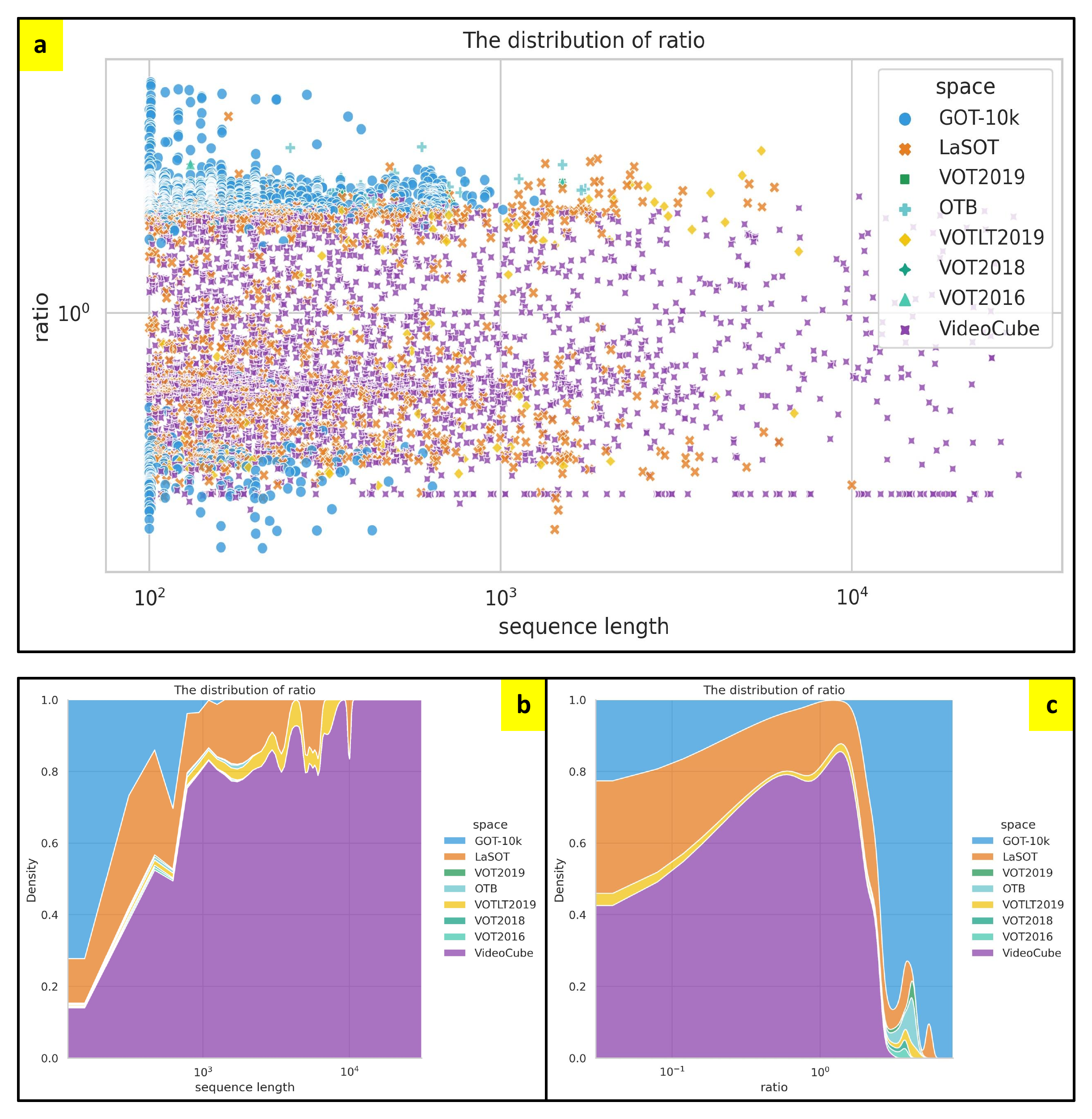}
\caption{The composition of abnormal ratio space. (a) The distribution of attribute values and sequence lengths, each point representing a sub-sequence. (b) The distribution of sequence lengths. (c) The distribution of attribute values.}
\label{fig:ratio}
\end{figure}

\clearpage
\onecolumn
\subsection{Abnormal Scale}

\begin{figure}[h!]
\centering
\includegraphics[width=0.95\linewidth]{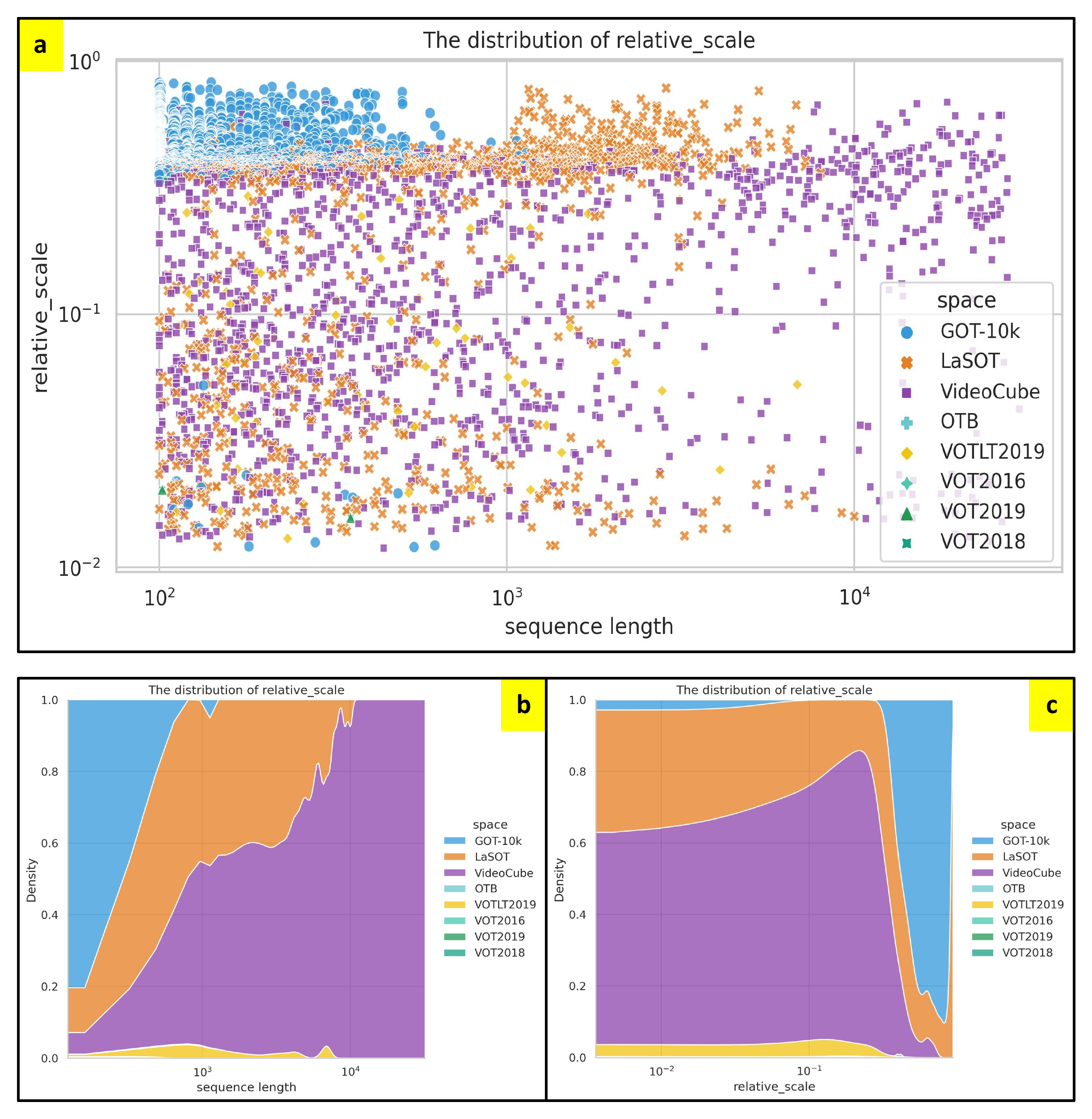}
\caption{The composition of abnormal scale space. (a) The distribution of attribute values and sequence lengths, each point representing a sub-sequence. (b) The distribution of sequence lengths. (c) The distribution of attribute values.}
\label{fig:relative_scale}
\end{figure}

\clearpage
\onecolumn
\subsection{Abnormal Illumination}

\begin{figure}[h!]
\centering
\includegraphics[width=0.95\linewidth]{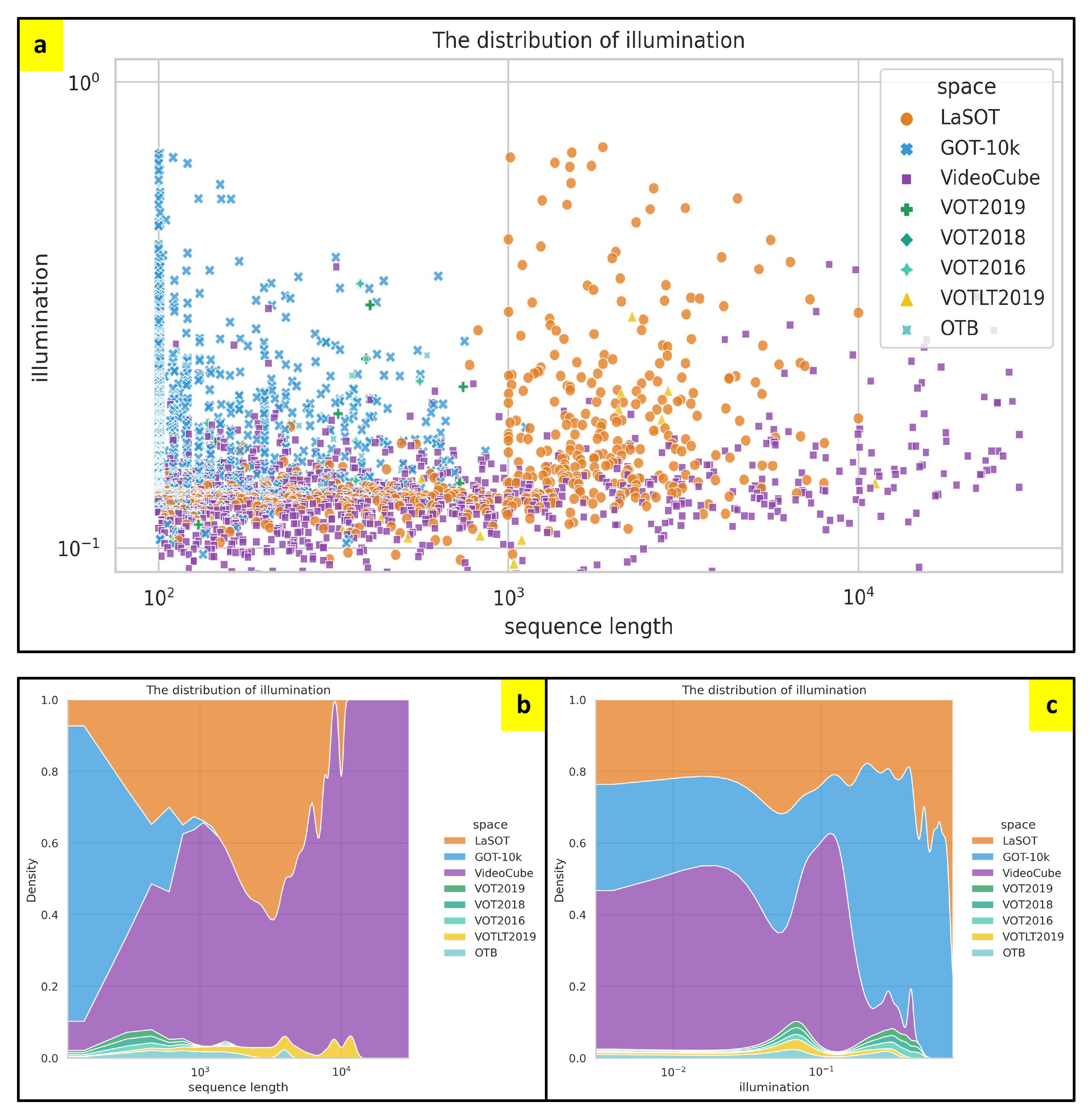}
\caption{The composition of abnormal illumination space. (a) The distribution of attribute values and sequence lengths, each point representing a sub-sequence. (b) The distribution of sequence lengths. (c) The distribution of attribute values.}
\label{fig:illumination}
\end{figure}

\clearpage
\onecolumn
\subsection{Blur Bounding-box}

\begin{figure}[h!]
\centering
\includegraphics[width=0.95\linewidth]{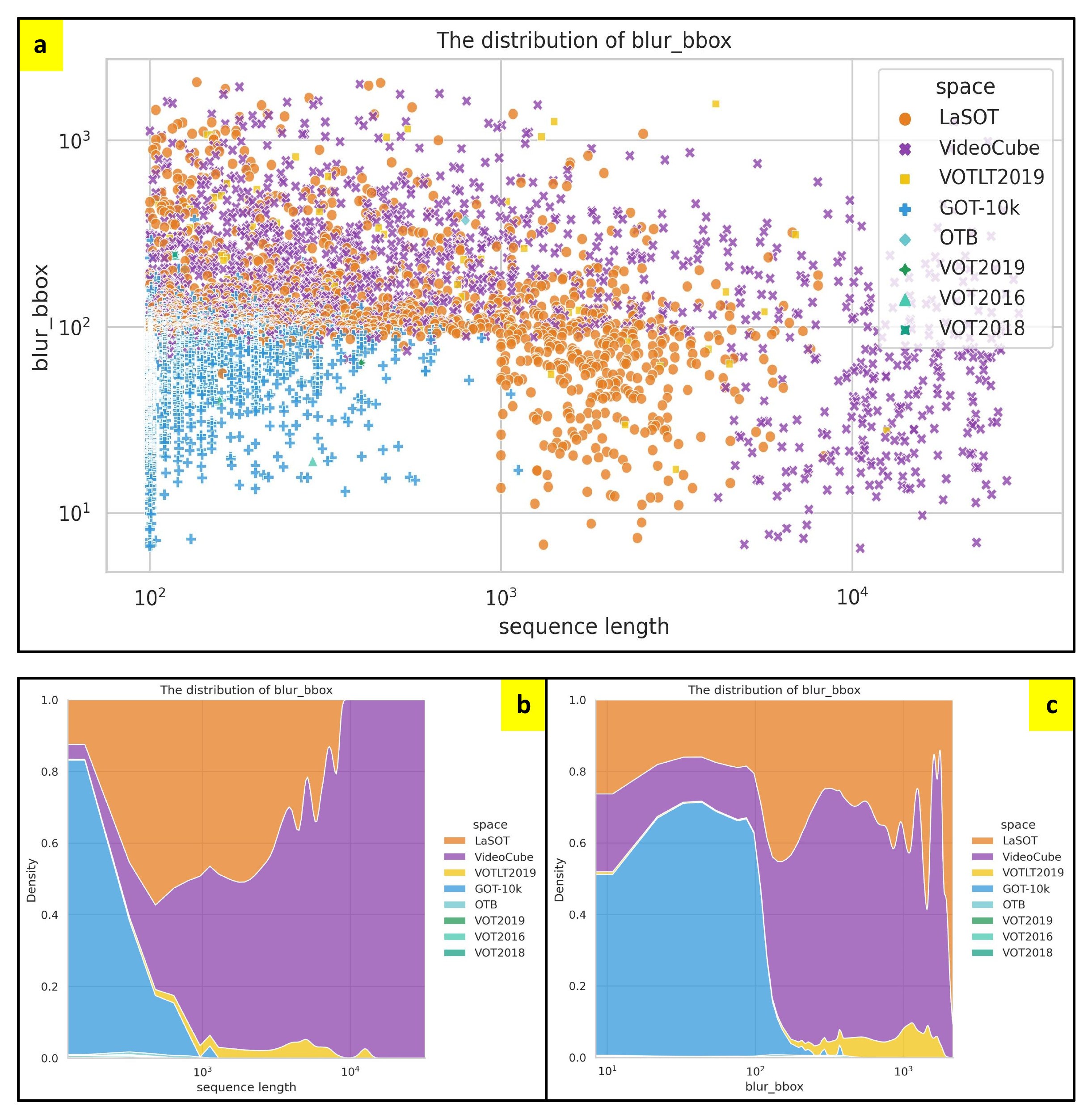}
\caption{The composition of blur bounding-box space. (a) The distribution of attribute values and sequence lengths, each point representing a sub-sequence. (b) The distribution of sequence lengths. (c) The distribution of attribute values.}
\label{fig:blur_bbox}
\end{figure}

\clearpage
\onecolumn
\subsection{Delta Ratio}

\begin{figure}[h!]
\centering
\includegraphics[width=0.95\linewidth]{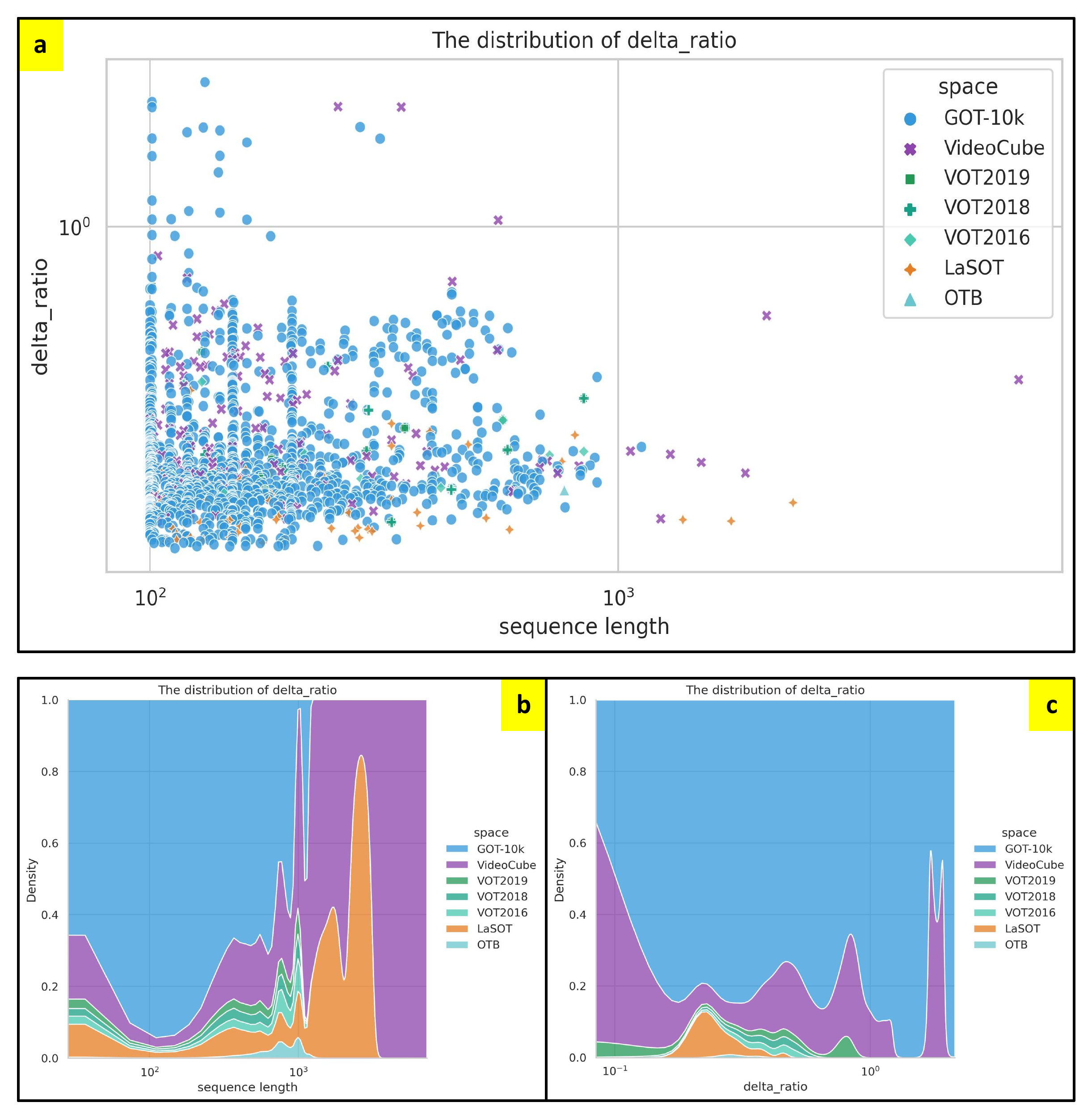}
\caption{The composition of delta ratio space. (a) The distribution of attribute values and sequence lengths, each point representing a sub-sequence. (b) The distribution of sequence lengths. (c) The distribution of attribute values.}
\label{fig:delta_ratio}
\end{figure}

\clearpage
\onecolumn
\subsection{Delta Scale}

\begin{figure}[h!]
\centering
\includegraphics[width=0.95\linewidth]{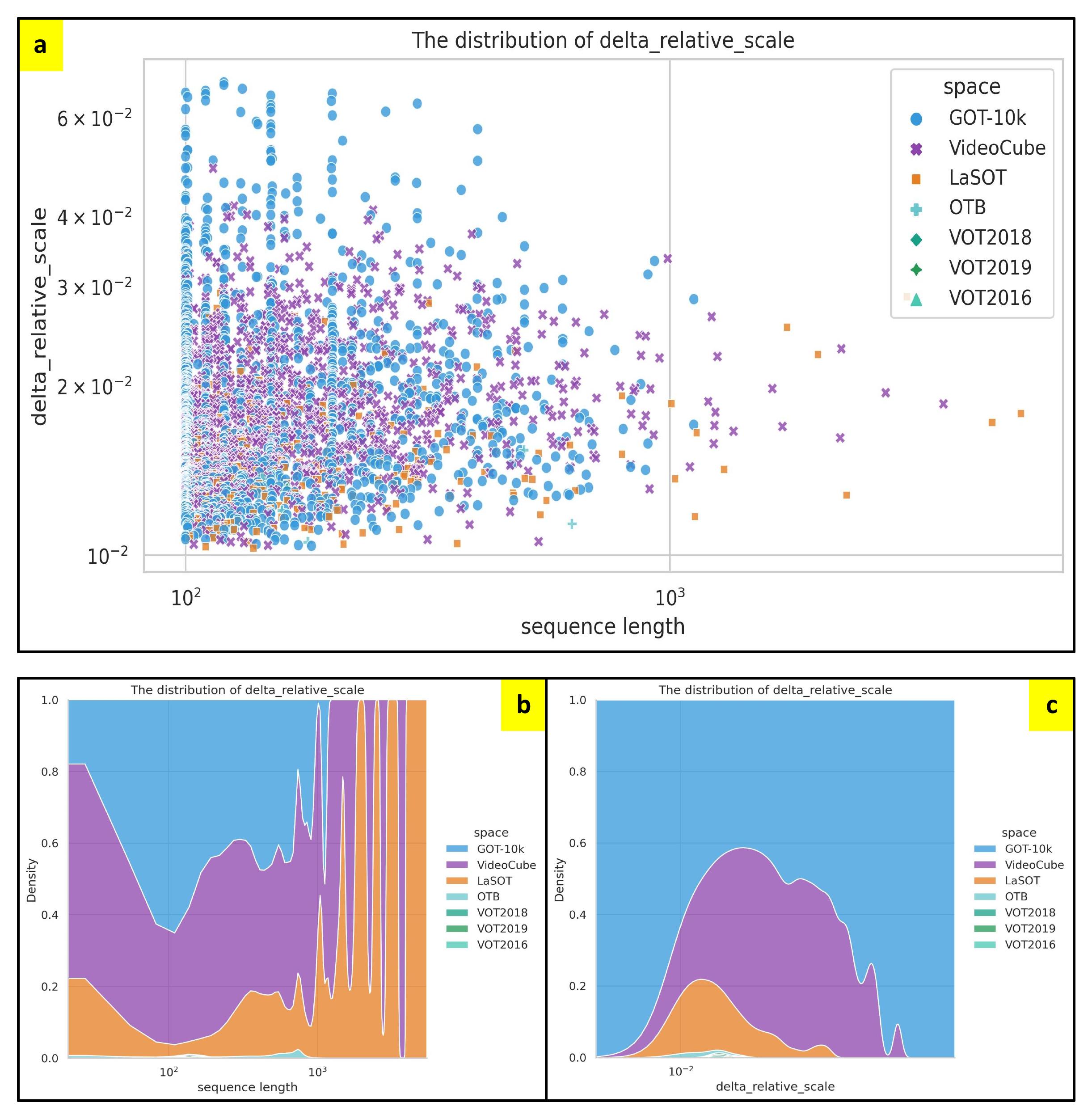}
\caption{The composition of delta scale space. (a) The distribution of attribute values and sequence lengths, each point representing a sub-sequence. (b) The distribution of sequence lengths. (c) The distribution of attribute values.}
\label{fig:delta_relative_scale}
\end{figure}

\clearpage
\onecolumn
\subsection{Delta Illumination}

\begin{figure}[h!]
\centering
\includegraphics[width=0.95\linewidth]{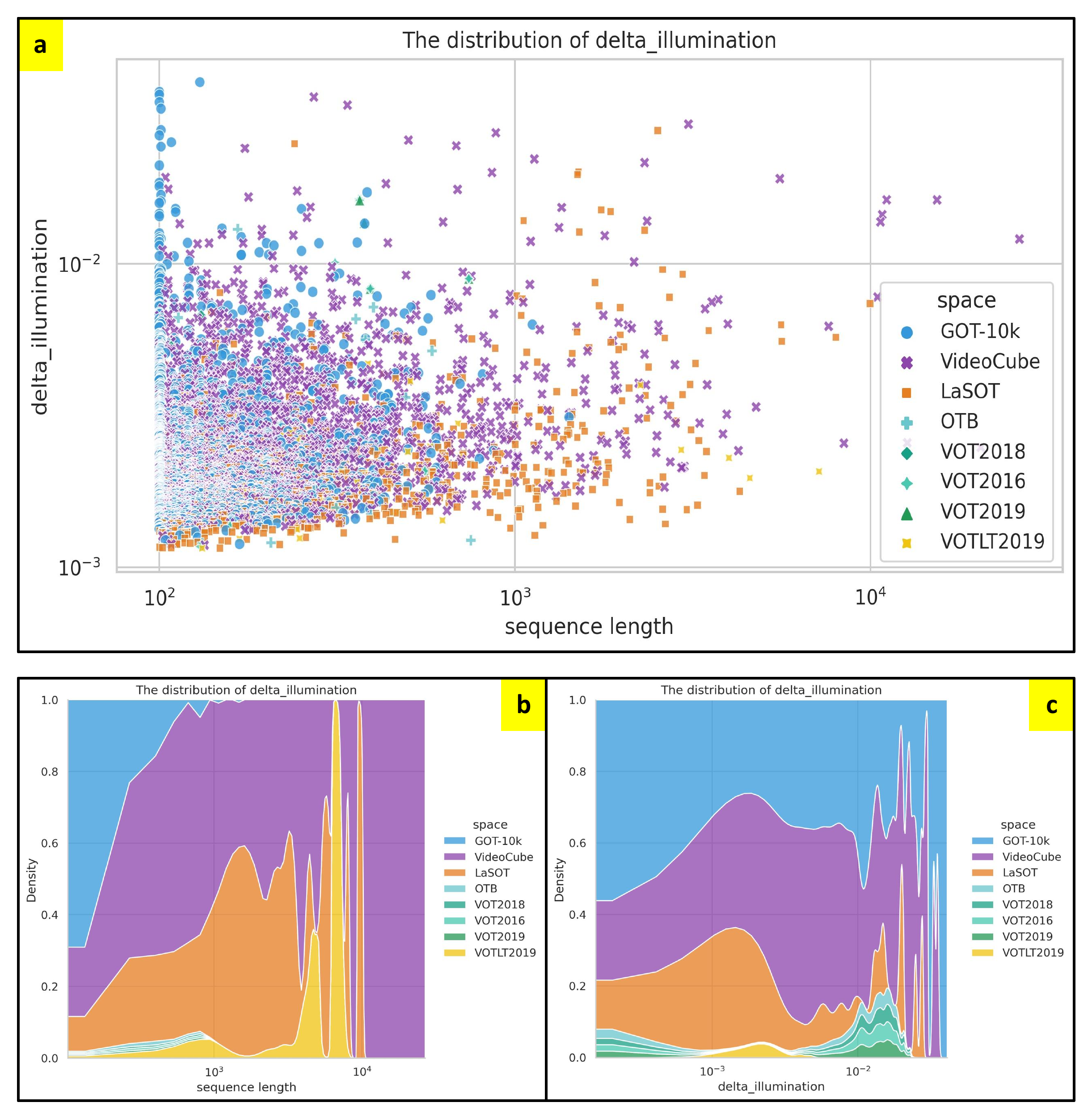}
\caption{The composition of delta illumination. (a) The distribution of attribute values and sequence lengths, each point representing a sub-sequence. (b) The distribution of sequence lengths. (c) The distribution of attribute values.}
\label{fig:delta_illumination}
\end{figure}

\clearpage
\onecolumn
\subsection{Delta Blur Bounding-box}

\begin{figure}[h!]
\centering
\includegraphics[width=0.95\linewidth]{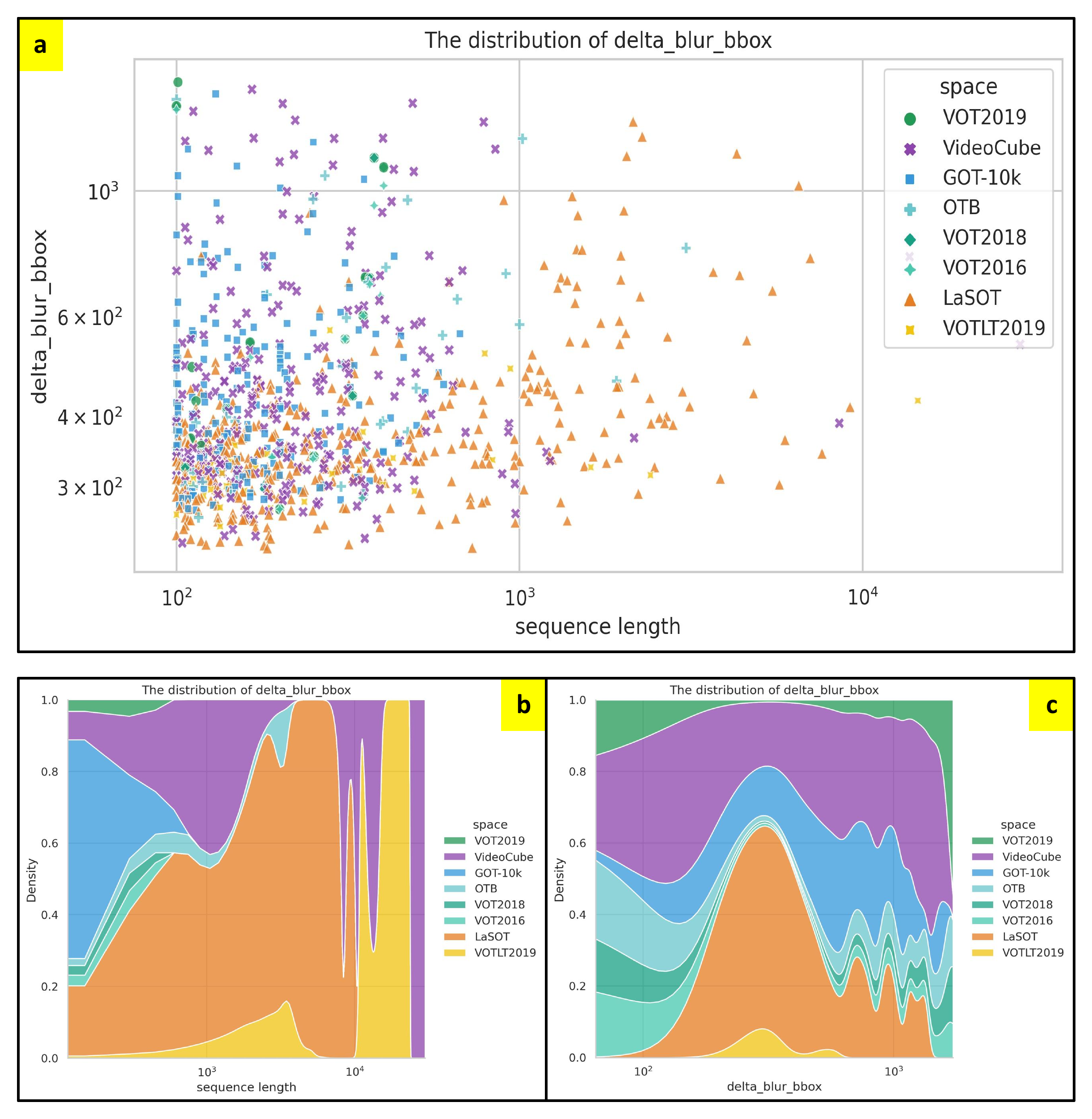}
\caption{The composition of delta blur bounding-box. (a) The distribution of attribute values and sequence lengths, each point representing a sub-sequence. (b) The distribution of sequence lengths. (c) The distribution of attribute values.}
\label{fig:delta_blur_bbox}
\end{figure}

\clearpage
\onecolumn
\subsection{Fast Motion}

\begin{figure}[h!]
\centering
\includegraphics[width=0.95\linewidth]{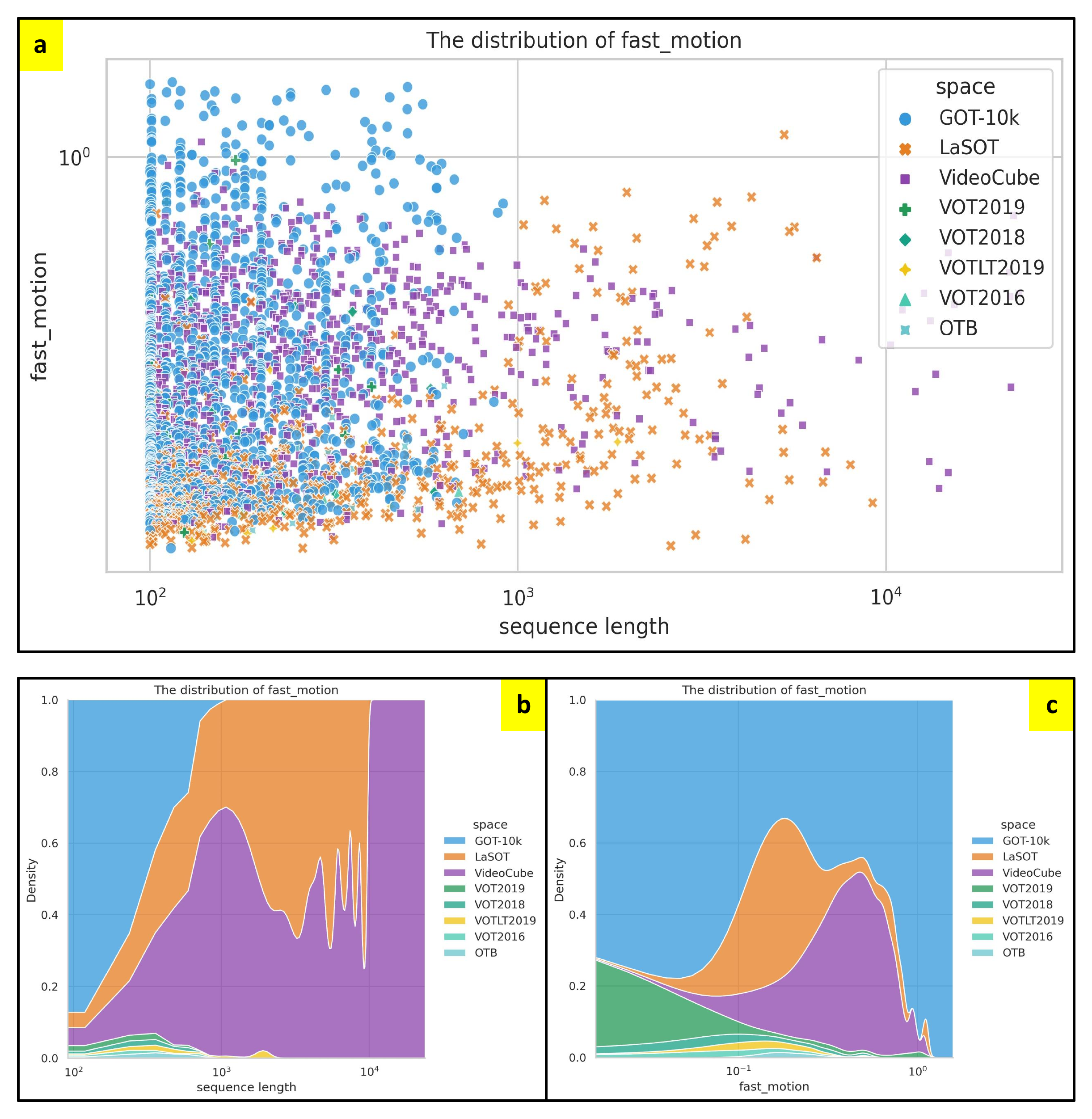}
\caption{The composition of fast motion. (a) The distribution of attribute values and sequence lengths, each point representing a sub-sequence. (b) The distribution of sequence lengths. (c) The distribution of attribute values.}
\label{fig:fast_motion}
\end{figure}

\clearpage
\onecolumn
\subsection{Low Correlation Coefficient}

\begin{figure}[h!]
\centering
\includegraphics[width=0.95\linewidth]{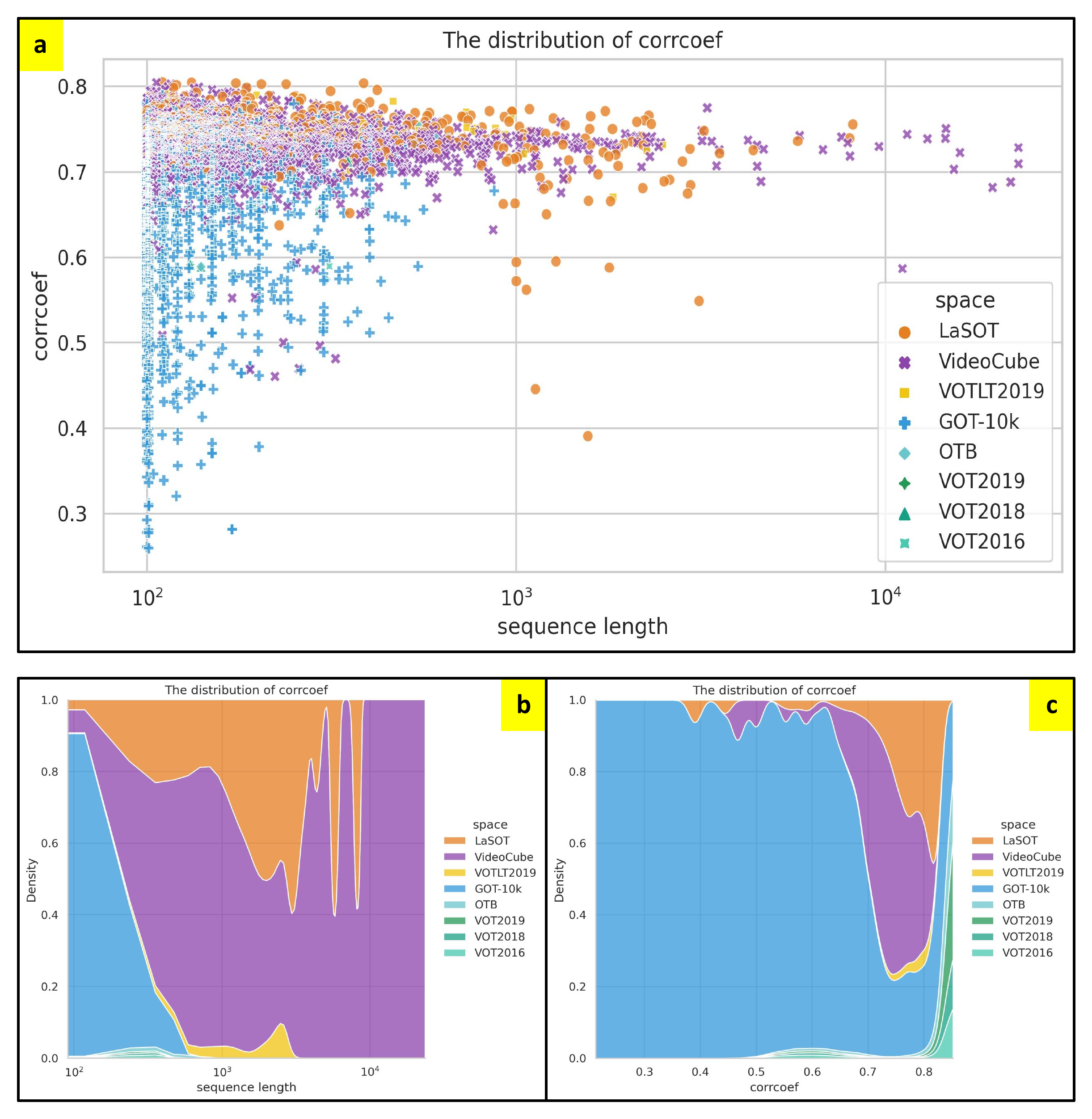}
\caption{The composition of low correlation coefficient. (a) The distribution of attribute values and sequence lengths, each point representing a sub-sequence. (b) The distribution of sequence lengths. (c) The distribution of attribute values.}
\label{fig:corrcoef}
\end{figure}

\clearpage
\onecolumn
\section{Experiments in Challenging Space}
\label{supsec:experiment-cs}

\subsection{Static Attributes}

\begin{figure*}[h!]
	\centering 
	\includegraphics[width=0.95\textwidth]{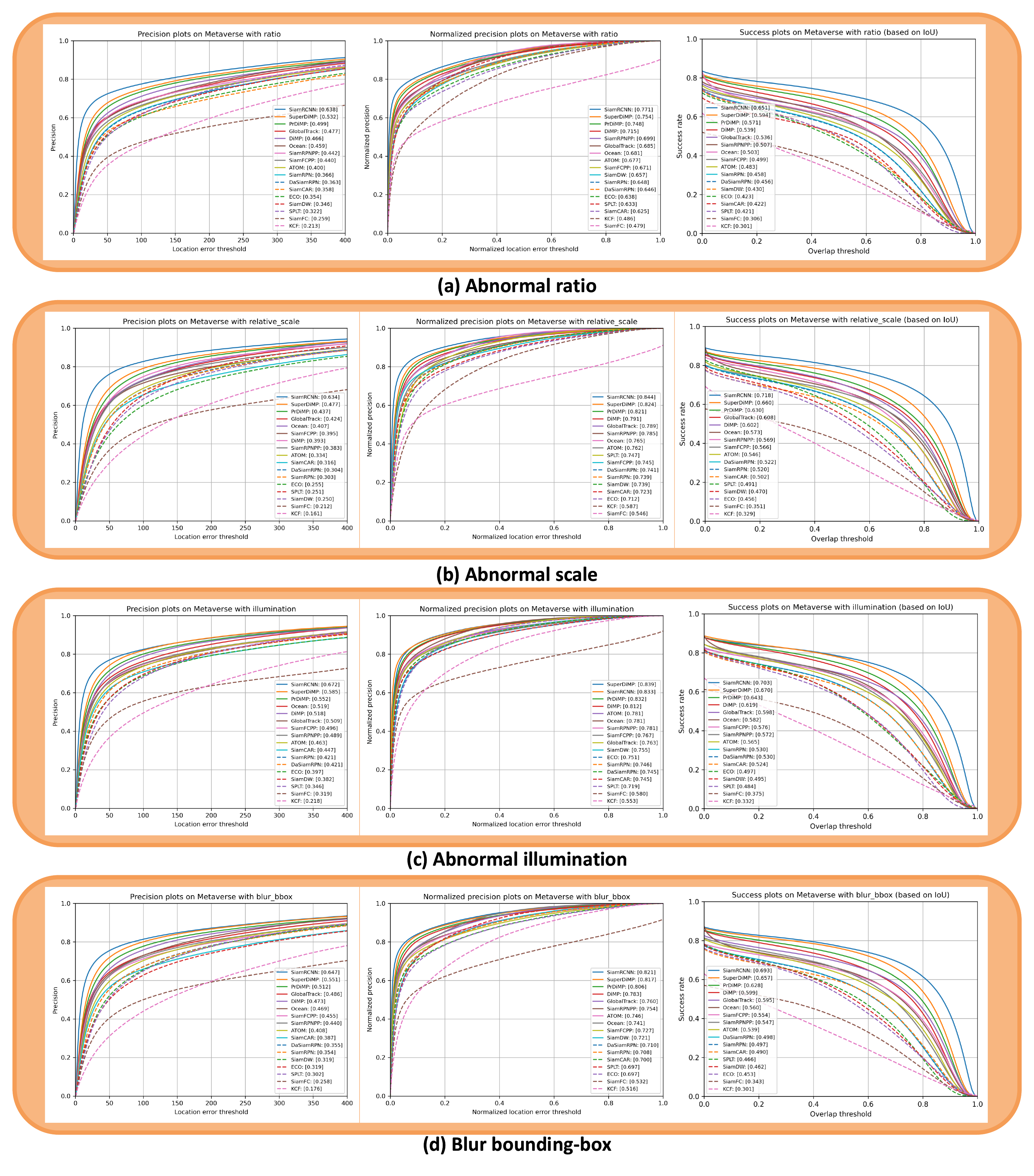}
	\caption{Experiments in challenging space with static attributes. (a, b, c, d) represent the tracking results in different challenging factors. Each task is evaluated by precision plot, normalized precision plot, and success plot with OPE mechanism. }
	\label{fig:challenging-static}
	\end{figure*}

\clearpage
\onecolumn
\subsection{Dynamic Attributes}

\begin{figure*}[h!]
	\centering 
	\includegraphics[width=0.95\textwidth]{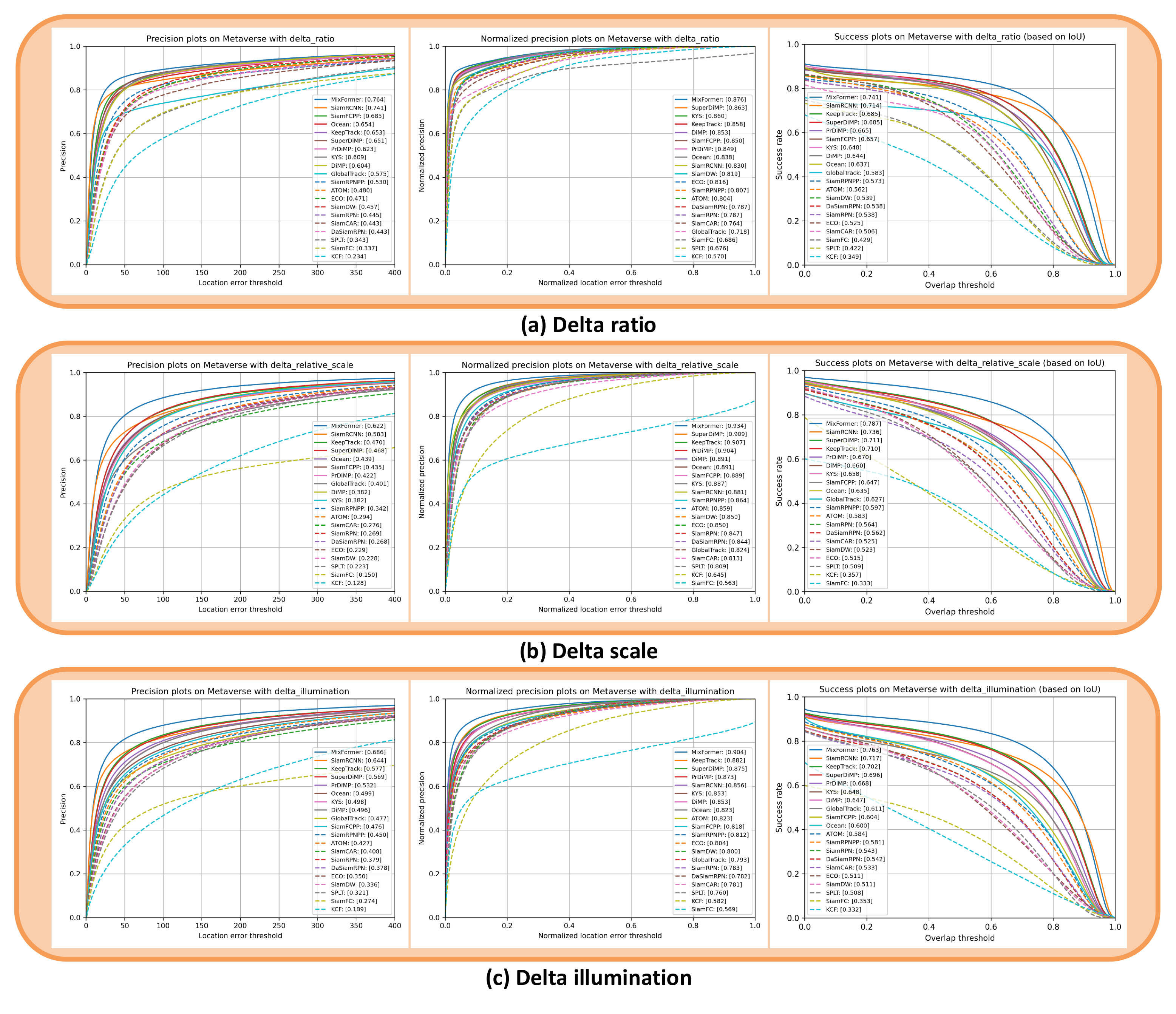}
	\caption{Experiments in challenging space with dynamic attributes. (a, b, c) represent the tracking results in different challenging factors. Each task is evaluated by precision plot, normalized precision plot, and success plot with OPE mechanism.}
	\label{fig:challenging-dynamic1}
	\end{figure*}

\clearpage
\onecolumn
\begin{figure*}[h!]
\centering 
\includegraphics[width=0.95\textwidth]{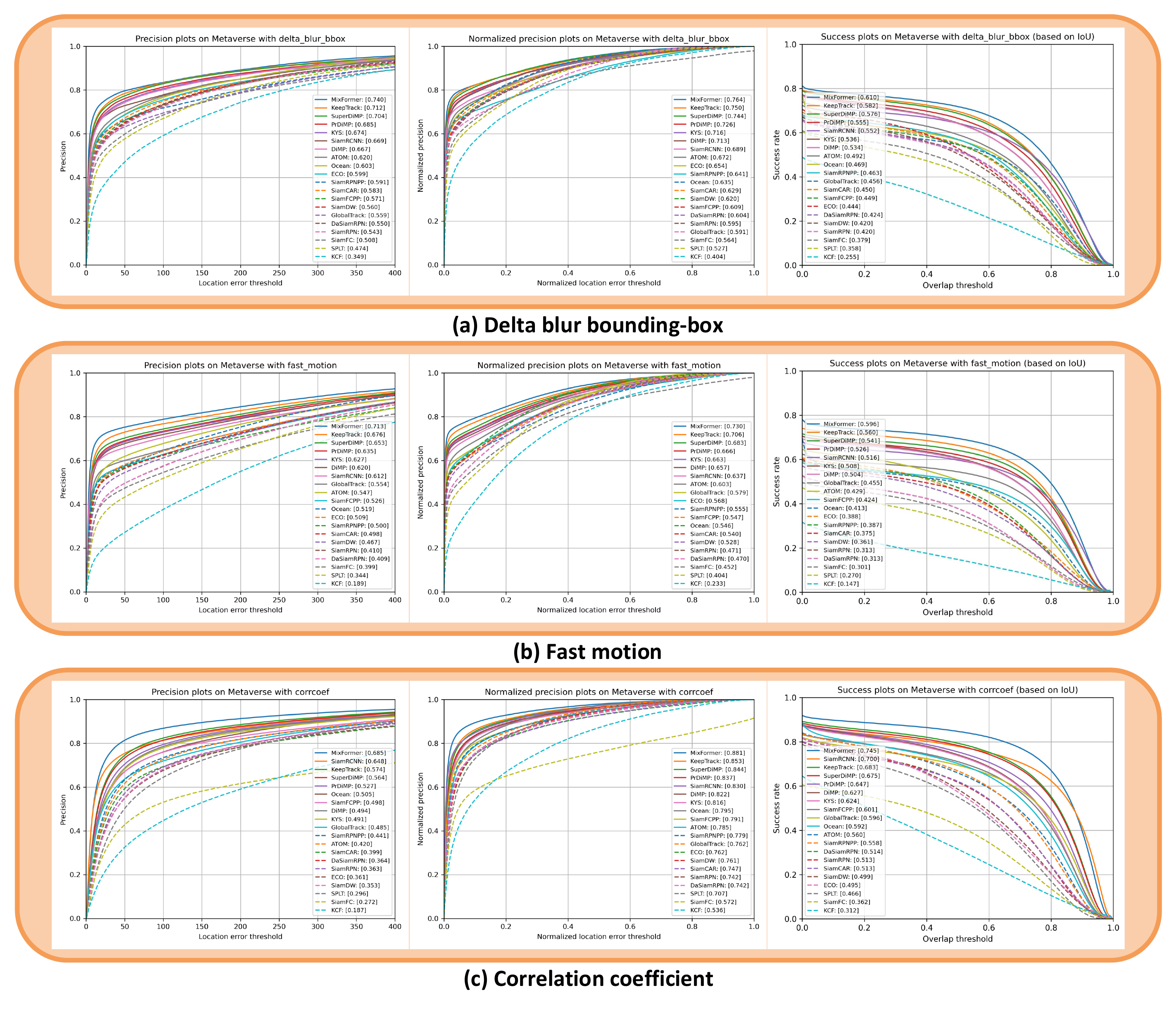}
\caption{Experiments in challenging space with dynamic attributes. (a, b, c) represent the tracking results in different challenging factors. Each task is evaluated by precision plot, normalized precision plot, and success plot with OPE mechanism.}
\label{fig:challenging-dynamic2}
\end{figure*}

\end{appendices}

\end{document}